\pdfoutput=1
\documentclass[10pt,twocolumn,letterpaper]{article}

\usepackage[pagenumbers]{cvpr}  

\usepackage{epsfig} 
\usepackage{times} 
\usepackage{amsmath} 
\usepackage{amssymb}  
\usepackage{graphicx}
\usepackage[pagebackref,breaklinks,colorlinks]{hyperref}
\usepackage{multirow}
\usepackage{siunitx}
\usepackage{array}
\usepackage{diagbox}
\usepackage{adjustbox}
\usepackage{booktabs}
\usepackage{pifont}
\usepackage{url}
\usepackage{colortbl}
\usepackage{caption}
\usepackage{subcaption}
\usepackage{cuted}

\usepackage{nicefrac} 
\usepackage{xcolor}
\usepackage{makecell}
\usepackage[inline]{enumitem}
\usepackage{microtype}
\usepackage{flushend}
\usepackage[linesnumbered,ruled,vlined]{algorithm2e}

\usepackage[accsupp]{axessibility}  

\definecolor{magenta}{RGB}{255,32,255}
\definecolor{green}{RGB}{0,255,0}
\definecolor{blue}{RGB}{0,0,255}
\definecolor{tablehighlight}{rgb}{0.8,0.8,0.8}

\newcommand{\xmark}{\ding{55}}%

\newcolumntype{P}[1]{>{\centering\arraybackslash}p{#1}}
\newcolumntype{M}[1]{>{\centering\arraybackslash}m{#1}}

\def\cvprPaperID{902}
\def\confName{CVPR}
\def\confYear{2023}

\def\paperTitle{Learning and Aggregating Lane Graphs for Urban Automated Driving}

\def\authorBlock{
    Martin Büchner${}^{1}$\thanks{Equal contribution}\quad
    Jannik Zürn${}^{1*}$\quad
    Ion-George Todoran${}^{2}$\quad
    Abhinav Valada${}^1$\vspace{0.1cm}\quad
    Wolfram Burgard\textsuperscript{3} \\

    ${}^1$University of Freiburg \quad\quad
    ${}^2$Woven by Toyota \quad\quad
    ${}^3$University of Technology Nuremberg
}

\newif\ifreview 
\newif\ifarxiv 
\newif\ifcamera 
\newif\ifrebuttal

\begin{document}
\title{\paperTitle}
\author{\authorBlock}

\maketitle

\begin{abstract}
Lane graph estimation is an essential and highly challenging task in automated driving and HD map learning. Existing methods using either onboard or aerial imagery struggle with complex lane topologies, out-of-distribution scenarios, or significant occlusions in the image space. Moreover, merging overlapping lane graphs to obtain consistent large-scale graphs remains difficult. To overcome these challenges, we propose a novel bottom-up approach to lane graph estimation from aerial imagery that aggregates multiple overlapping graphs into a single consistent graph. Due to its modular design, our method allows us to address two complementary tasks: predicting ego-respective successor lane graphs from arbitrary vehicle positions using a graph neural network and aggregating these predictions into a consistent global lane graph. Extensive experiments on a large-scale lane graph dataset demonstrate that our approach yields highly accurate lane graphs, even in regions with severe occlusions. The presented approach to graph aggregation proves to eliminate inconsistent predictions while increasing the overall graph quality. We make our large-scale urban lane graph dataset and code publicly available at \mbox{\small{\url{http://urbanlanegraph.cs.uni-freiburg.de}}}.
\end{abstract}

\section{Introduction}
\label{sec:introduction}

\begin{figure}
\centering
\includegraphics[width=8.3cm]{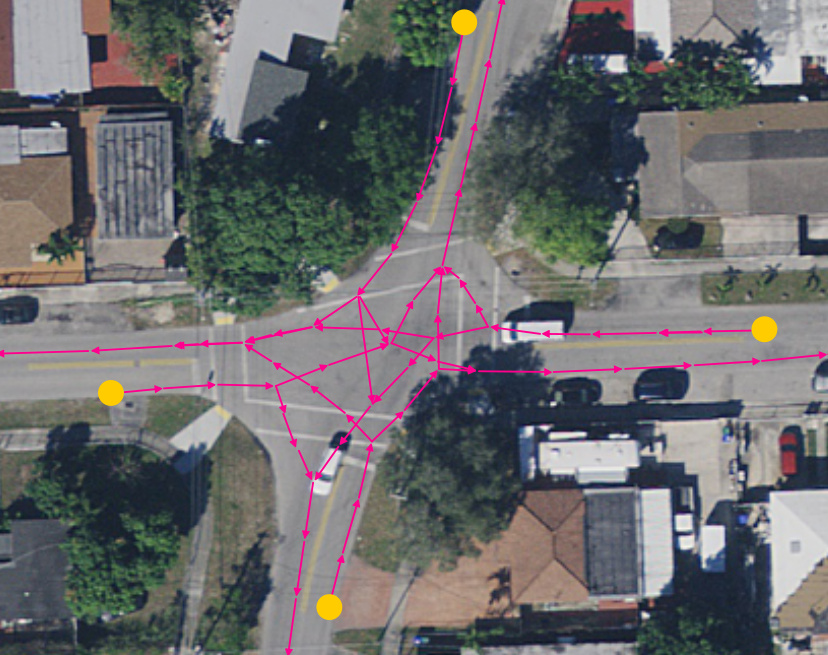}
\caption{Our approach predicts accurate lane graphs from aerial images of complex urban environments. We visualize the estimated lane graph in magenta and indicate model initialization points with yellow circles.}
\label{fig:covergirl}
\vspace{-0.3cm}
\end{figure}

Most automated driving vehicles rely on the knowledge of their immediate surroundings to safely navigate urban environments. Onboard sensors including LiDARs and cameras provide perception inputs that are utilized in multiple tasks such as localization~\cite{vodisch2022continual, petek2022robust, cattaneo2022lcdnet}, tracking~\cite{buchner20223d}, or scene understanding~\cite{zurn2022trackletmapper, mohan2022amodal, valada2016convoluted, vertens2022usegscene} to aggregate representations of the environment. However, robust planning and control typically require vastly more detailed and less noisy world models in the form of HD map data~\cite{gosala2022bird}. In particular, information on lane parametrization and connectivity is essential for both planning future driving maneuvers as well as high-level navigation tasks. 
Creating and maintaining HD maps in the form of lane graphs is a time-consuming and arduous task due to the large amount of detail required in the annotation and the data curation process including map updates based on local environment changes such as construction sites.

Previous approaches to lane graph estimation have shown shortcomings in predicting lane graphs due to multiple deficiencies: On the one hand, methods using onboard imagery typically degrade at complex real-world intersections and under significant occlusions, e.g., when following another vehicle~\cite{can2021structured,can2022topology}. On the other hand, methods based on aerial imagery show reduced performance when confronted with occlusions in the bird's-eye-view (BEV) due to, e.g., vegetation or shadows, and suffer from catastrophic drift when unconstrained in out-of-distribution scenarios~\cite{xu2022rngdet}. Previous works treat intersections and non-intersections inherently differently~\cite{he2022lane} and thus require elaborated heuristics and post-processing to merge single predictions into a consistent lane graph. Moreover, prior works do not focus on use cases where multiple predicted graphs must be merged into a single consistent solution, which is essential for enabling the automatic generation of highly detailed lane graphs of large contiguous regions.

Related to the aforementioned challenges, we propose a novel two-stage graph neural network (GNN) approach termed LaneGNN that operates on single aerial color images for lane graph prediction. Inspired by methods in the field of trajectory prediction~\cite{chai2020multipath}, we formulate a bottom-up approach according to which we place a virtual agent into a local crop of the aerial image and predict reachable successor lane graphs from its positions. To transform multiple disjoint local solutions into a single global solution, we aggregate a global representation by iteratively inferring the lane graph from consecutive poses, ultimately imitating real-world driving behavior. This iterative approach not only increases the predicted area covered but also improves graph accuracy based on data association and rejection. Note that we do not require any human in the loop to perform the graph aggregation. We visualize the output of our graph aggregation procedure in Fig.~\ref{fig:covergirl}, in which we superimpose the predicted graph on the aerial image input. Using this framework, we envision two applications: ego-centered successor path prediction and full lane graph estimation by aggregation.

\begin{figure*}[h]
\centering
\scriptsize
\includegraphics[width=\textwidth]{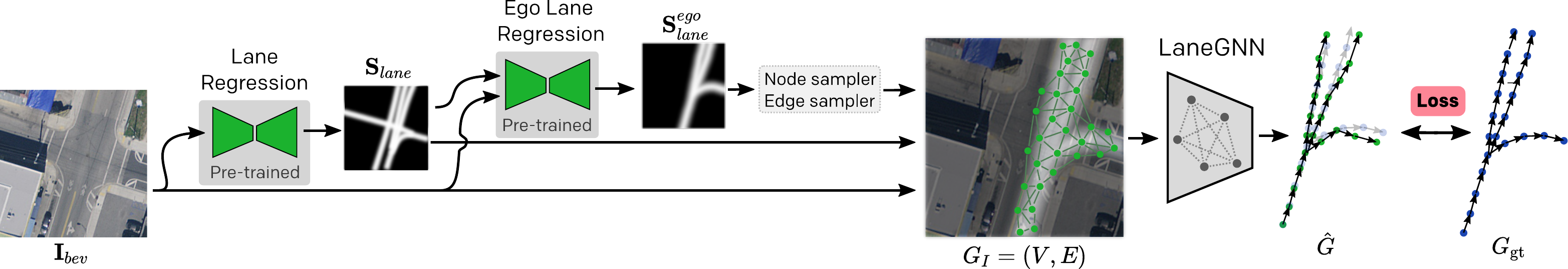}
\vspace{-0.3cm}
\caption{Overview of our LaneGNN model $\mathcal{M}$ predicting successor lane graphs. As a pre-training step, we train lane centerline and ego lane centerline regressor models. The ego lane regression $\mathbf{S}_{lane}^{ego}$ is used as a prior for sampling proposal nodes $V$ in corridors that have a high likelihood of entailing the successor graph. The model learns a binary classification of node and edge scores while also predicting the probability of a node being an endpoint of a given lane segment.}
\label{fig:approach-model}
\vspace{-0.3cm}
\end{figure*}

To summarize, the main contributions of this work are:
\begin{itemize}[noitemsep, topsep=0pt]
    \item An innovative bottom-up approach to lane graph estimation in challenging environments that explicitly encodes graph-level lane topology from input aerial images in a scenario-agnostic manner.
    \item A novel graph aggregation scheme enabling robust and method-agnostic merging of graph-level predictions.
    \item The large-scale lane graph dataset \textit{UrbanLaneGraph} comprising high-resolution aerial images aligned with dense lane graph annotations aggregated from the Argoverse2 dataset that we make publicly available.
    \item Extensive experiments and ablation studies demonstrating the significance of our findings.
\end{itemize}
\section{Related Works}
\label{sec:relatedworks}

In recent years, the prediction of topological road features such as road graphs and lane graphs have been extensively studied. In our discussion, we differentiate between road graph learning and lane graph learning. While road graphs encode the topological connections between road segments, lane graphs describe the locations and connectivity between all lanes, resulting in a spatially much denser graph. Many prior works focus on vehicle trajectory prediction, conditioned on HD map features such as lane centerline and boundary positions~\cite{djuric2021multixnet, chen2022learning, wang2022ltp}. These models do not aim at exclusively predicting the road or lane graph structure from onboard vehicle images or aerial images but instead at predicting future vehicle states such as position and orientation.\looseness=-1

{\parskip=3pt
\noindent\textbf{Road Graph Learning}: 
Prior works investigate estimating road graphs from both onboard sensors~\cite{liang2019convolutional} and from aerial images~\cite{zhou2018d, tan2020vecroad, bandara2022spin} or focus on extracting pixel-level road segmentation from images and extracting graphical road representations, i.e., using morphological image operators or graph neural networks to extract the connectivity between different roads within the image~\cite{mattyus2017deeproadmapper, bandara2022spin}. Other approaches investigate iterative methods and interpret road graph prediction as a sequential prediction task~\cite{bastani2018roadtracer, tan2020vecroad}.
}

{\parskip=3pt
\noindent\textbf{Lane Graph Learning from Vehicle Data}:
Some earlier works in lane graph learning from onboard vehicle sensors such as cameras and LiDAR formulate lane extraction as an image-based lane centerline regression task~\cite{homayounfar2018hierarchical}. Homayounfar~\etal~\cite{homayounfar2019dagmapper} aggregate onboard LiDAR data on highways and leverage a recurrent neural network to generate highway lane graphs in an iterative manner. Zhou~\etal~\cite{zhou2021automatic} utilize the OpenStreetMap database and a semantic particle filter to accumulate projected semantic predictions from a vehicle ego-view into a map representation. Zhang~\etal~\cite{zhang2021hierarchical} propose an online road map extraction system for a sensor setup onboard a moving vehicle and construct a graph representation of the road network using a fully convolutional neural network. More recently, Can~\etal proposed two different methods~\cite{can2021structured, can2022topology} for lane connectivity learning in intersection scenarios from onboard camera images.\looseness=-1

\noindent\textbf{Lane Graph Learning from Birds-Eye-View Data}:
Despite the advantages of leveraging readily available aerial data for training lane graphs, only a few works considered using aerial images as an input modality for the graph learning task. Z\"urn~\etal~\cite{zurn2021lane} propose a lane centerline regression model jointly with a Graph R-CNN backbone to predict nodes and edges of the lane graph from a local aggregated bird's-eye-view image crop. More recently, He~\etal~\cite{he2022lane} propose a two-stage graph estimation pipeline. They first extract lanes at non-intersection areas and subsequently predict the connectivity of each pair of lanes, and extract the valid turning lanes to complete the map at intersections.

In contrast to existing works, we do not estimate the complete lane graph visible in a given crop but only the part of the graph that is reachable from a virtual agent pose located within the crop, simplifying the graph estimation problem based on reduced topological complexity. Furthermore, we leverage a GNN to explicitly model the relationships between graph nodes, allowing us to leverage recent developments in the field of geometric deep learning. This explicit graph encoding and prediction allows us to formulate a model that does not internally differentiate between intersection areas and non-intersection areas, in contrast to some related works. Additionally, we propose a novel large-scale dataset for lane graph estimation from aerial images, allowing the research community to easily evaluate and compare their approaches. 
}

\section{Dataset}
\label{sec:dataset}

\begin{table}[t]
\scriptsize
\centering
\setlength\tabcolsep{3.7pt}
 \begin{tabular}{l|ccccc}
 \toprule
 \multirow{2}{*}{Dataset} & \multirow{2}{*}{City} & Lane & Total \\
& & Splits/Merges & Length [km] \\
 \midrule
 & Palo Alto    & 4752   &  796.4 \\
 & Austin       &  8495  &  531.7 \\
UrbanLaneGraph & Miami        &  7642  &  850.8 \\
\textit{(from Argoverse2)} & Pittsburgh   & 14610 & 1314.3  \\
&  Washington D.C.  & 2066  & 739.6  \\
& Detroit      & 5424  & 990.5 \\
 \midrule
 \multirow{2}{*}{LaneExtraction \cite{he2022lane}} & Boston, Seattle,   & \multirow{2}{*}{2262} & \multirow{2}{*}{398.6} \\
 & Phoenix, Miami   &  \\
  \midrule
NuScenes\cite{fong2022panoptic} & Boston &  1630 & 76.9  \\
 \bottomrule
 \end{tabular}
\caption{Key statistics of our \textit{UrbanLaneGraph} dataset, aggregated from the Argoverse2 dataset \cite{wilson2021argoverse}, used for our experiments, compared with other recent datasets for lane graph estimation.}
\label{tab:dataset-stats}
\vspace{-0.3cm}
\end{table}

To evaluate our approach on challenging real-world data, we compiled the \textit{UrbanLaneGraph} dataset. It is a first-of-its-kind dataset for large-scale lane graph estimation from aerial images. The dataset contains aerial images from the cities of Austin, Miami, Pittsburgh, Palo Alto, Detroit, and Washington DC. The images have a resolution of $\SI{15}{cm}$ per pixel. To obtain the corresponding lane annotations, we leverage the Argoverse2 dataset \cite{wilson2021argoverse} which entails lane graphs for large sections of the respective cities. The lane graphs in the  Argoverse2 dataset are provided on a per-scenario basis, covering only small local areas in one graph. Therefore, we collect all local lane graphs of each city and aggregate them into one globally consistent graph per city, thereby removing inconsistent or redundant nodes or edges. The annotated regions feature a diverse range of environments including urban, suburban, and rural regions with complex lane topologies. Accumulated over all cities, the overall length of all lanes spans over $\SI{5,000}{km}$. We split each city into disjoint training and testing regions. We list key dataset statistics in Tab.~\ref{tab:dataset-stats}, indicating the scale of our generated dataset compared with other recently proposed datasets containing graph annotations. For more details on the dataset and exemplary visualizations, please refer to Sec.~\ref{supp:dataset_details} in the supplementary material.\looseness=-1

\begin{figure*}
\centering
\scriptsize
\includegraphics[width=\textwidth]{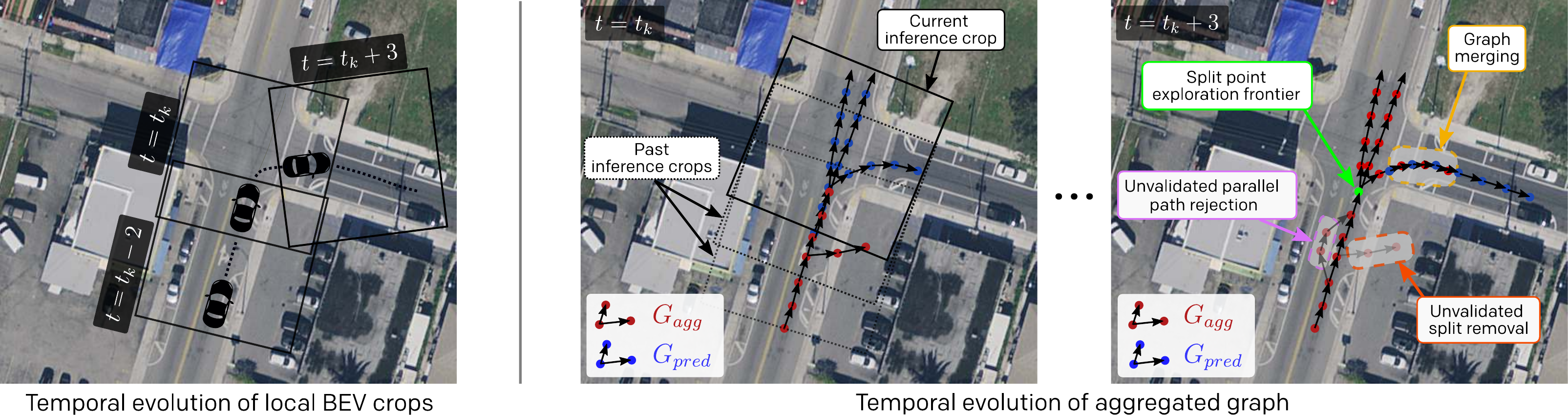}
\vspace{-0.3cm}
\caption{In our graph aggregation procedure, we iteratively obtain oriented image crops based on virtual agent poses along the currently predicted successor graph. For each crop, our LaneGNN model predicts a successor graph $G_{pred}$, which is aggregated into a globally consistent lane graph $G_{agg}$.}
\label{fig:approach-agg}
\vspace{-0.3cm}
\end{figure*}

\section{Technical Approach}
\label{sec:approach}

Our approach is divided into two stages: lane graph learning and lane graph aggregation. In the first stage (Sec.~\ref{subsec:lglearning}), we train our GNN model, denoted as LaneGNN, to predict the successor lane graph, entailing the nodes and edges that can be logically visited from the pose of a virtual vehicle agent. In the second phase (Sec.~\ref{subsec:lgaggregation}), we use our trained LaneGNN model to traverse a large map area. This is achieved by selecting an initial starting pose and predicting the successor lane graph from this pose. Subsequently, we iteratively estimate the traversable lane graph from the current pose and move forward along the predicted graph while aggregating. 
In the following, we detail both stages of our approach.


\subsection{Lane Graph Learning}
\label{subsec:lglearning} 

We formulate the task as a supervised learning problem where a successor lane graph $\hat{G}$ is estimated based on an aerial image $\mathbf{I}_{bev}$. First, a directed graph $G_{I}=(V, E)$ covering relevant regions is constructed by sampling from likely regions of $\mathbf{I}_{bev}$ (see Fig.~\ref{fig:approach-model}). For all our models and experiments we choose a spatial resolution of $256 \times 256$ pixels. The graph comprises a set of nodes $i \in V$ that are connected via directed edges $E \subseteq \{(i, j) | (i, j) \in V^2 \text{ and } i \neq j\}$ that constitute potentially valid lane graph edges. The graph is attributed using both node features $\mathbf{X} \in \mathbb{R}^{\|V\| \times D}$ and edge features $\mathbf{X}_{e} \in \mathbb{R}^{\|E\| \times (D_{geo} + D_{bev})}$, where $D_{geo}$ and $D_{bev}$ denote the dimensionality of the involved edge features (see Sec.~\ref{subsubsec:mp}). The overall model estimates an output graph $\hat{G} = \mathcal{M}(G_{I}|\theta)$, where $\mathcal{M}$ is parameterized by network model weights $\theta := (\theta_{reg}, \theta_{GNN})$.

{\parskip=5pt
\noindent\textbf{Lane Regression and Graph Construction:}\label{subsubsec:graphconstr}
We train two regression networks: Firstly, a centerline regression network predicting the likelihood map of lane centerlines $\mathbf{S}_{lane}$, and secondly, a segmentation network predicting all reachable lanes $\mathbf{S}_{lane}^{ego}$ starting from the initial virtual agent pose at the bottom center of $\mathbf{I}_{bev}$. We use identical PSPNet~\cite{zhao2017pyramid} architectures with a ResNet-152 feature extractor for this task. We sample equally-distributed node positions using Halton sequences~\cite{halton1964radical} that are later filtered based on the obtained ego-lane segmentation mask $\mathbf{S}_{lane}^{ego}$, which serves as a region of interest for sampling (see Fig.~\ref{fig:approach-model}). Directed edges $E$ among nodes are initialized for pairs of nodes with a Euclidean distance $d_{ij} \in [d_{min}, d_{max}]$. Initial node features $\mathbf{X}$ are crafted solely based on their 2D positions $\mathbf{x}_{i} = (x_i, y_i)$, while geometric edge features are defined as

\begin{equation}\small
    \mathbf{x}_{ij,geo} = \left(\operatorname{\tan^{-1}}\frac{\Delta y_{ij}}{\Delta x_{ij}}, \sqrt{{\Delta x_{ij}}^2 + {\Delta y_{ij}}^2}, \overline{x}_{ij}, \overline{y}_{ij}\right),
\end{equation}
where $\Delta x_{ij}$ and $\Delta y_{ij}$ represent node position differences and $\overline{x}_{ij}$, $\overline{y}_{ij}$ are edge middle point coordinates. In addition to the geometric edge feature, we generate aerial edge features: Per edge, a small oriented region of $\mathbf{I}_{bev}$ and the lane segmentation $\mathbf{S}_{lane}$ is obtained based on the direction of the edge, which provides $\mathbf{X}_{ij,bev} = [\mathbf{I}_{bev}^{*}, \mathbf{S}_{lane}^{*}]$ with dimension $D_{bev} = 4 \times 32 \times 32$.

\noindent\textbf{Feature Encoding and Message Passing:}\label{subsubsec:mp}
In our approach, we estimate edge probabilities, node probabilities, and whether a node is terminal. While the nodes themselves hold only unidirectional information, we encode the notion of direction using the proposed edge feature as outlined above.
We utilize a causal variant of neural message passing as proposed by Bras\'{o}~\emph{et~al.}~\cite{braso2020}. By imposing a causality prior, our network encodes predecessor and successor features during message passing. This formulation of message passing renders our approach direction-aware. Initial node and geometric edge features are encoded using multi-layer perceptrons $f_{enc}^{\Box}$ (MLP) while the aerial edge feature is transformed using a ResNet-18 architecture $f_{enc}^{e,bev}$. The geometric and aerial edge features are concatenated and fused subsequently to arrive at various node and edge embeddings $\mathbf{H}_{v}^{(0)}$ and $\mathbf{H}_{e}^{(0)}$: 
\begin{equation}\small
    f_{enc}^{e,bev}(\mathbf{X}_{e,bev}) = \mathbf{H}_{e,bev}^{(0)}\,, \,\,
    f_{enc}^{e,geo}(\mathbf{X}_{e,geo}) = \mathbf{H}_{e,geo}^{(0)}\,, 
\end{equation}
\begin{equation}\small
    f_{enc}^{v}(\mathbf{X}) = \mathbf{H}_{v}^{(0)},  \,\, 
    f_{fuse}^{e}([\mathbf{H}_{e,geo}^{(0)}, \mathbf{H}_{e,bev}^{(0)}]) = \mathbf{H}_{e}^{(0)}.
\end{equation}
In the following, multiple message-passing steps are performed using various ReLU-activated MLPs denoted by $f_{\Box}$ as follows.
The edge feature is updated based on the current neighboring node features $\mathbf{h}_{i}^{(l-1)}$, $\mathbf{h}_{j}^{(l-1)}$ and the edge feature $\mathbf{h}_{ij}^{(l-1)}$:
\begin{equation}\small
\mathbf{h}^{(l)}_{ij} = f_{e}\left(\left[\mathbf{h}_{i}^{(l-1)},\mathbf{h}_{j}^{(l-1)},\mathbf{h}_{ij}^{(l-1)}\right]\right).
\end{equation}
Messages $\mathbf{m}_{ij}^{(l)}$ are crafted based on either predecessors $\mathcal{N}_{\mathit{pred}}(i)$ or successors $\mathcal{N}_{\mathit{succ}}(i)$ of a node $i$ and propagated based on the constructed adjacency. In the next step, all predecessors and successor messages, respectively, are aggregated using a permutation-invariant sum:
\begin{equation}\small
\mathbf{h}_{i, \mathit{pred}}^{(l)} = \sum_{j \in \mathcal{N}_{\mathit{pred}}(i)} \underbrace{f_{v}^{\mathit{pred}}\left(\left[\mathbf{h}_{j}^{(l-1)}, \mathbf{h}_{ij}^{(l)}, \mathbf{h}_{j}^{(0)}\right]\right)}_{\mathbf{m}_{ji}^{(l)}} ,
\end{equation}
\begin{equation}\small
\mathbf{h}_{i, \mathit{succ}}^{(l)} = \sum_{j \in \mathcal{N}_{\mathit{succ}}(i)} \underbrace{f_{v}^{\mathit{succ}}\left(\left[\mathbf{h}_{j}^{(l-1)}, \mathbf{h}_{ij}^{(l)}, \mathbf{h}_{j}^{(0)}\right]\right)}_{\mathbf{m}_{ij}^{(l)}}.
\end{equation}
Note that the message crafting includes skip connections to initial node embeddings $\mathbf{h}_{j}^{(0)}$. Nodes are updated by combining the two features using concatenation:
\begin{equation}\small
\mathbf{h}_{i}^{(l)}=f_{v}\left(\left[\mathbf{h}_{i, \mathit{pred}}^{(l)}, \mathbf{h}_{i, \mathit{succ}}^{(l)}\right]\right).
\end{equation}
Finally, sigmoid-valued edge scores $\hat{e}_{ij}$ and node score $\hat{s}_i$ are predicted from the obtained edge and node embeddings. In a separate network head, we classify each node as being a terminal node or not, denoted as a scalar $\hat{t}_i$. Therefore, we optimize the following combined binary cross entropy:
\begin{equation}\small
    \mathcal{L} = - \sum_{|V|} s_i \log \hat{s}_i - \sum_{|V|} t_i \log \hat{t}_i - \sum_{|E|} e_{ij} \log \hat{e}_{ij}.
\end{equation}
The ground truth graph $G_{GT}$ as a learning target ($s_i$, $t_i$, $e_{ij}$) is generated based on the map annotations for the given cropped region and the corresponding closest nodes. This is further outlined in the supplementary material in Sec.~\ref{supp:sampling}.
}

\subsection{Iterative Temporal Graph Aggregation}
\label{subsec:lgaggregation}

In the second stage of our approach, we aggregate local successor graphs into a globally consistent lane graph. First, we prune the predicted per-crop lane graph, and second, we  iteratively aggregate the sparse graphs. In the following, we detail both components.

\begin{figure}
\centering
\scriptsize
\includegraphics[width=8.3cm]{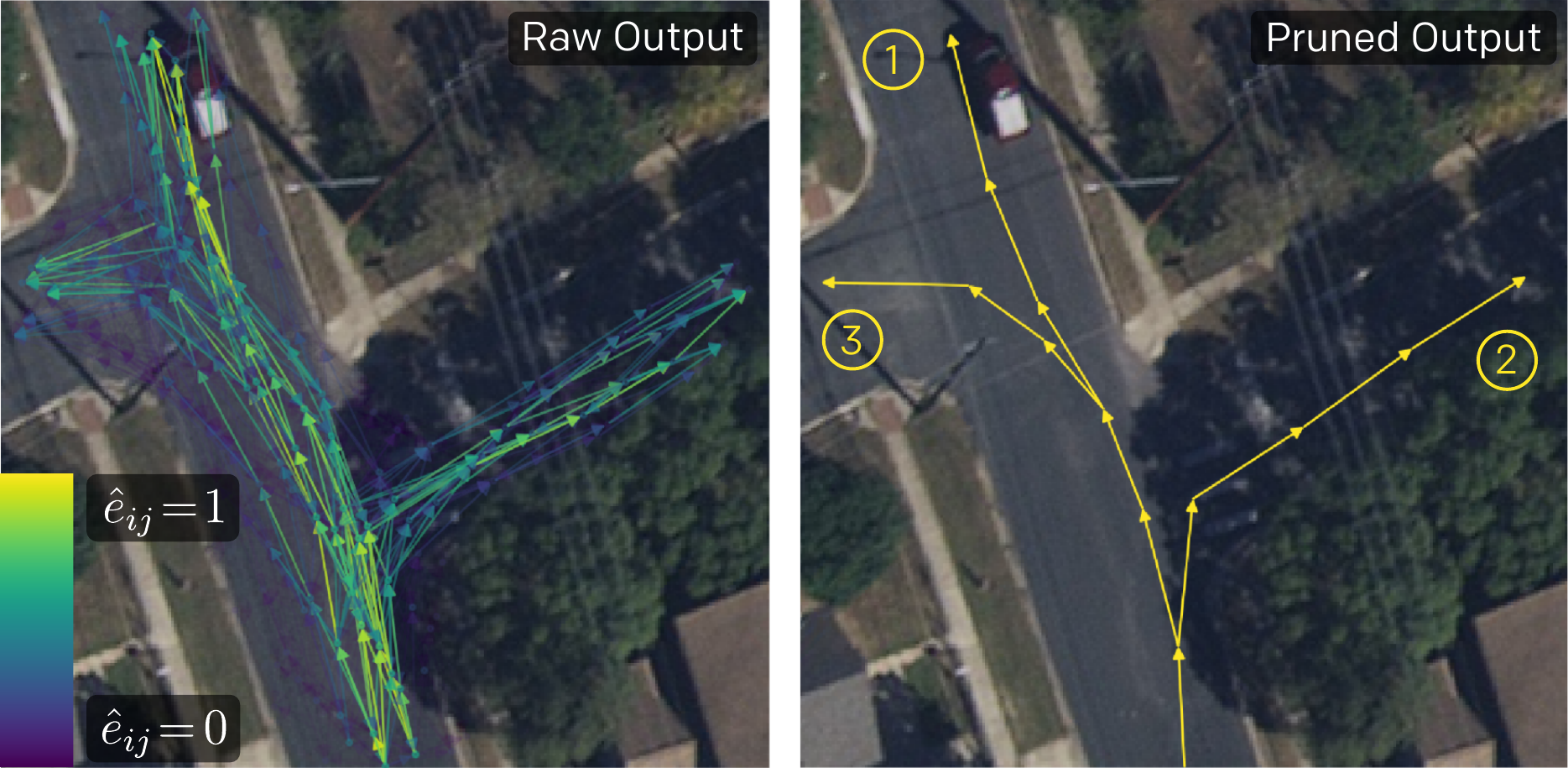}
\vspace{-0.3cm}
\caption{Comparison of raw and pruned lane graph predictions. On the left-hand side, we visualize estimated edge scores $\hat{e}_{ij}$ and node scores $\hat{s}_{i}$ using the same color scaling. On the right-hand side, we visualize the pruned graph. The circled numbers denote the order of traversal based on estimated terminal node scores.}
\label{fig:pruning}
\vspace{-0.4cm}
\end{figure}

{\parskip=5pt
\noindent\textbf{Pruning:} 
The graph prediction obtained from the LaneGNN model follows the graph connectivity initially generated during sampling. Since the model prediction contains a number of redundant paths with high predicted node and edge scores, we prune the obtained solution to obtain sparse lane representations. We formulate the graph pruning problem as a search problem from a starting node to possibly multiple predicted terminal nodes (see Sec.~\ref{subsec:lglearning}). Predicted lane split points should coincide with actual split points. Thus, different branches should share the same set of edges up to a split point. We use Dijkstra's algorithm to iteratively find high-score paths between the initial pose and terminal nodes, ordered from high to low scores until all terminal nodes are reached. Edge scores contained in found paths are set to zero. In Fig.~\ref{fig:pruning}, we visualize the output of this step.

\noindent\textbf{Aggregation:} 
We iteratively aggregate predicted successor lane graphs $G_{\mathit{pred}}=(V_{\mathit{pred}}, E_{\mathit{pred}})$ into a globally consistent and complete graph $G_{\mathit{agg}}=(V_{\mathit{agg}}, E_{\mathit{agg}})$ as depicted in Fig.~\ref{fig:approach-agg}. The predicted successor graph $G_{\mathit{pred}}^{ t}$ is added to the current aggregated graph $G_{\mathit{agg}}^{t-1}$ at time step $t$: $G_{\mathit{agg}}^{ t} \leftarrow \texttt{aggregate}(G_{\mathit{pred}}^{t}, G_{\mathit{agg}}^{t-1})$.
Plausible branches of $G_{\mathit{pred}}$ are merged into $G_{\mathit{agg}}$ and thus extend the aggregated graph with every iteration if new grounds are covered in the respective crops. 
The next virtual agent pose is extracted from the set of forward-facing edges of $G_{\mathit{agg}}$ given the current pose. Only edges with a significant weight due to previous aggregations are selected for this set.
}{\parskip=0pt
As the node positions differ slightly with each model forward pass we observe roughly similar paths in the lateral sense wrt.\ the ground-truth graph. However, along the longitudinal dimension of a branch, we observe deviations in node position, which must be circumvented when aggregating the graph. Based on this, we only take the lateral deviation wrt.\ $G_{\mathit{agg}}^{t}$
into account when merging $G_{\mathit{pred}}^{t}$. Thus, $G_{\mathit{agg}}^{t}$ is only updated in a lateral sense while the longitudinal misalignment of the two sets of nodes is neglected (see also Sec.~\ref{supp:aggregation} in the supplementary material). The nodes of $G_{\mathit{pred}}^{t}$ have a weight of $1$ while a node of $G_{\mathit{agg}}^{t}$ holds a weight equal to the number of merges it has observed so far. If a novel node $i \in V_{\mathit{pred}}$ is not close to any other node in $k \in V_{\mathit{agg}}$ it is added to the aggregated graph including its incident edges. This ultimately allows a weighting-based merging of arbitrary pairs of graphs, which is used for global lane graph estimation (see Sec.~\ref{sec:experiments-fulllgp}).

We observe that the more graphs we aggregate, the more we are certain about the significance of particular graph branches. Following a weighting-based approach allows us to set certain thresholds that allow modification of $G_{\mathit{agg}}$. Thus, implausible graph branches, semantically similar parallel paths, redundant edges, and isolated nodes are deleted based on confidence and distance as well as angle criteria (see Fig.~\ref{fig:approach-agg}). As a result, we are able to decrease the number of false positive split and merge points to obtain more consistent global graphs. We observe that this approach greatly improves lane graph prediction accuracy in difficult occluded and out-of-distribution scenarios since the sum of model predictions covering the same region shed light onto what potentially constitutes, e.g., a valid and an invalid branch.\looseness=-1

Multiple forward passes naturally lead to a multitude of lane split points (both true positive and false positive splits). We interpret each split point as an element of an exploration frontier. A queue of unexplored, high-probability graph branches is maintained, which is queried in a depth-first manner as soon as the currently traversed branch terminates. Following the weighting-based approach, a branch is only explored if its depth-limited oriented successor tree weight exceeds a certain level of confidence. Due to this flexible approach, we can essentially handle arbitrary lane graph topologies with a single holistic approach. For more details on the graph pruning and aggregation procedures, please refer to the supplementary material.
Finally, we apply multiple iterations of Laplacian smoothing to $G_{\mathit{pred}}^{t}$, which modifies the original node positions in order to even out position irregularities caused by sampling while keeping the adjacency represented by $E_{\mathit{pred}}^{t}$ constant.
}

\section{Experimental Results}
\label{sec:experiments}

In the following, we present our experimental findings. We first define and illustrate three tasks on which we benchmark our method. Subsequently, we describe the evaluation metrics and compare against other methods as well as our own baselines. We provide extensive qualitative and quantitative evaluations on our \textit{UrbanLaneGraph} dataset.

\begin{table*}
\centering
\scriptsize
\setlength\tabcolsep{3.7pt}
\caption{Quantitative results of our model including ablation studies, in comparison with baseline models for the Successor-LGP task on our \textit{UrbanLaneGraph} dataset. P/R denotes Precision/Recall. For our LaneGNN model variants, we denote CMP as causal message passing, aerial node features as AerN, aerial edge features as AerE, and the ego-lane regression with $\mathbf{S}_{lane}^{ego}$. For all metrics, higher values mean better results.}
\begin{tabular}{p{3cm}|m{0.4cm}p{0.4cm}p{0.4cm}p{0.6cm} | p{1.8cm} p{1.8cm} p{1.4cm} p{1.0cm} p{1.0cm} p{1.7cm}}
 \toprule
 Model &   CMP &  AerN & AerE & $\mathbf{S}_{lane}^{ego}$ &  TOPO P/R$\,\uparrow$ & GEO P/R$\,\uparrow$ & APLS$\,\uparrow$ & SDA$_\text{20}$$\,\uparrow$ & SDA$_\text{50}$$\,\uparrow$ & Graph IoU$\,\uparrow$ \\
 \midrule
Skeletonized Regression          & -- & -- & --  & --	&  0.597/0.613 & 0.578/0.601	& \textbf{0.315} & 0.020 & 0.185  & 0.180 \\
LaneGraphNet \cite{zurn2021lane} & -- & -- & --  & --	&  0.0/0.0 & 0.0/0.0	& 0.179 & 0.0 & 0.0  &  0.063 \\
 \midrule
 \multirow{5}{*}{LaneGNN (ours)} & \xmark      & \xmark & \checkmark   &  \checkmark    & 0.549/0.677 & 0.548/0.671 & 0.188 & 0.168 & 0.323 & 0.312  \\  
 & \checkmark  & \checkmark     & \xmark   &  \checkmark	  & 0.562/0.656 & 0.562/0.655	& 0.192 & 0.151 & 0.298 & 0.320  \\  
 & \checkmark  & \xmark & \xmark       &  \checkmark   & 0.545/0.693 & 0.545/0.688   & 0.200 & 0.188 & 0.311 & 0.336  \\  
 & \checkmark  & \xmark & \checkmark   &  \xmark 	  & 0.578/0.669 & 0.577/0.659   & 0.150 & 0.132 & 0.227 & 0.250  \\  
 & \checkmark  & \xmark & \checkmark   &  \checkmark   & \textbf{0.600}/\textbf{0.699} & \textbf{0.599}/\textbf{0.695} &  0.202 & \textbf{0.227} & \textbf{0.377} & \textbf{0.347}  \\  
 \bottomrule
\end{tabular}
\label{tab:results-successor}
\vspace{-0.3cm}
\end{table*}

\subsection{Proposed Tasks}

We propose three distinct and complementary tasks. In the first task, successor lane graph prediction \textit{(Successor-LGP, Sec.~\ref{sec:succlgp})}, we aim at predicting a feasible ego-reachable successor lane graphs from the current pose of the virtual agent. The purpose of the Successor-LGP task is to measure the prediction quality of potential future driving paths when no HD map coverage is available. In the second task, full lane graph prediction (\textit{Full-LGP}, Sec.~\ref{sec:experiments-fulllgp}), we evaluate the quality of regionally aggregated lane graphs in the context of HD map estimation. This task aims at measuring the predictive power of our full two-stage model performing lane graph inference and graph aggregation in conjunction. For such purposes, the full ground truth lane graph of a given map area is compared to the aggregated predictions of our model. 
Finally, we carry out a high-level path planning task (Sec.~\ref{sec:planning}) on the predicted lane graphs, intended to analyze the fidelity of routes planned on the predicted graphs.

\subsection{Evaluation Metrics}

We leverage multiple complementary metrics for performance evaluation as detailed below.

{\parskip=0pt
\noindent\textbf{Graph IoU:} This metric measures the intersection-over-union (IoU) of two graphs rendered as a binary image~\cite{zurn2021lane}, where pixels closer than $d=5$ pixels are assigned the label $1$ and the remaining pixels the label $0$. Equivalent to the evaluation of semantic segmentation models, we determine the IoU values for the non-zero pixels.

\noindent The \textbf{APLS metric} sums the differences in optimal path lengths between nodes in the ground truth graph $G$ and the proposal graph $G^\prime$ \cite{van2018spacenet}. The APLS metric scales from $0$ (worst) to $1$ (best). Formally, it is defined as
\begin{equation}\small
    \text{APLS} = 1 - \frac{1}{N_p} \sum \min \Big\{ 1, \frac{|d(v_1,v_2) - d(v_1',v_2') |}{d(v_1,v_2)} \Big\}, 
\end{equation}
where $v_i$ and $v_i'$ are nodes in $G$ and $G'$, respectively. $N_p$ denotes the number of paths in $G$ and $d(\cdot,\cdot)$ is the path length. For more details, please refer to \cite{van2018spacenet}.

\noindent\textbf{TOPO / GEO metrics:} Following previous works in road network extraction and lane graph estimation, we use the GEO metric and the TOPO metric. For definitions and details on these metrics, please refer to \cite{he2022lane} and to the supplementary material.

\noindent\textbf{Split Detection Accuracy (SDA$_\text{R}$):} This metric evaluates how accurately a model predicts the lane split within a circle of radius $R$ pixels from a given ground truth lane split. 
}

\subsection{Successor Lane Graph Prediction}
\label{sec:succlgp}

\begin{figure*}
\centering
\footnotesize
\setlength{\tabcolsep}{0.0cm}
{\renewcommand{\arraystretch}{0.2}
    \begin{tabular}{m{1cm}m{2.2cm}m{2.2cm}m{2.2cm}m{2.2cm}m{2.2cm}!{\textcolor{red}{\vrule width 1pt}}m{2.2cm}m{2.2cm}}

\rotatebox{90}{Ground Truth} &
\includegraphics[width=2.2cm,trim={3cm 3cm 3cm 3cm},clip]{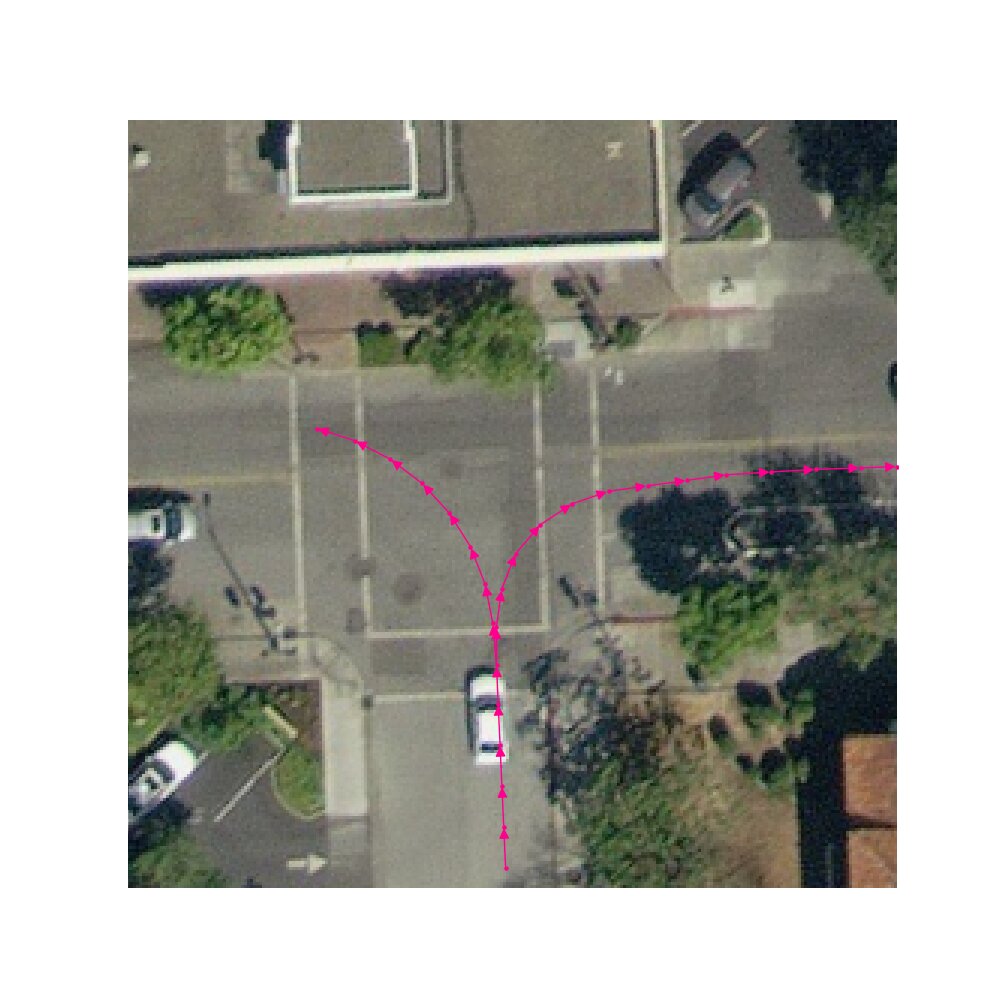} &
\includegraphics[width=2.2cm,trim={3cm 3cm 3cm 3cm},clip]{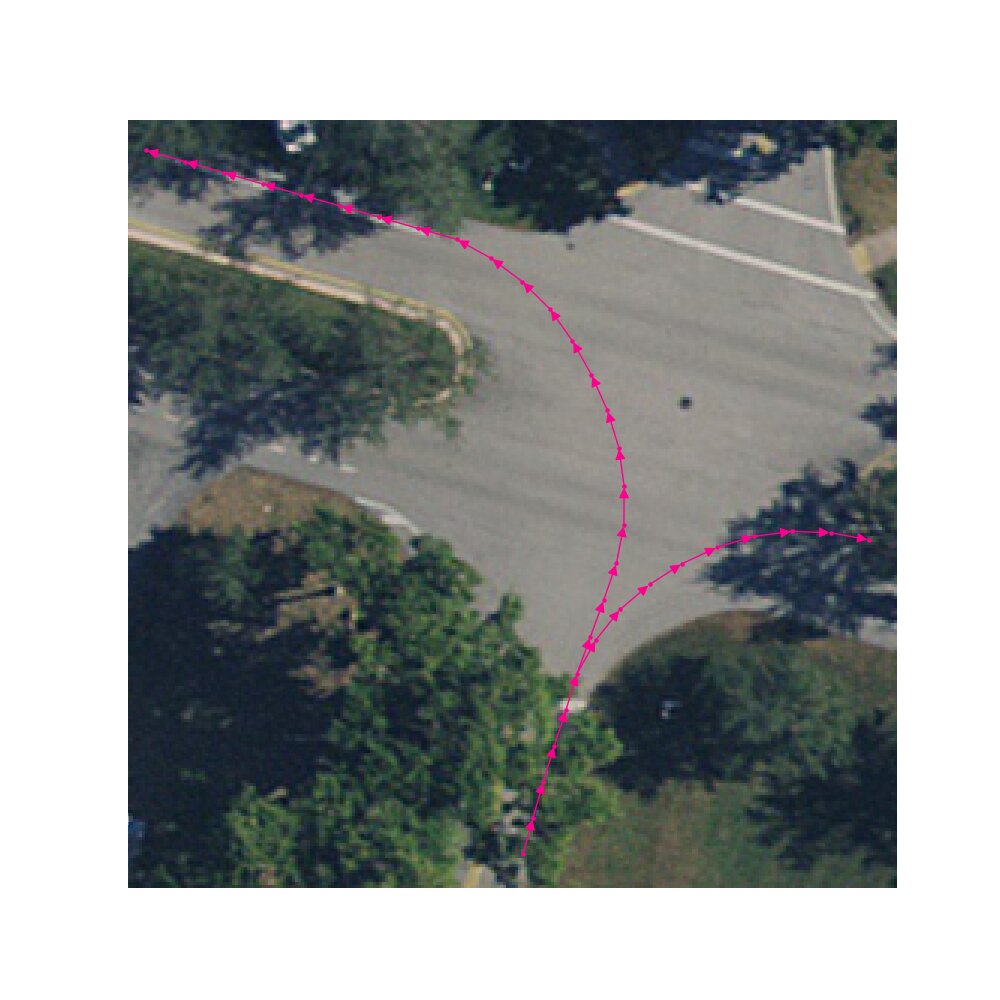} &
\includegraphics[width=2.2cm,trim={3cm 3cm 3cm 3cm},clip]{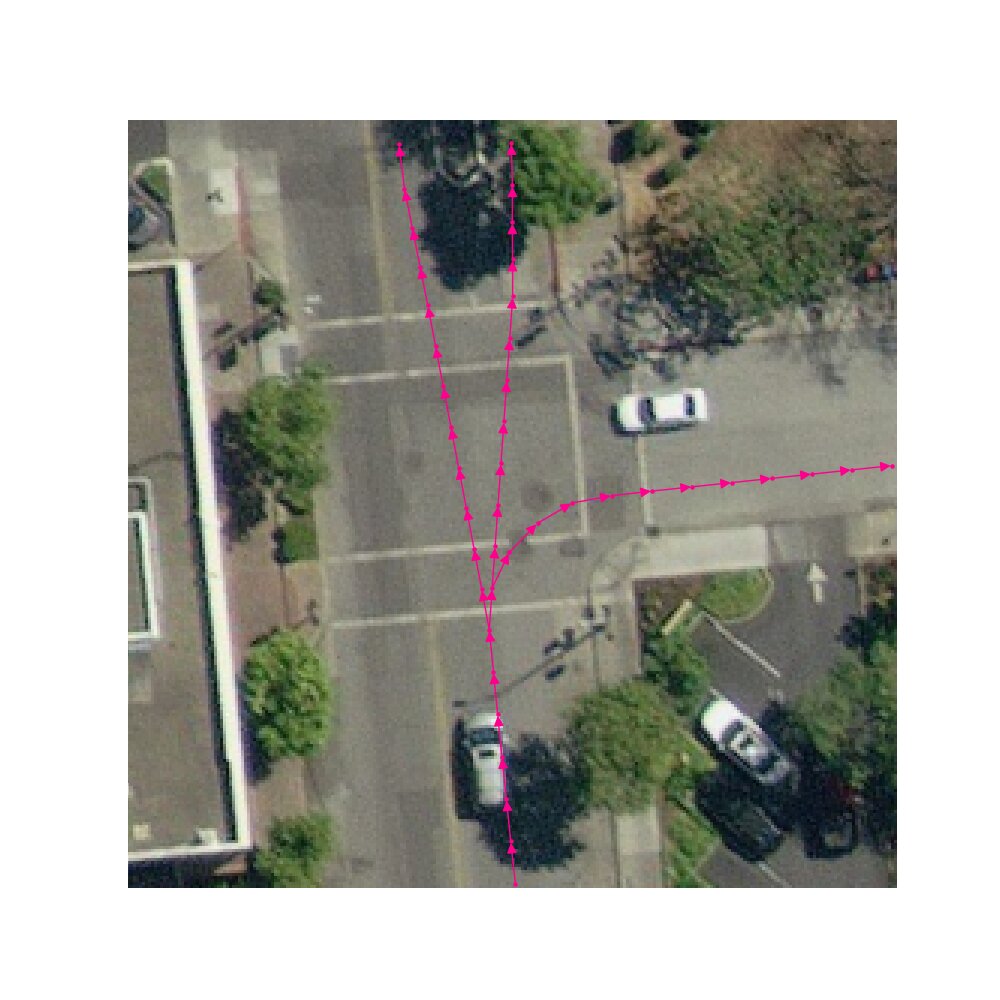} &
\includegraphics[width=2.2cm,trim={3cm 3cm 3cm 3cm},clip]{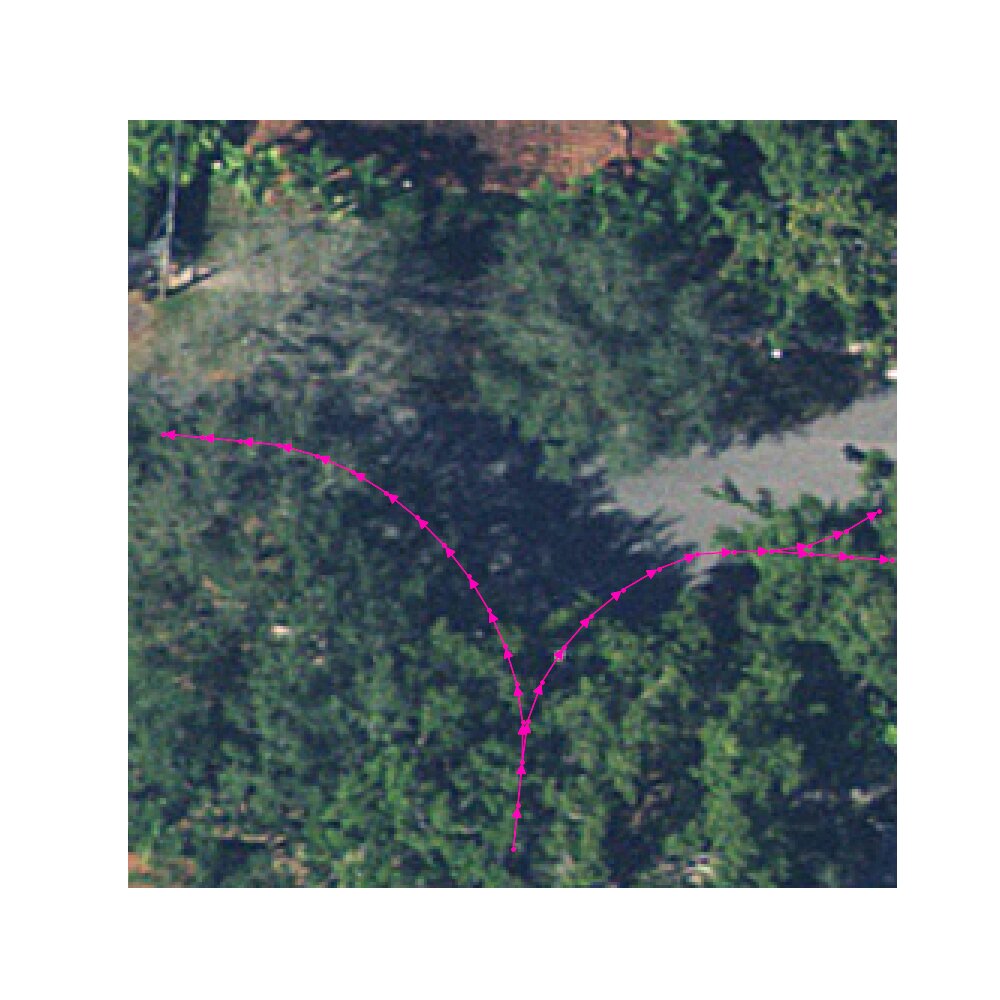} &
\includegraphics[width=2.2cm,trim={3cm 3cm 3cm 3cm},clip]{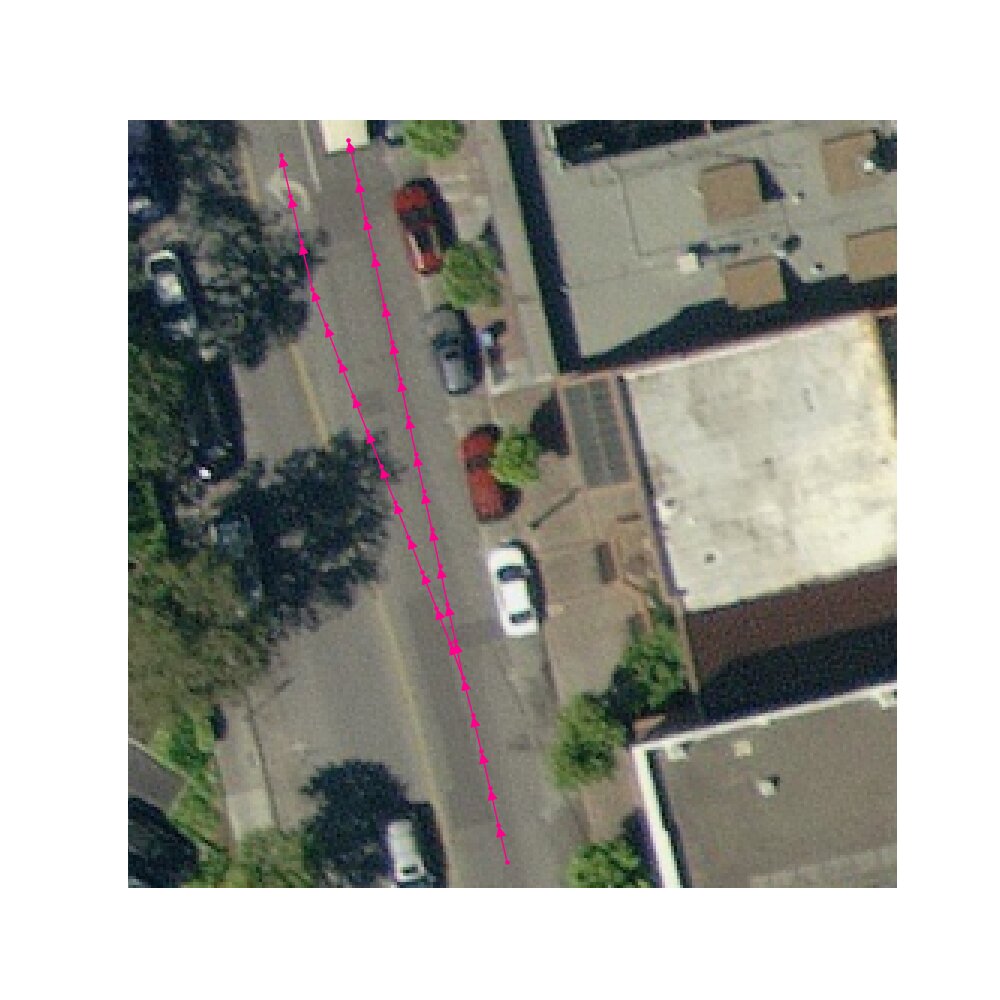} &
\includegraphics[width=2.2cm,trim={3cm 3cm 3cm 3cm},clip]{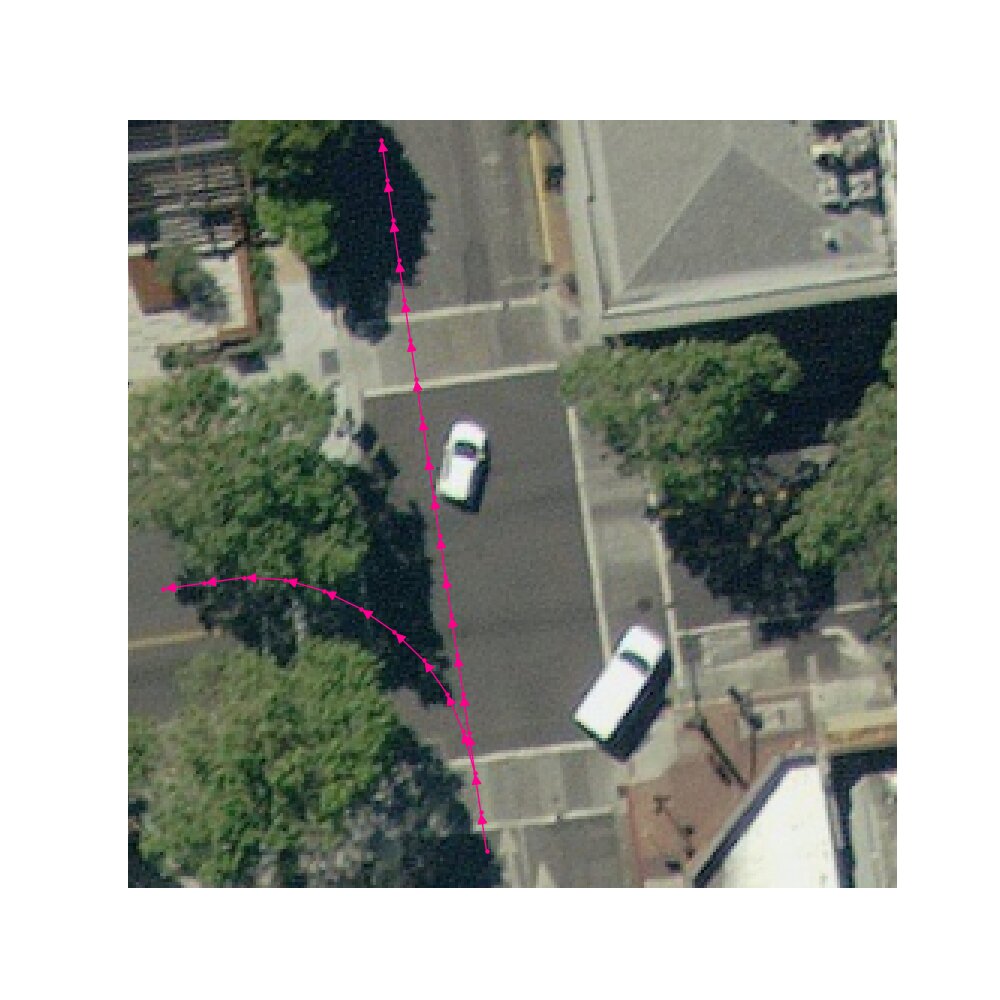} &
\includegraphics[width=2.2cm,trim={3cm 3cm 3cm 3cm},clip]{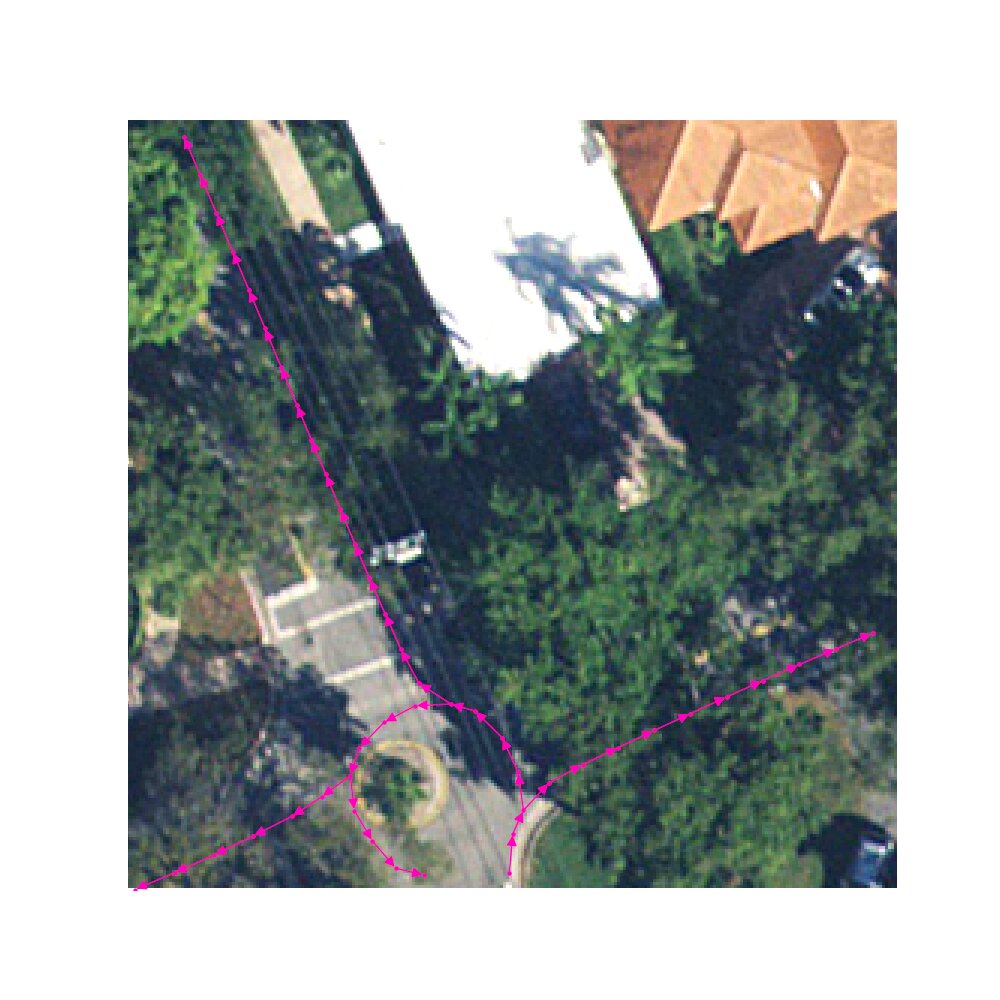} \\  

\rotatebox{90}{LaneGraphNet~\cite{zurn2021lane}}  &
\includegraphics[width=2.2cm,trim={3cm 3cm 3cm 3cm},clip]{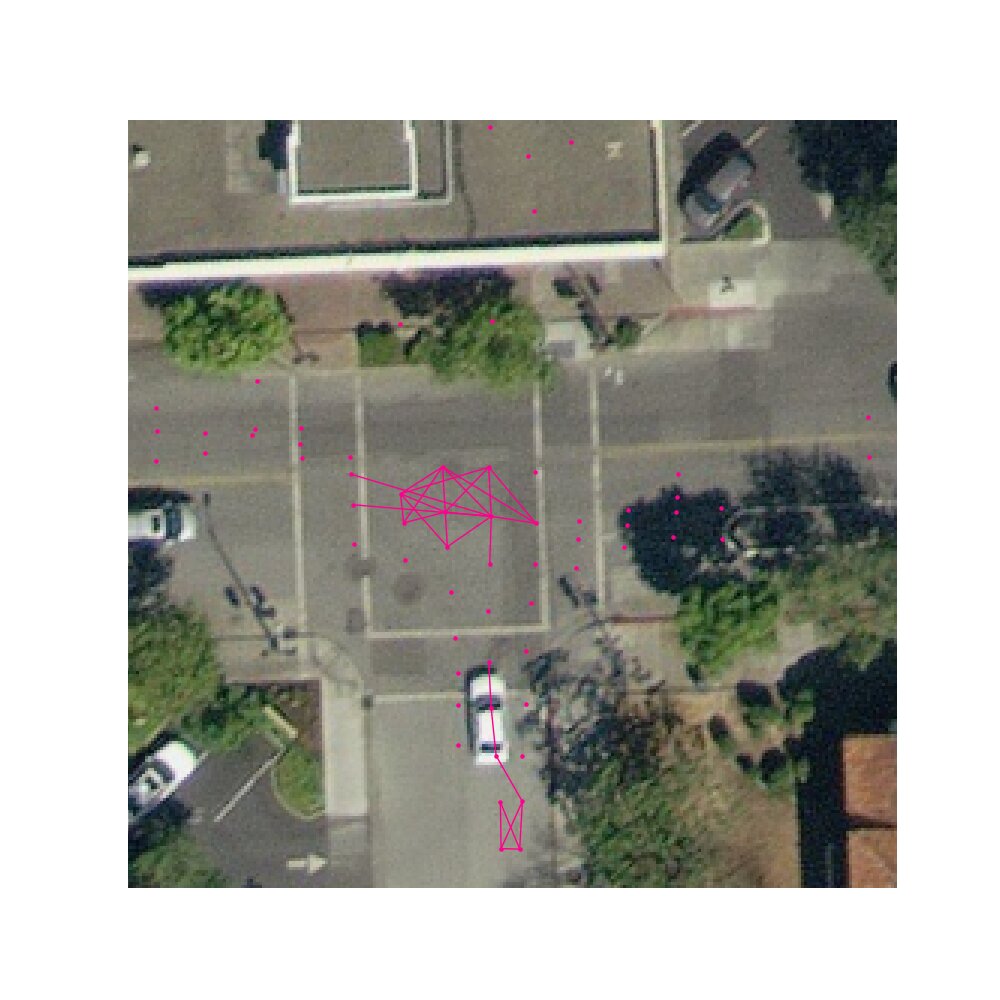} &
\includegraphics[width=2.2cm,trim={3cm 3cm 3cm 3cm},clip]{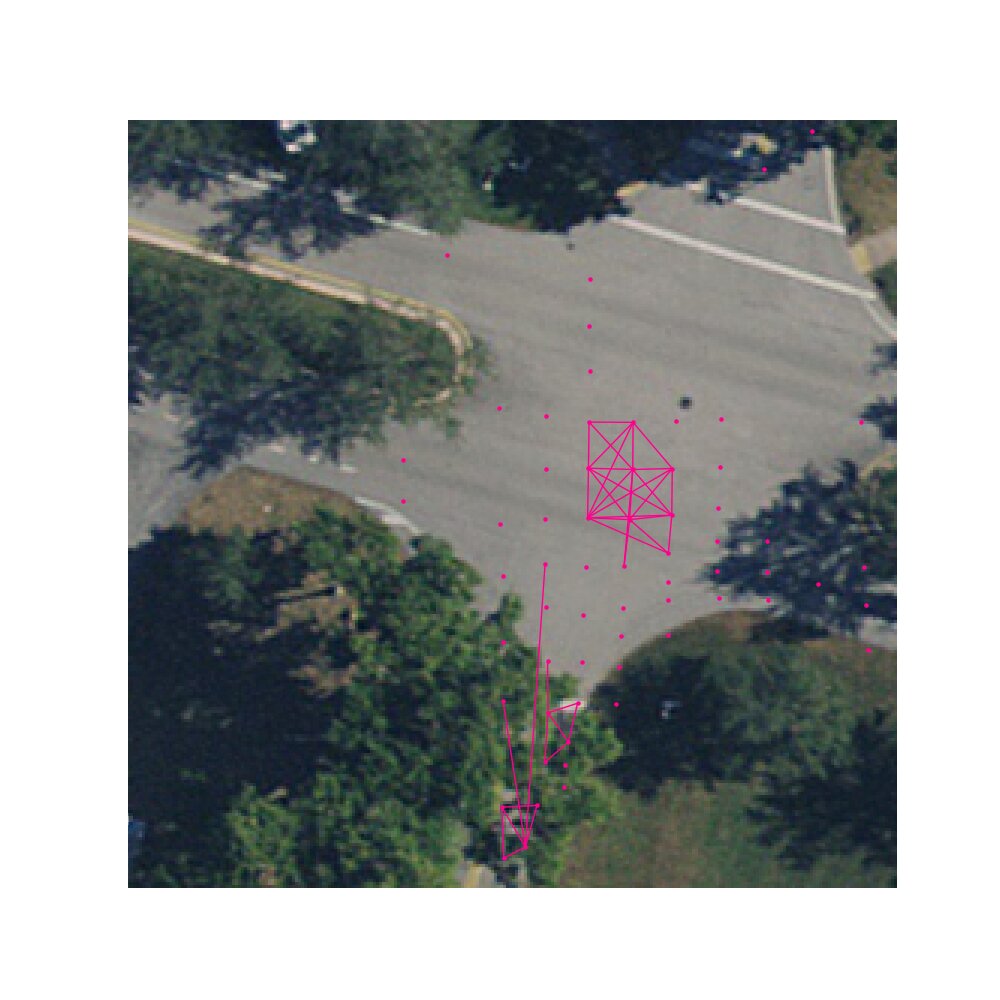} &
\includegraphics[width=2.2cm,trim={3cm 3cm 3cm 3cm},clip]{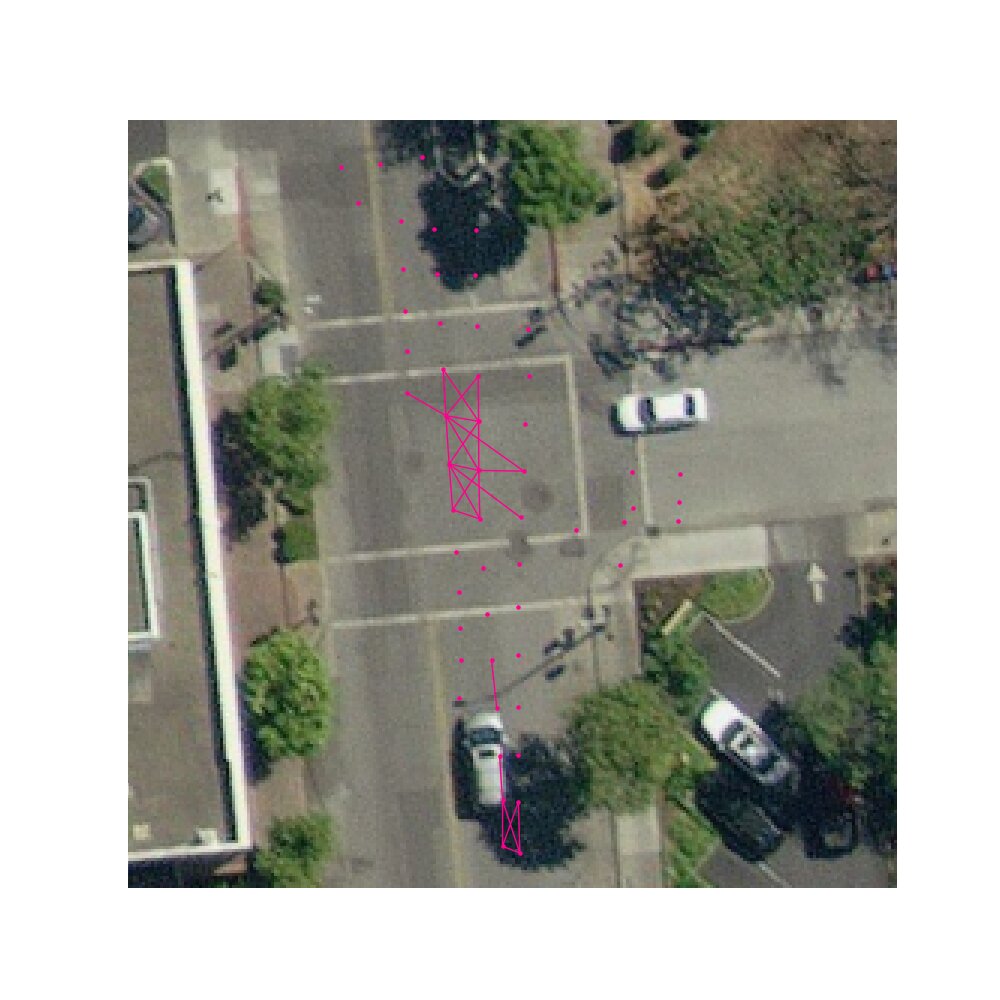} &
\includegraphics[width=2.2cm,trim={3cm 3cm 3cm 3cm},clip]{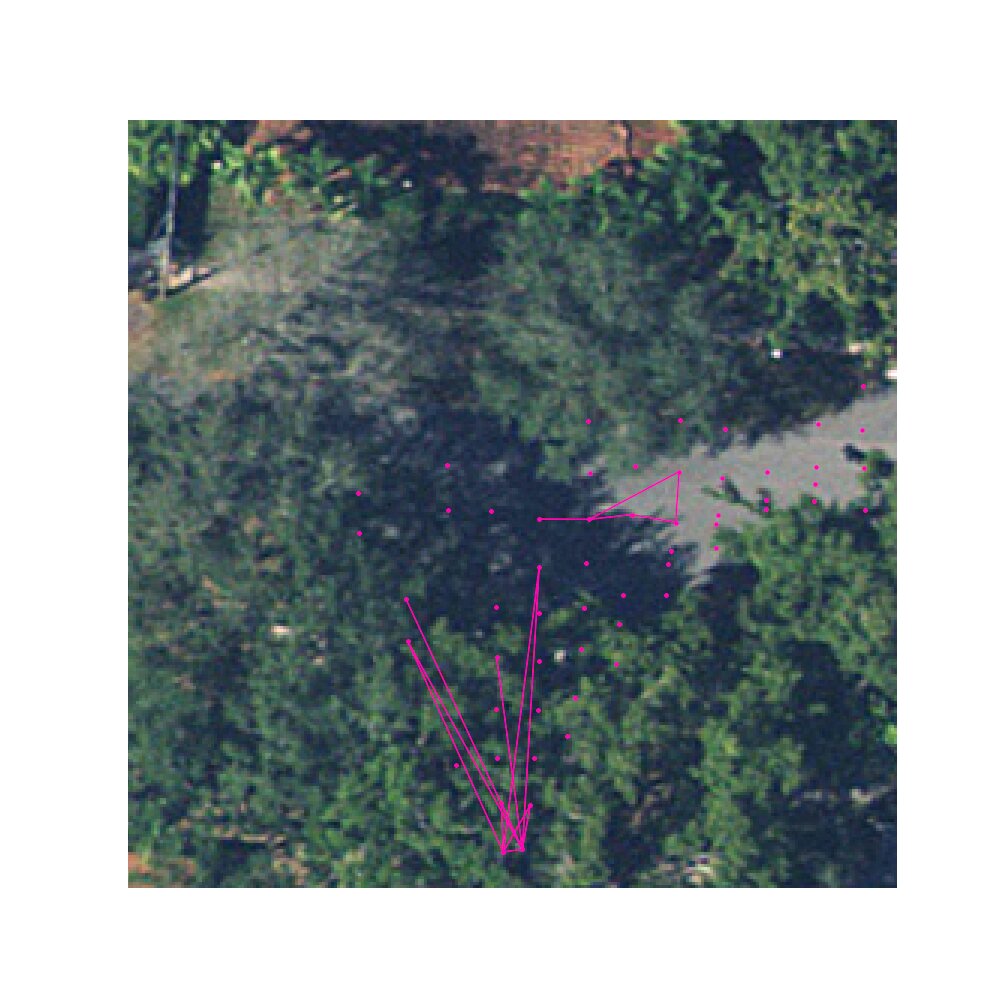} &
\includegraphics[width=2.2cm,trim={3cm 3cm 3cm 3cm},clip]{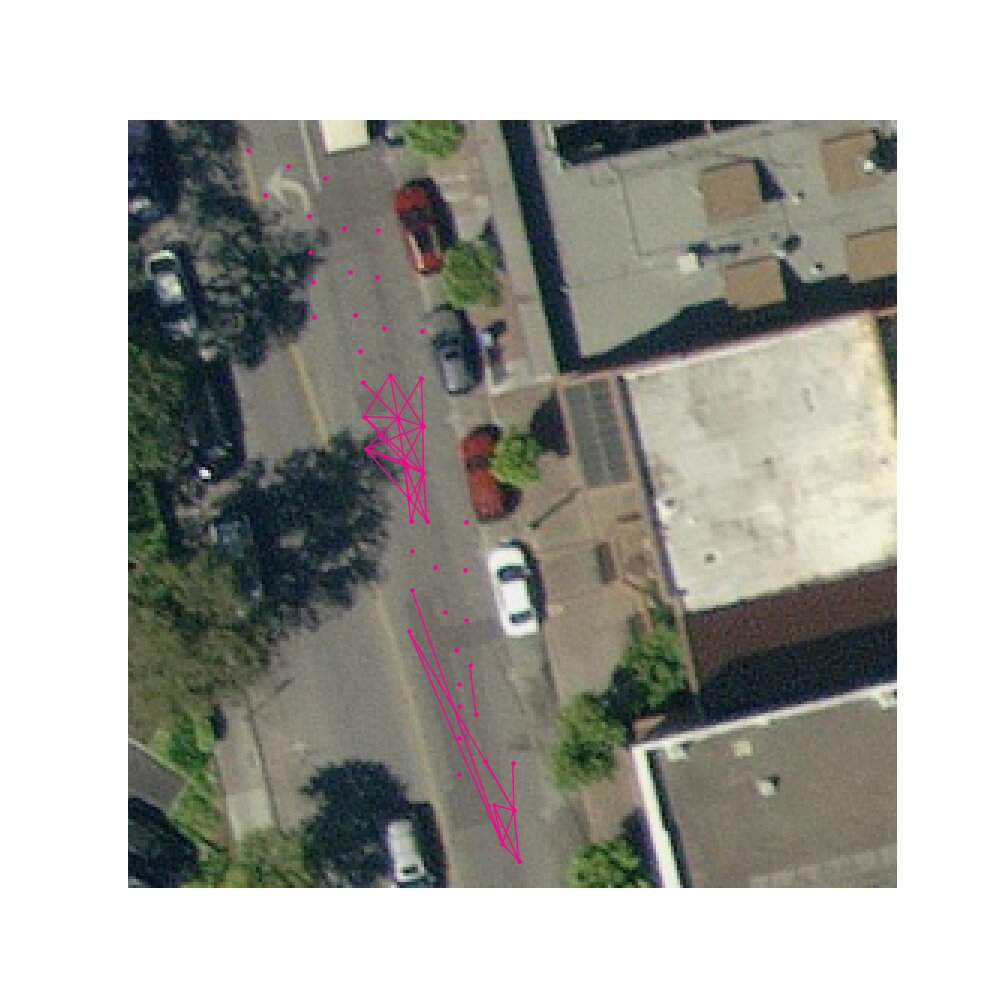} &
\includegraphics[width=2.2cm,trim={3cm 3cm 3cm 3cm},clip]{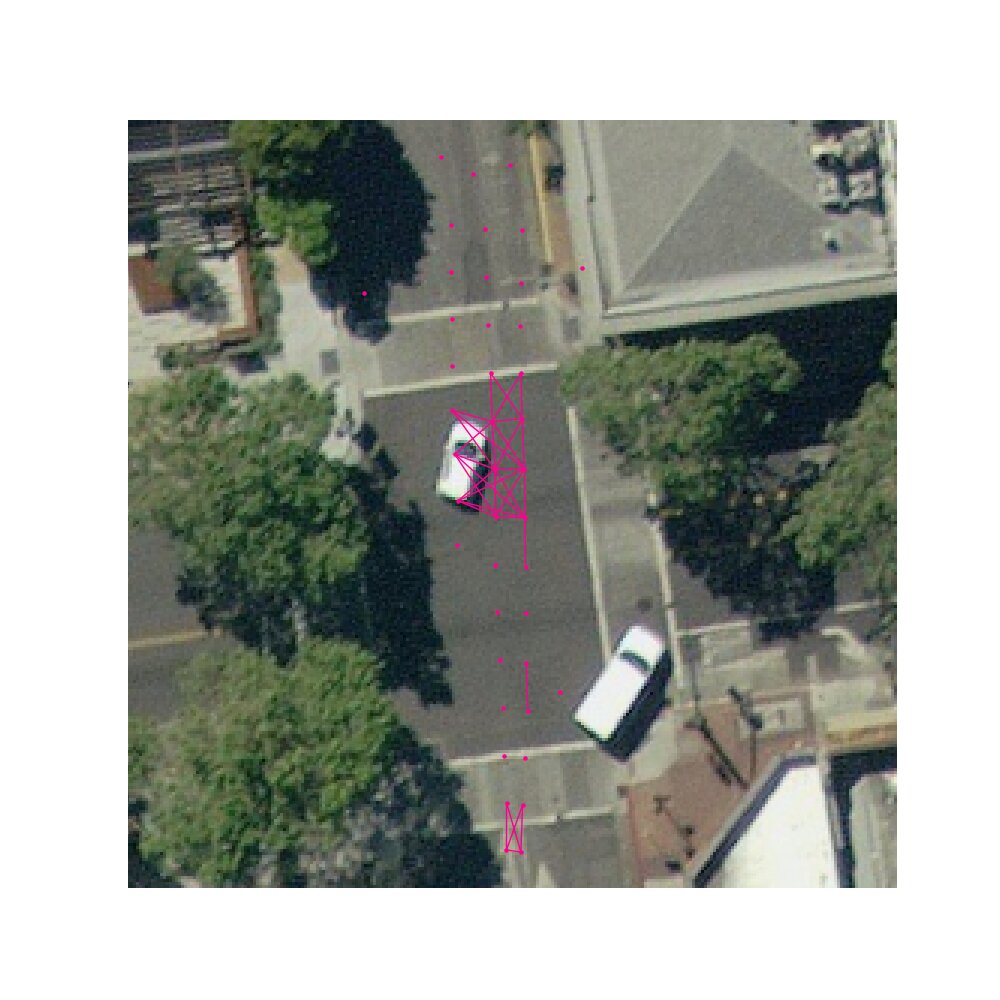} &
\includegraphics[width=2.2cm,trim={3cm 3cm 3cm 3cm},clip]{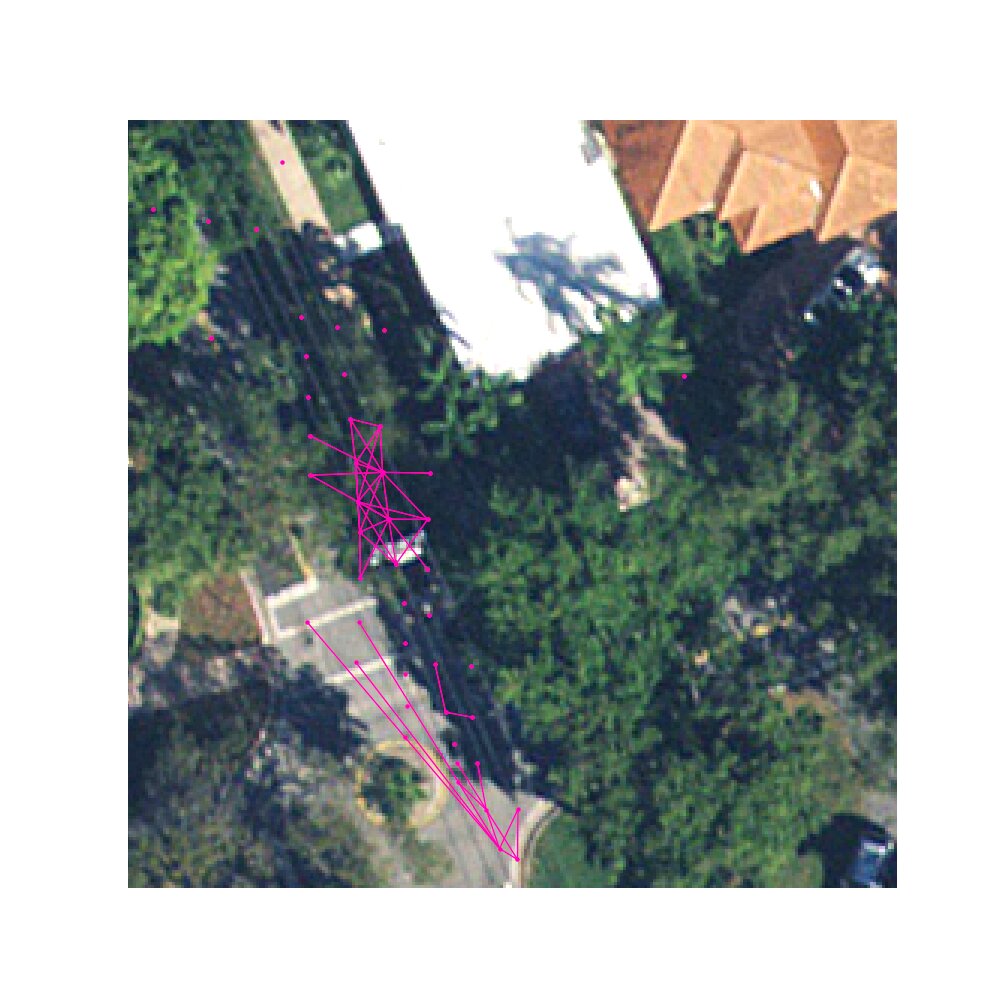} \\  

\rotatebox{90}{\thead{Skeletonized \\ Regression}}    &
\includegraphics[width=2.2cm,trim={0.8cm 0.8cm 0.8cm 0.8cm},clip]{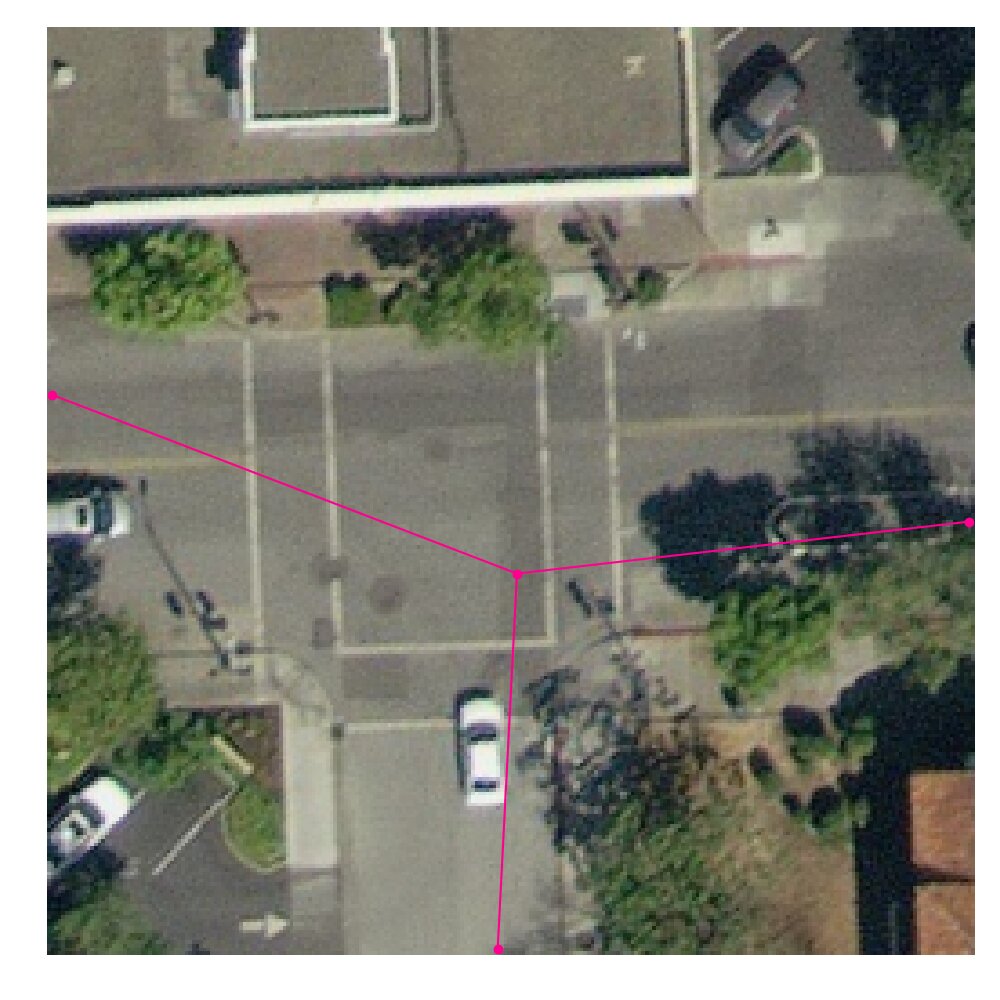} &
\includegraphics[width=2.2cm,trim={3cm 3cm 3cm 3cm},clip]{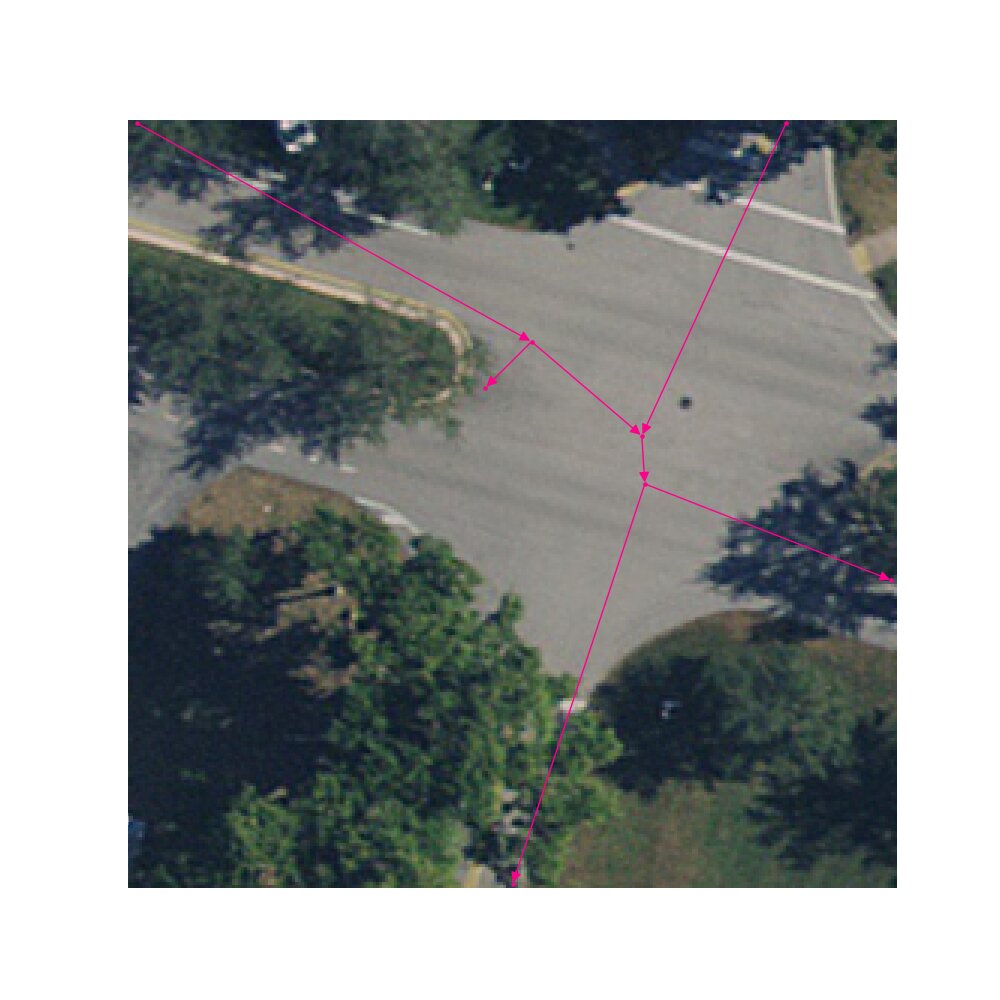} &
\includegraphics[width=2.2cm,trim={0.8cm 0.8cm 0.8cm 0.8cm},clip]{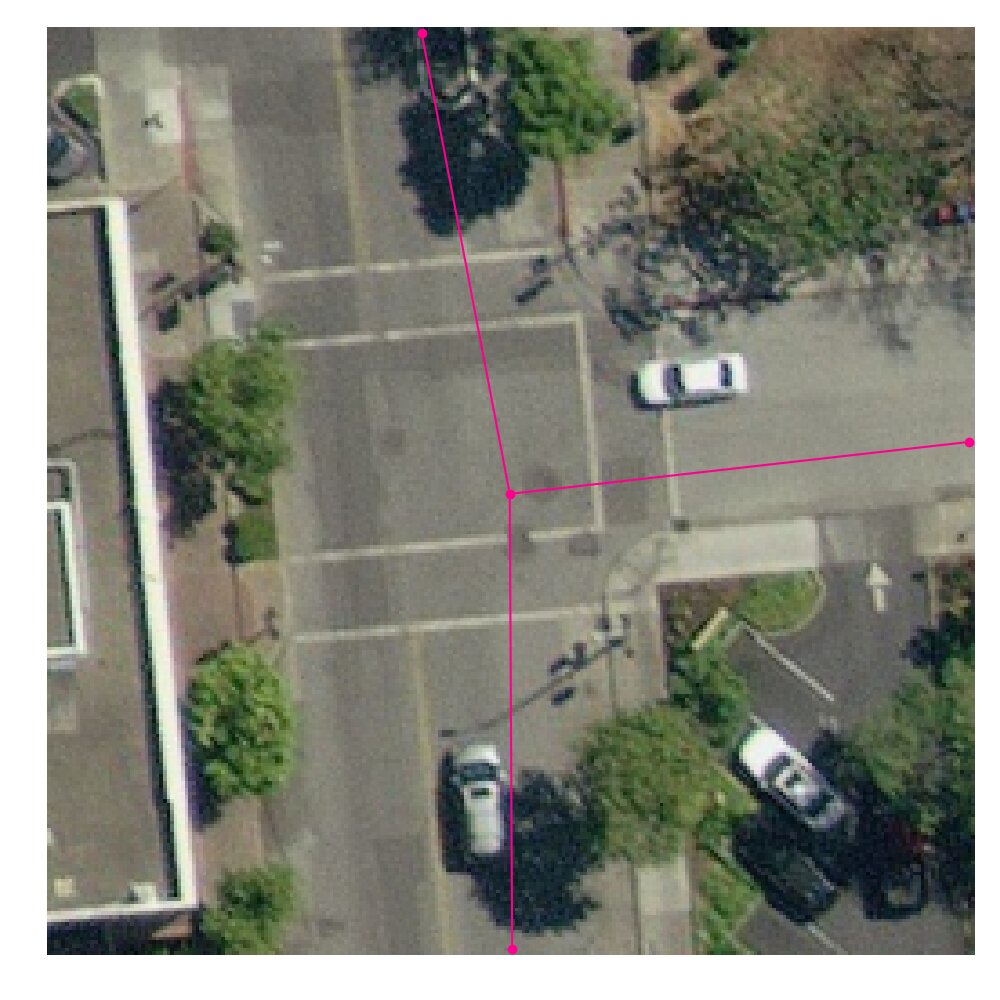} &
\includegraphics[width=2.2cm,trim={0.8cm 0.8cm 0.8cm 0.8cm},clip]{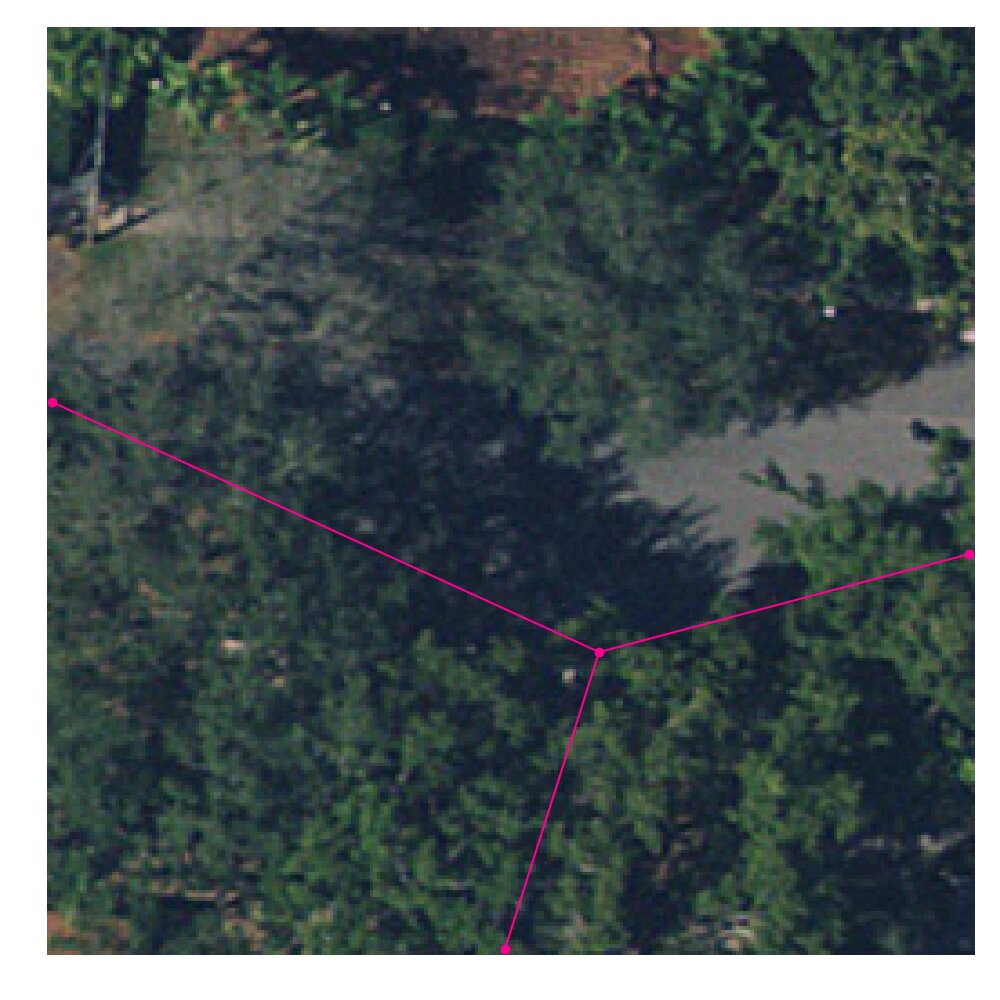} &
\includegraphics[width=2.2cm,trim={0.8cm 0.8cm 0.8cm 0.8cm},clip]{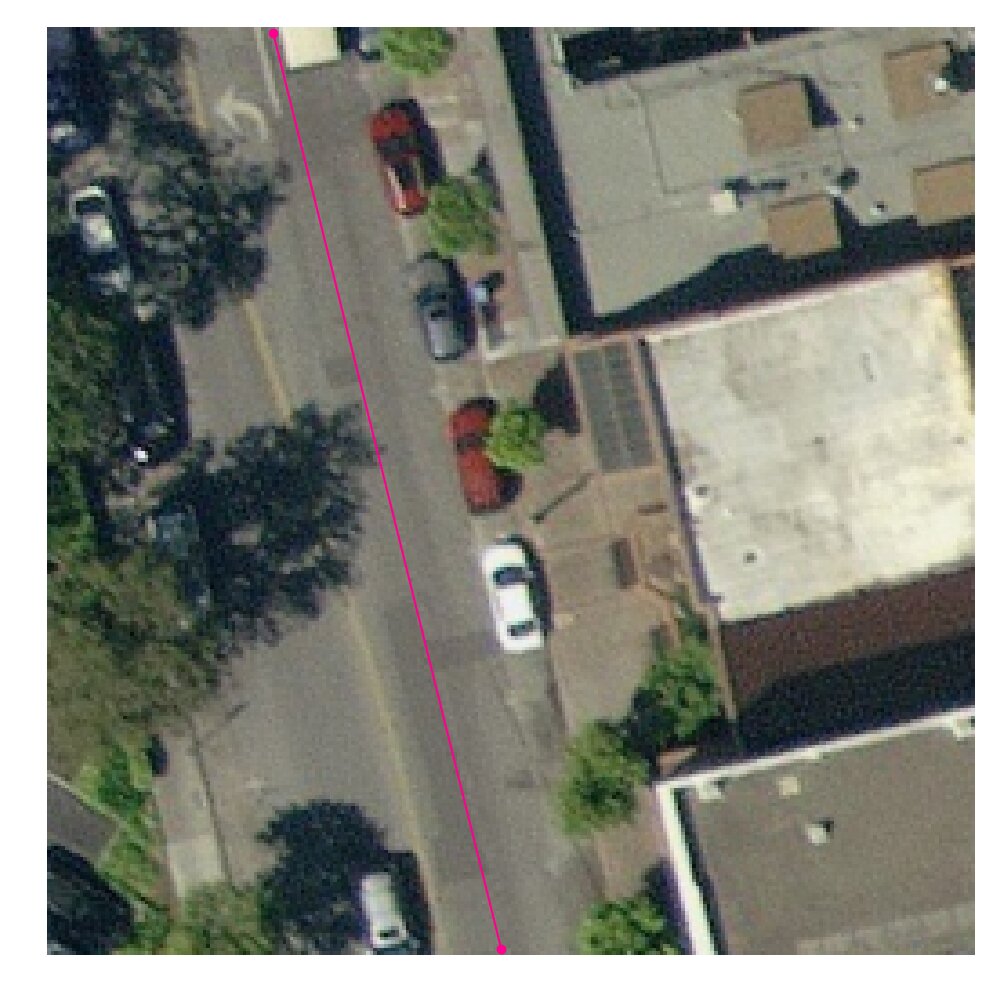} &
\includegraphics[width=2.2cm,trim={3cm 3cm 3cm 3cm},clip]{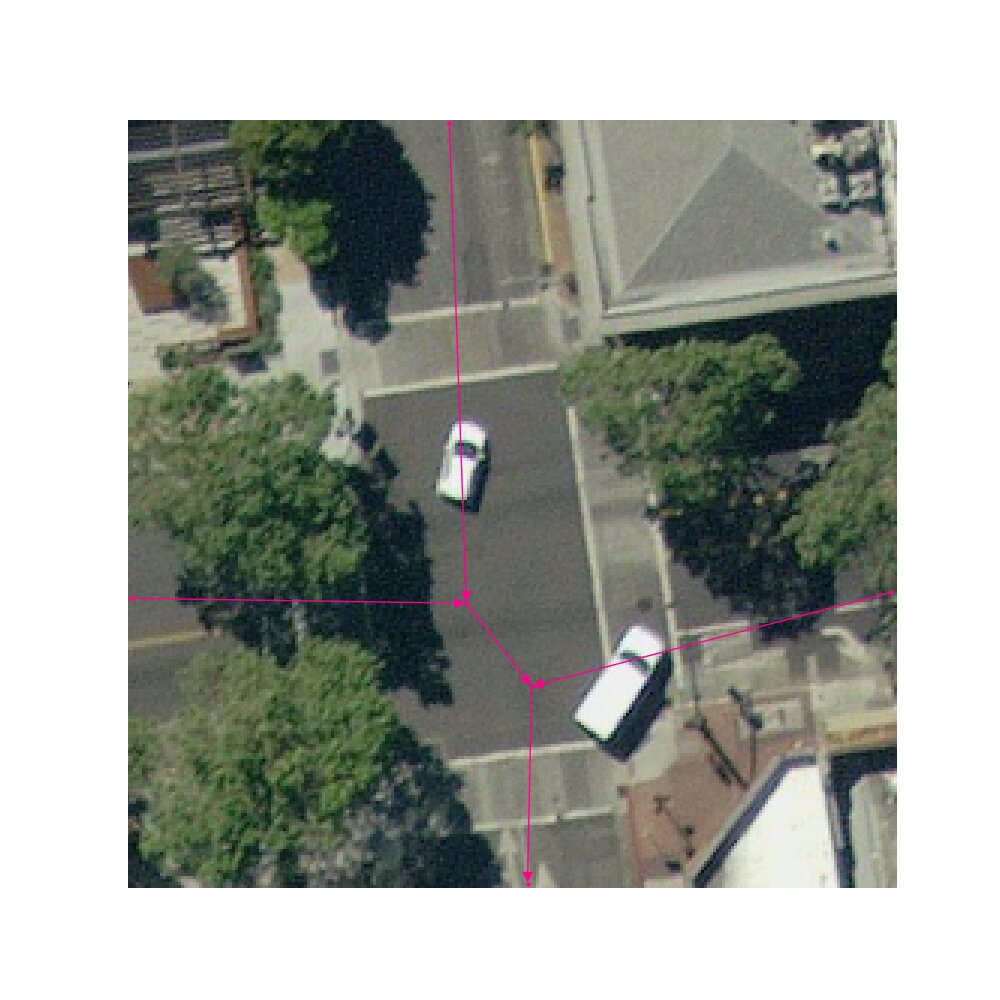} &
\includegraphics[width=2.2cm,trim={0.8cm 0.8cm 0.8cm 0.8cm},clip]{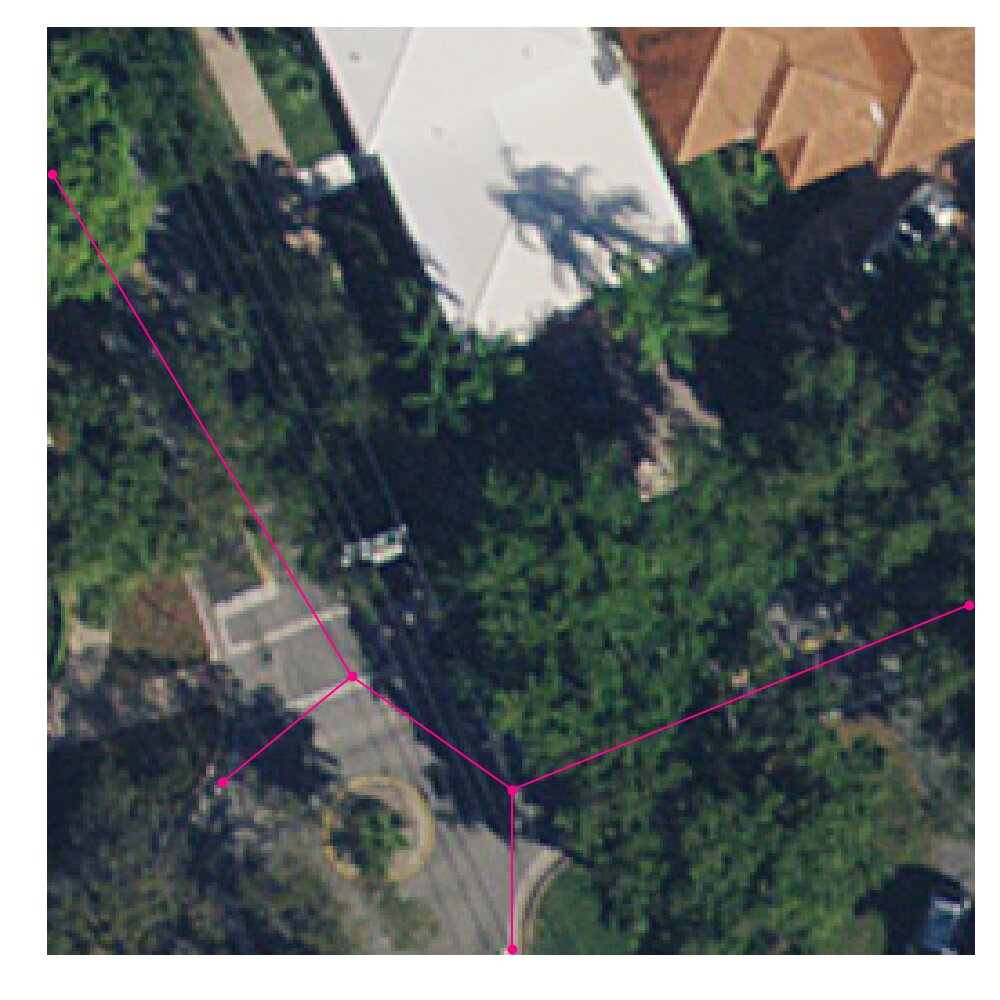} \\  

\rotatebox[origin=lB]{90}{LaneGNN (ours)}  &
\includegraphics[width=2.2cm,trim={3cm 3cm 3cm 3cm},clip]{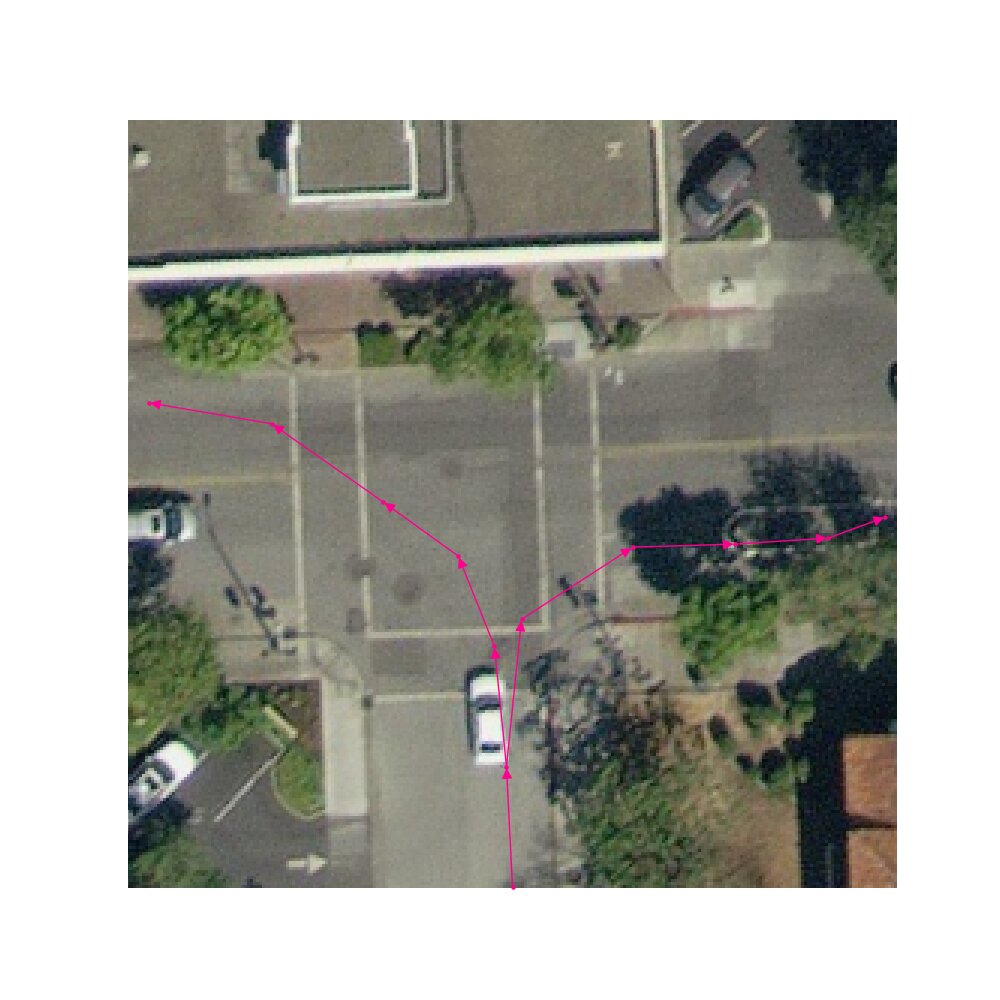} &
\includegraphics[width=2.2cm,trim={3cm 3cm 3cm 3cm},clip]{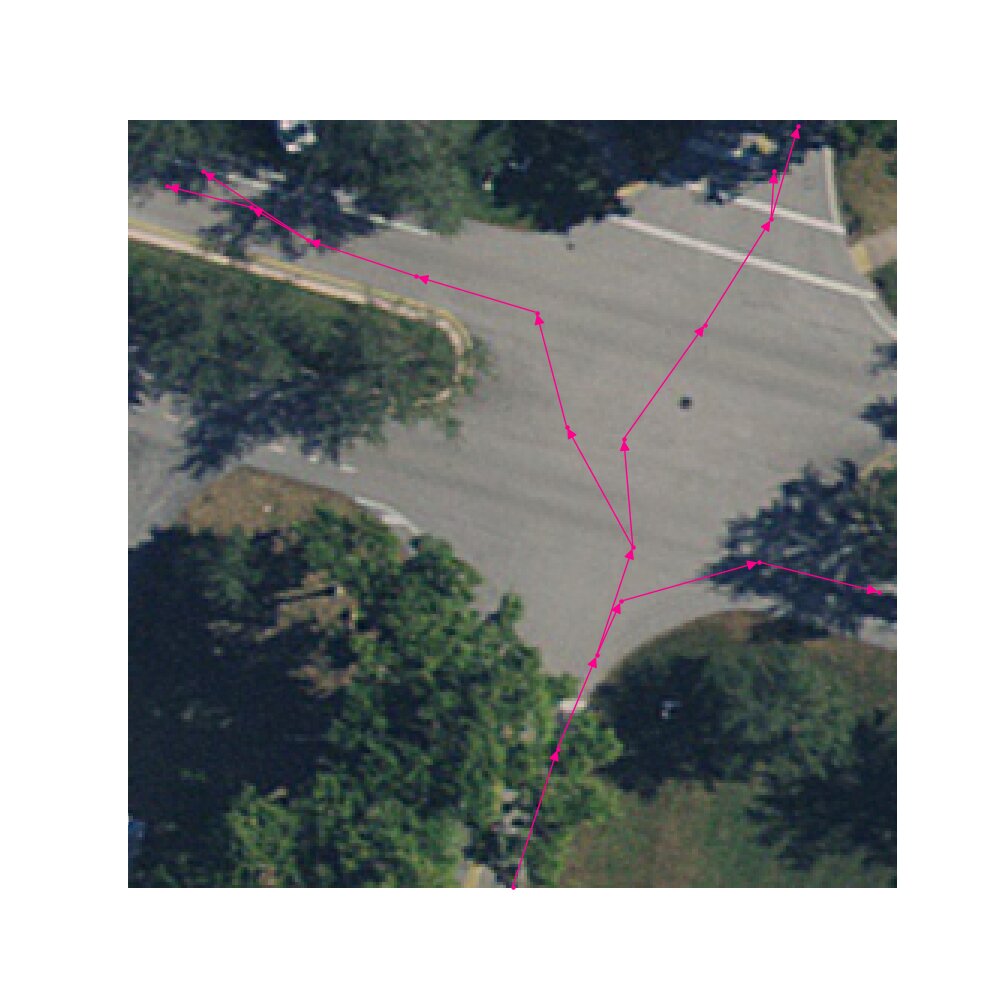} &
\includegraphics[width=2.2cm,trim={3cm 3cm 3cm 3cm},clip]{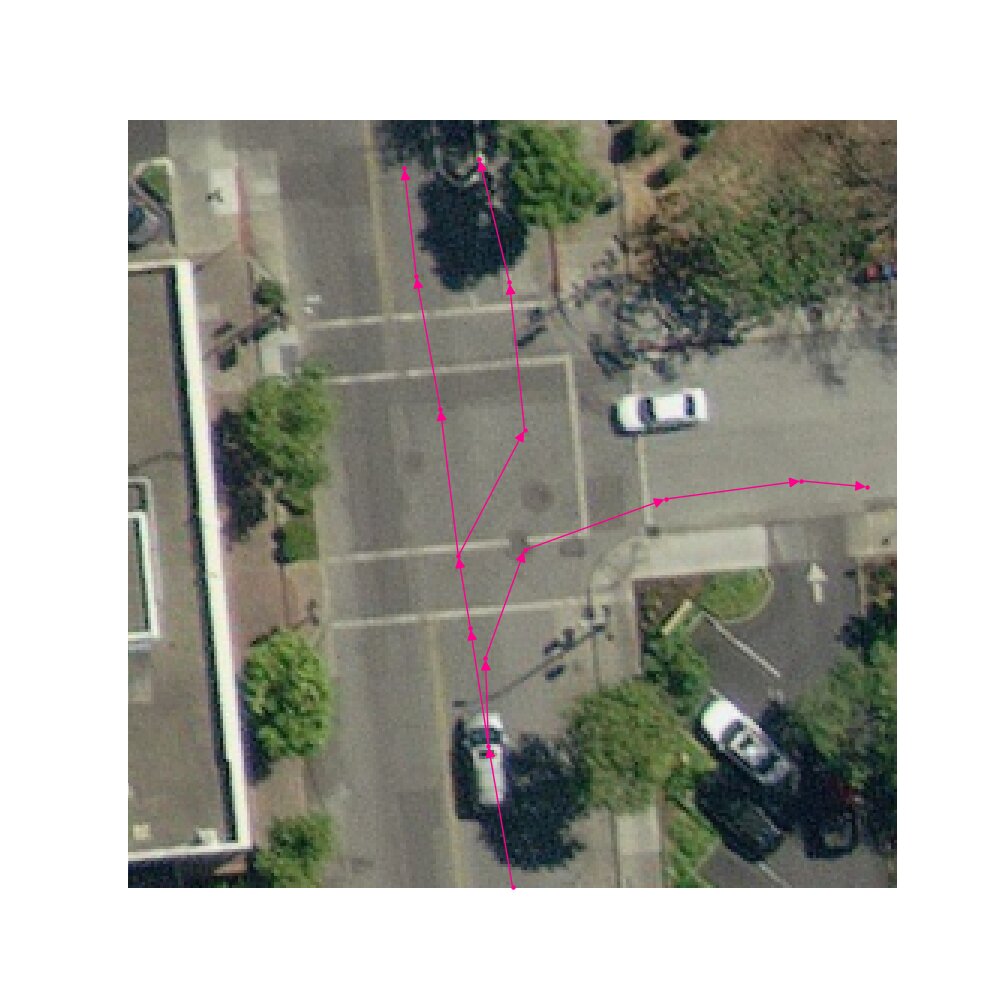} &
\includegraphics[width=2.2cm,trim={3cm 3cm 3cm 3cm},clip]{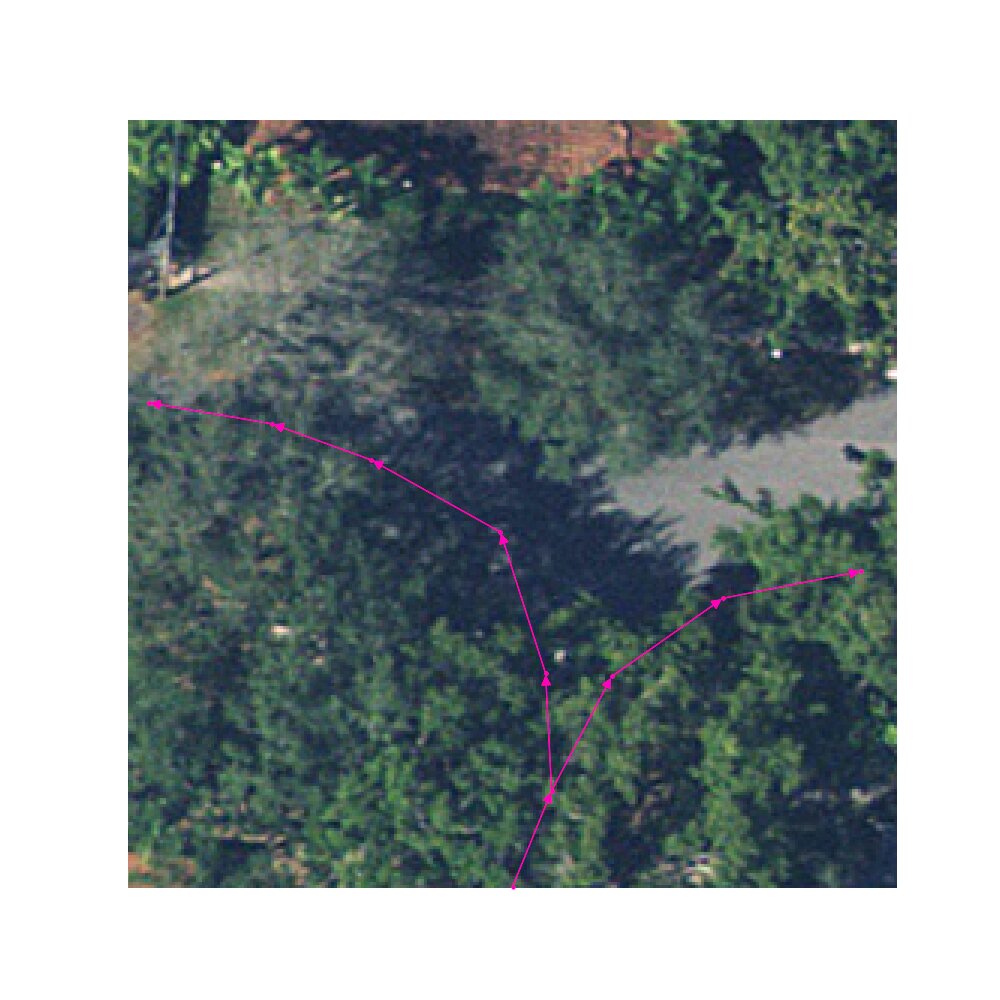} &
\includegraphics[width=2.2cm,trim={3cm 3cm 3cm 3cm},clip]{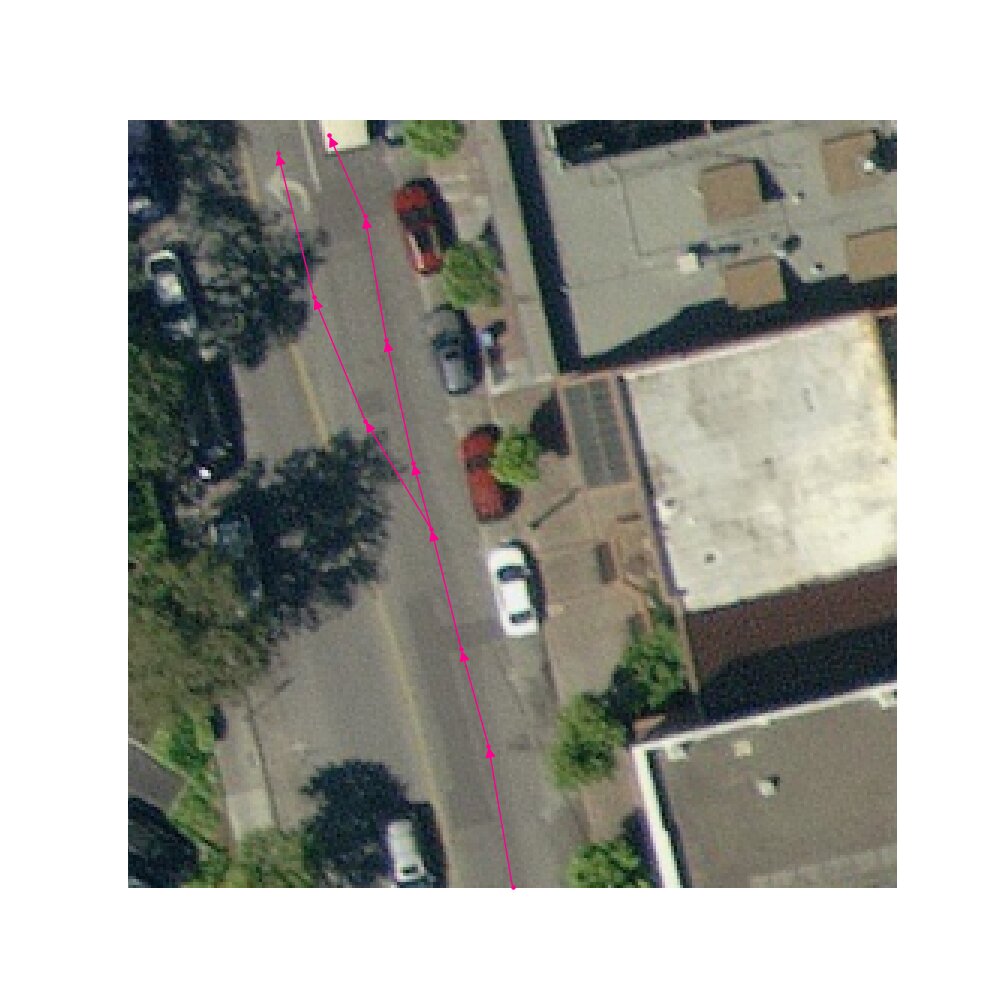} &
\includegraphics[width=2.2cm,trim={3cm 3cm 3cm 3cm},clip]{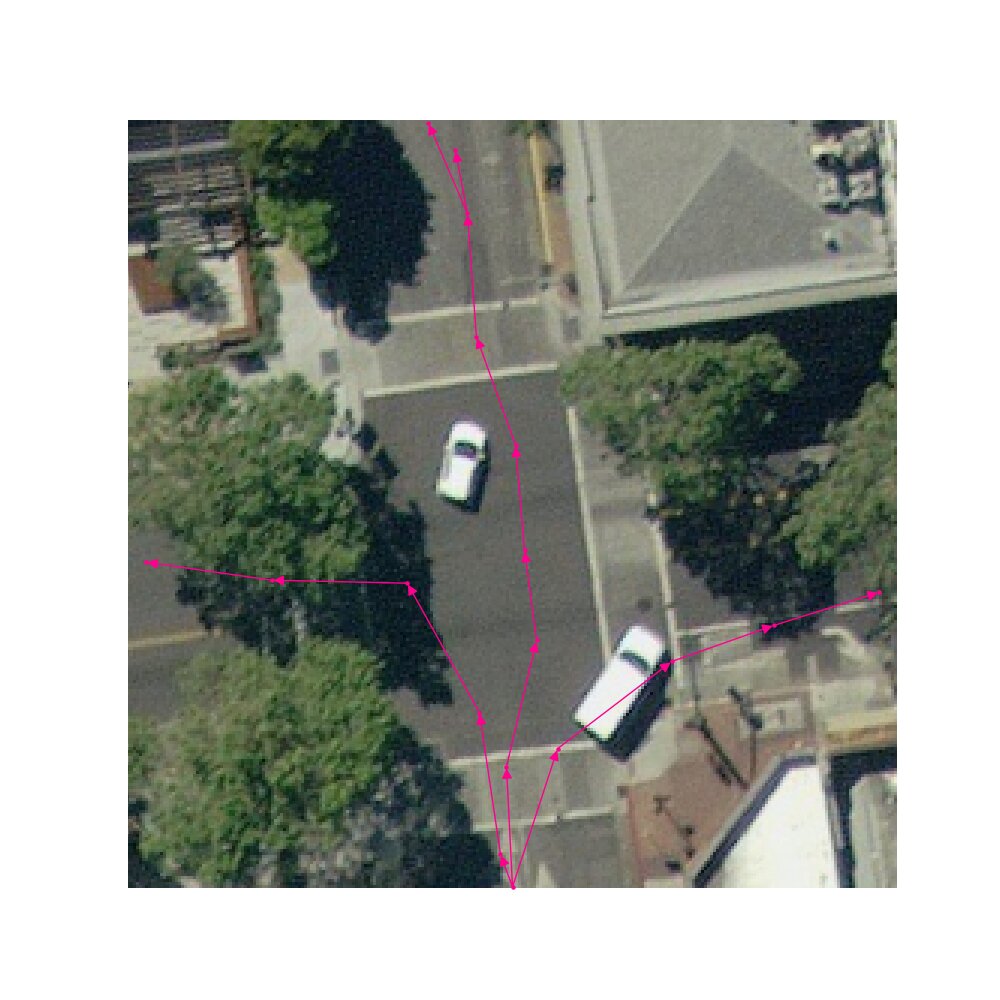} &
\includegraphics[width=2.2cm,trim={3cm 3cm 3cm 3cm},clip]{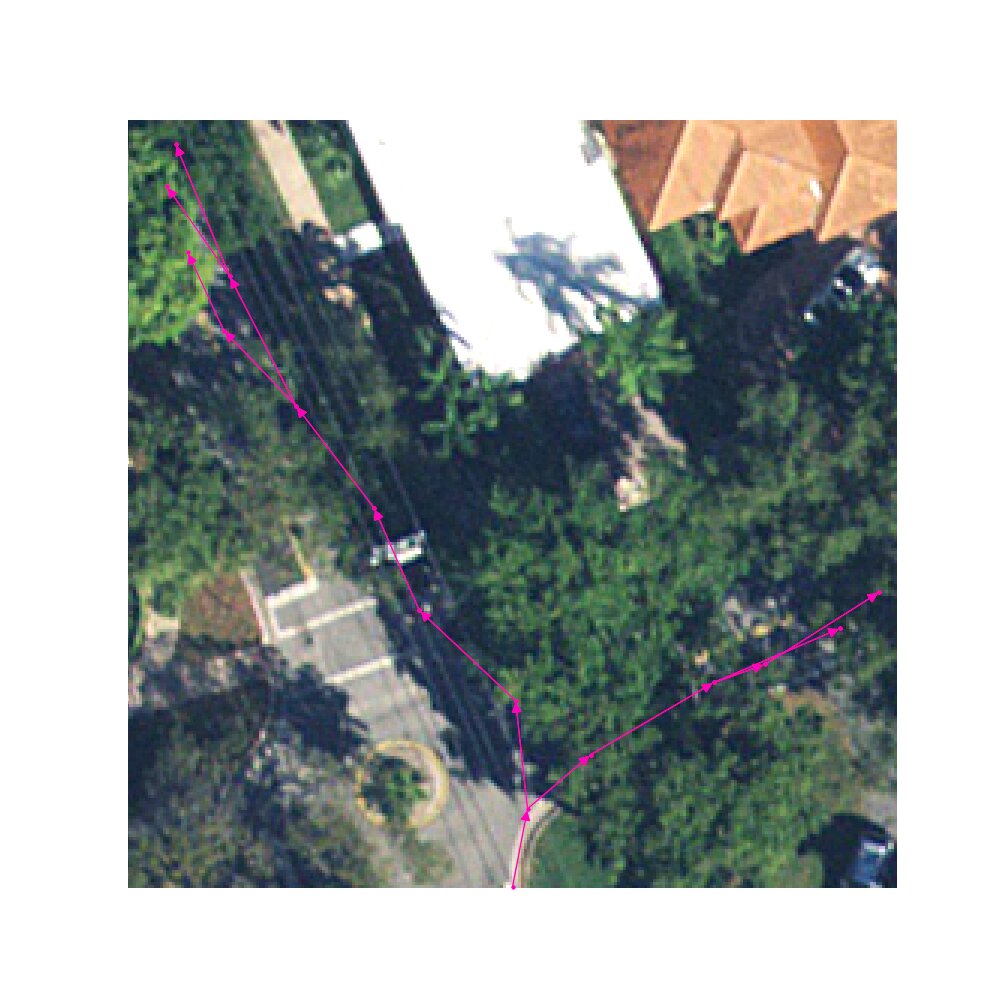} \\

    \end{tabular}
    }
    \caption{Qualitative results on the Successor-LGP task. We compare predictions of our model with LaneGraphNet \cite{zurn2021lane} and a morphological image skeletonization baseline. Predicted nodes are visualized with points while predicted edges are visualized as directed arrows. We illustrate failure cases in the two rightmost columns. Best viewed zoomed in.}
    \label{fig:qualitative-successor} 
\end{figure*}

In the following, we evaluate LaneGNN by ablating and comparing it with two baselines on the Successor-LGP task: morphological skeletonization of the ego-lane regression as well as a modified LaneGraphNet~\cite{zurn2021lane} to be used for successor lane graph prediction. We list quantitative results in Tab.~\ref{tab:results-successor}.
Our results demonstrate that none of the baseline methods are capable of estimating accurate lane graphs given the challenging topology of the lane graphs in the dataset. The LaneGraphNet\cite{zurn2021lane} model fails to model the graph for many samples, yielding low scores in all metrics. Despite its simplicity, the skeletonized regression model achieves the highest APLS score and comparably high GEO/TOPO scores. However, it fails to accurately predict lane split points, resulting in low SDA scores. Increasing the number of nodes of the skeleton leads to many more false positive splits and thus deteriorates further. 

Regarding different variants of LaneGNN, we find that, e.g., causal message passing (CMP) increases performance over standard message passing, which underlines the significance of the imposed causality prior for lane graph learning. In order to show the efficacy of computationally more demanding aerial image edge features (AerE) compared to uni-directional aerial image node features (AerN), we ablate on this in Tab.~\ref{tab:results-successor} as well. We observe stark increases across the TOPO, GEO and SDA metrics when utilizing aerial edge features. Lastly, we replace the ego-lane segmentation mask $\mathbf{S}_{lane}^{ego}$ used for sampling with the standard lane segmentation mask. Our findings show that especially APLS, SDA, and Graph IoU drastically decrease as graph estimation becomes more difficult due to a generally enlarged sampling region (see Fig.~\ref{fig:approach-model}).

These results are further illustrated in Fig.~\ref{fig:qualitative-successor}, where we show qualitative comparisons of predictions of our best-performing model with predictions from the two baselines. We find that the quality of predictions by LaneGraphNet~\cite{zurn2021lane} is generally low, rendering it unsuitable for the task. The skeletonized regression baseline is capable of following basic lane graph topologies, but lane split points cannot be resolved accurately. In contrast, our LaneGNN model is capable of modeling most graphs with high accuracy; both in intersection areas and in straight road sections. For more results, please refer to the supplementary material, Sec.~\ref{supp:successor-lgp}.

\subsection{Full Lane Graph Prediction}
\label{sec:experiments-fulllgp}
\begin{table}
\centering
\scriptsize
\setlength\tabcolsep{3.7pt}
\caption{Quantitative evaluation for the Full-LGP task on the test-set of our \textit{UrbanLaneGraph} dataset. We compare a baseline model with graphs aggregated with a naïve aggregation scheme and our iterative temporal aggregation scheme. P/R denotes Precision/Recall. Higher values mean better results.}
\begin{tabular}{p{2.1cm}|p{1.3cm} p{1.2cm} p{0.9cm} p{1.3cm}}
 \toprule
Model &    TOPO P/R$\,\uparrow$ &  GEO P/R$\,\uparrow$ & APLS$\,\uparrow$ & Graph IoU$\,\uparrow$ \\
 \midrule
LaneExtraction \cite{he2022lane}   &  0.405/0.507 & 0.491/0.454  & 0.072 & 0.213 \\
 \midrule 
Aggregation (naïve)               &   0.366/0.654 & 0.523/\textbf{0.727} & 0.101 & 0.376 \\
Aggregation (ours)  & \textbf{0.481}/\textbf{0.670} & \textbf{0.649}/0.689 &  \textbf{0.103} & \textbf{0.384} \\
 \bottomrule
\end{tabular}
\label{tab:results-fulllgp}
\vspace{-0.3cm}
\end{table}

\begin{figure*}
\renewcommand{\arraystretch}{2}
\centering
\footnotesize
\setlength{\tabcolsep}{0.1cm}
    \begin{tabular}{m{0.6cm}m{14cm}}
     \rotatebox{90}{LaneExtraction\cite{he2022lane}} & \includegraphics[width=14cm]{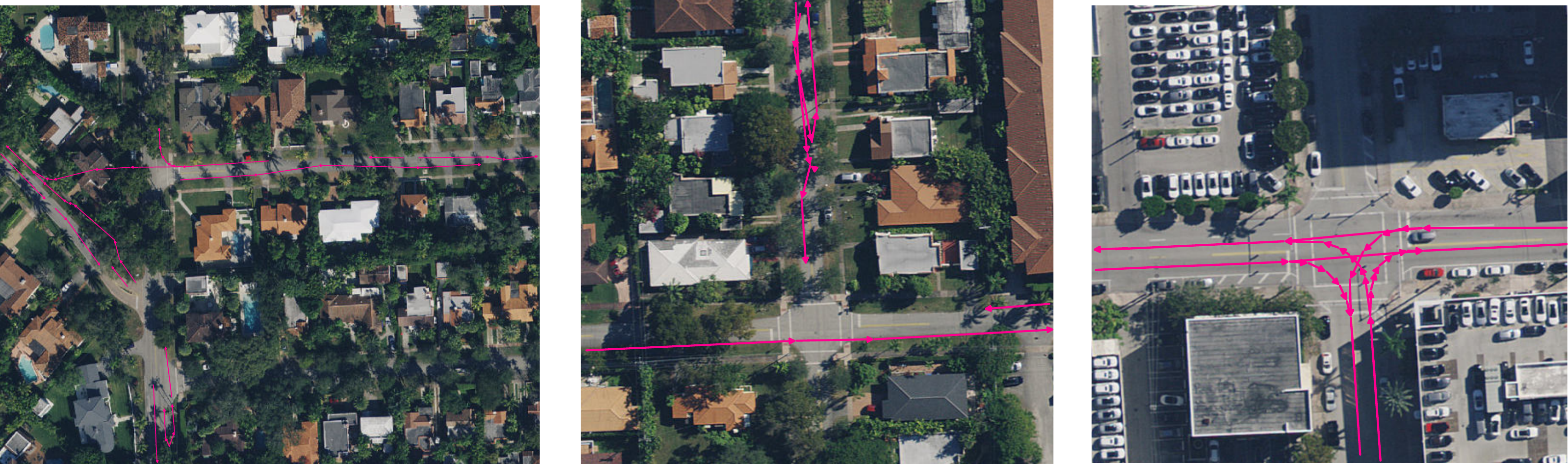}  \\
\rotatebox{90}{Aggregation (ours)} &    \includegraphics[width=14cm]{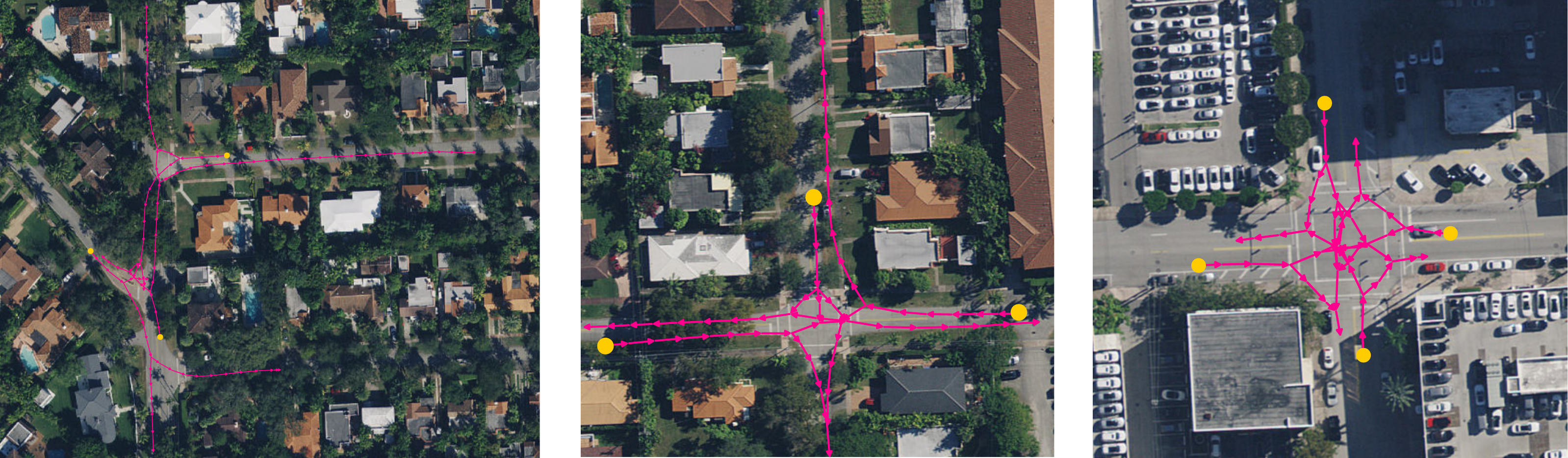}  \\
    \end{tabular}
    \vspace{-0.3cm}
    \caption{Qualitative results on the Full-LGP task. We visualize predictions of LaneExtraction~\cite{he2022lane} (top row) and aggregated LaneGNN predictions (bottom row). Our model is initialized at poses using predicted lane direction masks of LaneExtraction (indicated with yellow circles). Best viewed zoomed in.}
    \label{fig:qualitative-fulllgp} 
\end{figure*}
    


For the Full-LGP task, we compare our approach with LaneExtraction~\cite{he2022lane}. Since their used graph representation is incompatible with ours, we train it on their provided dataset. To allow for a fair comparison, we evaluate both our method and LaneExtraction only on scenes in the city of Miami, Florida, as it is contained in both of the datasets. We select a testing region that is not part of the training data for either of the models. To initialize our aggregation scheme, we select starting poses obtained from intermediate segmentation predictions including yaw angles of the LaneExtraction model. Tab.~\ref{tab:results-fulllgp} lists the evaluation results for the Full-LGP task obtained with LaneExtraction \cite{he2022lane} and the results obtained with our LaneGNN model in conjunction with two aggregation schemes: a na\"ive aggregation scheme baseline and our full aggregation scheme. The na\"ive aggregation scheme merges nodes in close proximity while not relying on unvalidated split/merge or parallel path removal as well as the lateral weighting-based merging (Sec.~\ref{subsec:lgaggregation}). Our experiments show that our aggregation scheme outperforms both the LaneExtraction model and the na\"ive aggregation scheme on nearly all evaluation metrics. We note that our method improves the TOPO/GEO precision metrics while maintaining similar recalls due to better handling of redundant nodes. 
Fig. \ref{fig:qualitative-fulllgp} illustrates successive aggregations from our model while indicating the used initialization points from the LaneExtraction model. Since our aggregation approach does not differentiate between intersection and non-intersection regions, it does not deteriorate in regions that do not exactly fit this categorization. Furthermore, our model exhibits superior performance in reduced visibility settings introduced by stark illumination changes or road occlusions from vegetation, as illustrated in Fig. \ref{fig:qualitative-fulllgp}. One of the decisive assets of our bottom-up method is that it allows to \textit{explore} regions that are missed by LaneExtraction~\cite{he2022lane} as they are entailed in their predicted segmentation masks. For more qualitative and quantitative results, please refer to the supplementary material, Sec.~\ref{supp:full-lgp}. 

\begin{table}
\centering
\scriptsize
\setlength\tabcolsep{3.7pt}
\caption{Quantitative evaluation for the planning task. MMD denotes mean minimum distance, MED denotes mean endpoint distance, and SR denotes the path planning success rate.}
\begin{tabular}{l|ccc}
 \toprule
Model &   MMD [m]$\,\downarrow$	&  MED [m]$\,\downarrow$	& SR$\,\uparrow$ \\
 \midrule
LaneExtraction \cite{he2022lane}   &  157.0  & 339.4 & 0.47 \\
 \midrule
Aggregation (ours)       &   2.2	&  19.7 & 0.46   \\
\bottomrule
\end{tabular}
\label{tab:results-planning}
\vspace{-0.3cm}
\end{table}

\subsection{Path Planning}
\label{sec:planning}

To illustrate the efficacy of our aggregation scheme, we evaluate the quality of the lane graph on a planning task. We generate 1000 randomly selected starting poses in the Miami graph test area from which a plan to a random goal within the graph must be found, using A* search. We place the points such that a maximal optimal route length of $\SI{200}{m}$ is not exceeded. To evaluate the planned routes, we compare the mean minimum distance (MMD) and the mean route endpoint distance (MED) between the paths on the predicted graph and the ground truth graph, respectively. We also report the success rate (SR), indicating the number of cases in which a path between start and goal exists. We list the results in Tab.~\ref{tab:results-planning}. While the SR for our aggregation scheme and the LaneExtraction predictions is similar, the low MMD and MED values of our aggregation scheme indicate that our generated lane graph entails shorter and more direct paths, compared to LaneExtraction. We show additional results in the supplementary material, Sec.~\ref{supp:planning}.

\subsection{Limitations}
Due to its bottom-up architecture, the proposed approach performs well for most evaluated scenes in urban and suburban surroundings but struggles with highly complex graph topologies such as multi-lane intersections or roundabouts. Moreover, due to the iterative formulation of our aggregation scheme, the inference time of our approach increases with the number of nodes and edges. To speed up inference time, future work might include adaptively changing the distance between consecutive virtual agent positions and leveraging efficient neighborhood lookup methods such as k-d trees. Parallel execution of multiple agents would additionally boost run time to match top-down approaches and is feasible in terms of the proposed aggregation scheme.

\section{Conclusion}
\label{sec:conclusion}

In this work, we presented a novel lane graph estimation framework complemented with a novel dataset comprising aerial images. We showed that formulating the lane graph estimation problem as bottom-up graph neural network approach leveraging agent-centric views yields promising results. In addition, we presented a novel aggregation scheme to merge successive lane graphs to produce large-scale solutions. A first-of-its-kind dataset and benchmark for lane graph estimation from aerial images will enable further research in this field. Future work could address end-to-end training and exploiting  further modalities such as onboard vehicle cameras for additional context information.

{\parskip=3pt
\noindent\textbf{Acknowledgements}: This work was partly funded by the German Research Foundation (DFG) Emmy Noether Program grant number 468878300, DFG grant number BU 865/10-2, and a hardware grant from NVIDIA.}

{\small
\bibliographystyle{ieee_fullname}
\bibliography{root}

\begin{thebibliography}{10}\itemsep=-1pt

\bibitem{bandara2022spin}
Wele Gedara~Chaminda Bandara, Jeya Maria~Jose Valanarasu, and Vishal~M Patel.
\newblock Spin road mapper: Extracting roads from aerial images via spatial and
  interaction space graph reasoning for autonomous driving.
\newblock In {\em 2022 International Conference on Robotics and Automation
  (ICRA)}, pages 343--350. IEEE, 2022.

\bibitem{bastani2018roadtracer}
Favyen Bastani, Songtao He, Sofiane Abbar, Mohammad Alizadeh, Hari
  Balakrishnan, Sanjay Chawla, Sam Madden, and David DeWitt.
\newblock Roadtracer: Automatic extraction of road networks from aerial images.
\newblock In {\em Proc.~of the IEEE Computer Society Conference on Computer
  Vision and Pattern Recognition (CVPR)}, pages 4720--4728, 2018.

\bibitem{braso2020}
Guillem Brasó and Laura Leal-Taixé.
\newblock Learning a neural solver for multiple object tracking.
\newblock In {\em The IEEE Conference on Computer Vision and Pattern
  Recognition (CVPR)}, June 2020.

\bibitem{buchner20223d}
Martin B{\"u}chner and Abhinav Valada.
\newblock 3d multi-object tracking using graph neural networks with cross-edge
  modality attention.
\newblock {\em IEEE Robotics and Automation Letters}, 7(4):9707--9714, 2022.

\bibitem{can2021structured}
Yigit~Baran Can, Alexander Liniger, Danda~Pani Paudel, and Luc Van~Gool.
\newblock Structured bird's-eye-view traffic scene understanding from onboard
  images.
\newblock In {\em Proceedings of the IEEE/CVF International Conference on
  Computer Vision}, pages 15661--15670, 2021.

\bibitem{can2022topology}
Yigit~Baran Can, Alexander Liniger, Danda~Pani Paudel, and Luc Van~Gool.
\newblock Topology preserving local road network estimation from single onboard
  camera image.
\newblock In {\em Proceedings of the IEEE/CVF Conference on Computer Vision and
  Pattern Recognition}, pages 17263--17272, 2022.

\bibitem{cattaneo2022lcdnet}
Daniele Cattaneo, Matteo Vaghi, and Abhinav Valada.
\newblock Lcdnet: Deep loop closure detection and point cloud registration for
  lidar slam.
\newblock {\em IEEE Transactions on Robotics}, 38(4):2074--2093, 2022.

\bibitem{chai2020multipath}
Yuning Chai, Benjamin Sapp, Mayank Bansal, and Dragomir Anguelov.
\newblock Multipath: Multiple probabilistic anchor trajectory hypotheses for
  behavior prediction.
\newblock In {\em Conference on Robot Learning}, pages 86--99. PMLR, 2020.

\bibitem{chen2022learning}
Dian Chen and Philipp Kr{\"a}henb{\"u}hl.
\newblock Learning from all vehicles.
\newblock In {\em Proceedings of the IEEE/CVF Conference on Computer Vision and
  Pattern Recognition}, pages 17222--17231, 2022.

\bibitem{djuric2021multixnet}
Nemanja Djuric, Henggang Cui, Zhaoen Su, Shangxuan Wu, Huahua Wang, Fang-Chieh
  Chou, Luisa San~Martin, Song Feng, Rui Hu, Yang Xu, et~al.
\newblock Multixnet: Multiclass multistage multimodal motion prediction.
\newblock In {\em 2021 IEEE Intelligent Vehicles Symposium (IV)}, pages
  435--442. IEEE, 2021.

\bibitem{fong2022panoptic}
Whye~Kit Fong, Rohit Mohan, Juana~Valeria Hurtado, Lubing Zhou, Holger Caesar,
  Oscar Beijbom, and Abhinav Valada.
\newblock Panoptic nuscenes: A large-scale benchmark for lidar panoptic
  segmentation and tracking.
\newblock {\em IEEE Robotics and Automation Letters}, 7(2):3795--3802, 2022.

\bibitem{gosala2022bird}
Nikhil Gosala and Abhinav Valada.
\newblock Bird’s-eye-view panoptic segmentation using monocular frontal view
  images.
\newblock {\em IEEE Robotics and Automation Letters}, 7(2):1968--1975, 2022.

\bibitem{hagberg2008exploring}
Aric Hagberg, Pieter Swart, and Daniel S~Chult.
\newblock Exploring network structure, dynamics, and function using networkx.
\newblock Technical report, Los Alamos National Lab.(LANL), Los Alamos, NM
  (United States), 2008.

\bibitem{halton1964radical}
J Halton and G Smith.
\newblock Radical inverse quasi-random point sequence, algorithm 247.
\newblock {\em Commun. ACM}, 7(12):701, 1964.

\bibitem{he2022lane}
Songtao He and Hari Balakrishnan.
\newblock Lane-level street map extraction from aerial imagery.
\newblock In {\em Proceedings of the IEEE/CVF Winter Conference on Applications
  of Computer Vision}, pages 2080--2089, 2022.

\bibitem{homayounfar2018hierarchical}
Namdar Homayounfar, Wei-Chiu Ma, Shrinidhi Kowshika~Lakshmikanth, and Raquel
  Urtasun.
\newblock Hierarchical recurrent attention networks for structured online maps.
\newblock In {\em Proc.~of the IEEE Computer Society Conference on Computer
  Vision and Pattern Recognition (CVPR)}, pages 3417--3426, 2018.

\bibitem{homayounfar2019dagmapper}
Namdar Homayounfar, Wei-Chiu Ma, Justin Liang, Xinyu Wu, Jack Fan, and Raquel
  Urtasun.
\newblock Dagmapper: Learning to map by discovering lane topology.
\newblock In {\em Proceedings of the IEEE International Conference on Computer
  Vision}, pages 2911--2920, 2019.

\bibitem{liang2019convolutional}
Justin Liang, Namdar Homayounfar, Wei-Chiu Ma, Shenlong Wang, and Raquel
  Urtasun.
\newblock Convolutional recurrent network for road boundary extraction.
\newblock In {\em Proc.~of the IEEE Computer Society Conference on Computer
  Vision and Pattern Recognition (CVPR)}, pages 9512--9521, 2019.

\bibitem{mattyus2017deeproadmapper}
Gell{\'e}rt M{\'a}ttyus, Wenjie Luo, and Raquel Urtasun.
\newblock Deeproadmapper: Extracting road topology from aerial images.
\newblock In {\em Proceedings of the IEEE International Conference on Computer
  Vision}, pages 3438--3446, 2017.

\bibitem{mohan2022amodal}
Rohit Mohan and Abhinav Valada.
\newblock Amodal panoptic segmentation.
\newblock In {\em Proceedings of the IEEE/CVF Conference on Computer Vision and
  Pattern Recognition}, pages 21023--21032, 2022.

\bibitem{petek2022robust}
K{\"u}rsat Petek, Kshitij Sirohi, Daniel B{\"u}scher, and Wolfram Burgard.
\newblock Robust monocular localization in sparse hd maps leveraging multi-task
  uncertainty estimation.
\newblock In {\em 2022 International Conference on Robotics and Automation
  (ICRA)}, pages 4163--4169. IEEE, 2022.

\bibitem{tan2020vecroad}
Yong-Qiang Tan, Shang-Hua Gao, Xuan-Yi Li, Ming-Ming Cheng, and Bo Ren.
\newblock Vecroad: Point-based iterative graph exploration for road graphs
  extraction.
\newblock In {\em Proc.~of the IEEE Computer Society Conference on Computer
  Vision and Pattern Recognition (CVPR)}, pages 8910--8918, 2020.

\bibitem{umeyama1991least}
Shinji Umeyama.
\newblock Least-squares estimation of transformation parameters between two
  point patterns.
\newblock {\em IEEE Transactions on Pattern Analysis \& Machine Intelligence},
  13(04):376--380, 1991.

\bibitem{valada2016convoluted}
Abhinav Valada, Ankit Dhall, and Wolfram Burgard.
\newblock Convoluted mixture of deep experts for robust semantic segmentation.
\newblock In {\em IEEE/RSJ International conference on intelligent robots and
  systems (IROS) workshop, state estimation and terrain perception for all
  terrain mobile robots}, volume~2, 2016.

\bibitem{van2018spacenet}
Adam Van~Etten, Dave Lindenbaum, and Todd~M Bacastow.
\newblock Spacenet: A remote sensing dataset and challenge series.
\newblock {\em arXiv preprint arXiv:1807.01232}, 2018.

\bibitem{vertens2022usegscene}
Johan Vertens and Wolfram Burgard.
\newblock Usegscene: Unsupervised learning of depth, optical flow and
  ego-motion with semantic guidance and coupled networks.
\newblock {\em arXiv preprint arXiv:2207.07469}, 2022.

\bibitem{vodisch2022continual}
Niclas V{\"o}disch, Daniele Cattaneo, Wolfram Burgard, and Abhinav Valada.
\newblock Continual slam: Beyond lifelong simultaneous localization and mapping
  through continual learning.
\newblock {\em arXiv preprint arXiv:2203.01578}, 2022.

\bibitem{wang2022ltp}
Jingke Wang, Tengju Ye, Ziqing Gu, and Junbo Chen.
\newblock Ltp: Lane-based trajectory prediction for autonomous driving.
\newblock In {\em Proceedings of the IEEE/CVF Conference on Computer Vision and
  Pattern Recognition}, pages 17134--17142, 2022.

\bibitem{wilson2021argoverse}
Benjamin Wilson, William Qi, Tanmay Agarwal, John Lambert, Jagjeet Singh,
  Siddhesh Khandelwal, Bowen Pan, Ratnesh Kumar, Andrew Hartnett,
  Jhony~Kaesemodel Pontes, et~al.
\newblock Argoverse 2: Next generation datasets for self-driving perception and
  forecasting.
\newblock In {\em Thirty-fifth Conference on Neural Information Processing
  Systems Datasets and Benchmarks Track (Round 2)}, 2021.

\bibitem{xu2022rngdet}
Zhenhua Xu, Yuxuan Liu, Lu Gan, Yuxiang Sun, Xinyu Wu, Ming Liu, and Lujia
  Wang.
\newblock Rngdet: Road network graph detection by transformer in aerial images.
\newblock {\em IEEE Transactions on Geoscience and Remote Sensing}, 2022.

\bibitem{zhang2021hierarchical}
Li Zhang, Faezeh Tafazzoli, Gunther Krehl, Runsheng Xu, Timo Rehfeld, Manuel
  Schier, and Arunava Seal.
\newblock Hierarchical road topology learning for urban map-less driving.
\newblock {\em arXiv preprint arXiv:2104.00084}, 2021.

\bibitem{zhang1984fast}
Tongjie~Y Zhang and Ching~Y. Suen.
\newblock A fast parallel algorithm for thinning digital patterns.
\newblock {\em Communications of the ACM}, 27(3):236--239, 1984.

\bibitem{zhao2017pyramid}
Hengshuang Zhao, Jianping Shi, Xiaojuan Qi, Xiaogang Wang, and Jiaya Jia.
\newblock Pyramid scene parsing network.
\newblock In {\em Proceedings of the IEEE conference on computer vision and
  pattern recognition}, pages 2881--2890, 2017.

\bibitem{zhou2018d}
Lichen Zhou, Chuang Zhang, and Ming Wu.
\newblock D-linknet: Linknet with pretrained encoder and dilated convolution
  for high resolution satellite imagery road extraction.
\newblock In {\em Proceedings of the IEEE Conference on Computer Vision and
  Pattern Recognition Workshops}, pages 182--186, 2018.

\bibitem{zhou2021automatic}
Yiyang Zhou, Yuichi Takeda, Masayoshi Tomizuka, and Wei Zhan.
\newblock Automatic construction of lane-level hd maps for urban scenes.
\newblock In {\em 2021 IEEE/RSJ International Conference on Intelligent Robots
  and Systems (IROS)}, pages 6649--6656. IEEE, 2021.

\bibitem{zurn2021lane}
Jannik Z{\"u}rn, Johan Vertens, and Wolfram Burgard.
\newblock Lane graph estimation for scene understanding in urban driving.
\newblock {\em IEEE Robotics and Automation Letters}, 6(4):8615--8622, 2021.

\bibitem{zurn2022trackletmapper}
Jannik Z{\"u}rn, Sebastian Weber, and Wolfram Burgard.
\newblock Trackletmapper: Ground surface segmentation and mapping from traffic
  participant trajectories.
\newblock In {\em 6th Annual Conference on Robot Learning}, 2022.

\end{thebibliography}
}

\flushcolsend








\clearpage

\begin{strip}
\begin{center}
\vspace{-5ex}
\textbf{\Large \bf
Learning and Aggregating Lane Graphs for Urban Automated Driving
} \\
\vspace{2ex}

\Large{\bf- Supplementary Material -}\\
\vspace{0.4cm}
\normalsize{Martin B\"uchner$^{*}$, Jannik Z\"urn$^{*}$, Ion-George Todoran, Abhinav Valada, and Wolfram Burgard}
\end{center}
\end{strip}

\setcounter{section}{0}
\setcounter{equation}{0}
\setcounter{figure}{0}
\setcounter{table}{0}
\makeatletter

\renewcommand{\thesection}{S.\arabic{section}}
\renewcommand{\thesubsection}{S.\arabic{section}.\arabic{subsection}}
\renewcommand{\thetable}{S.\arabic{table}}
\renewcommand{\thefigure}{S.\arabic{figure}}

\normalsize

In our supplementary material, we expand upon multiple aspects of our paper. In Sec.~\ref{supp:dataset_details}, we visualize exemplary data from our compiled \textit{UrbanLaneGraph} dataset and detail pre- and post-processing methods. We also discuss the proposed benchmark for evaluating lane graph prediction models.

In Sec.~\ref{supp:metrics} we give additional detail on evaluation metrics. In Sec.~\ref{supp:sampling}, we discuss the sampling of the annotated graph into a representation suitable to train our LaneGNN model. In Sec.~\ref{supp:training}, we provide additional explanations on the model architectures, the training procedures, and hyperparameter selection. In Sec.~\ref{supp:aggregation}, we explain our graph aggregation in more detail compared to the main manuscript. Finally, in Sec..~\ref{supp:successor-lgp}, \ref{supp:full-lgp}, and ~\ref{supp:planning}, we provide additional ablation studies and evaluations for our Successor-LGP, Full-LGP, and Planning tasks, respectively.

\section{UrbanLaneGraph Dataset Details}
\label{supp:dataset_details}

\begin{figure*}
\centering
\includegraphics[width=\textwidth]{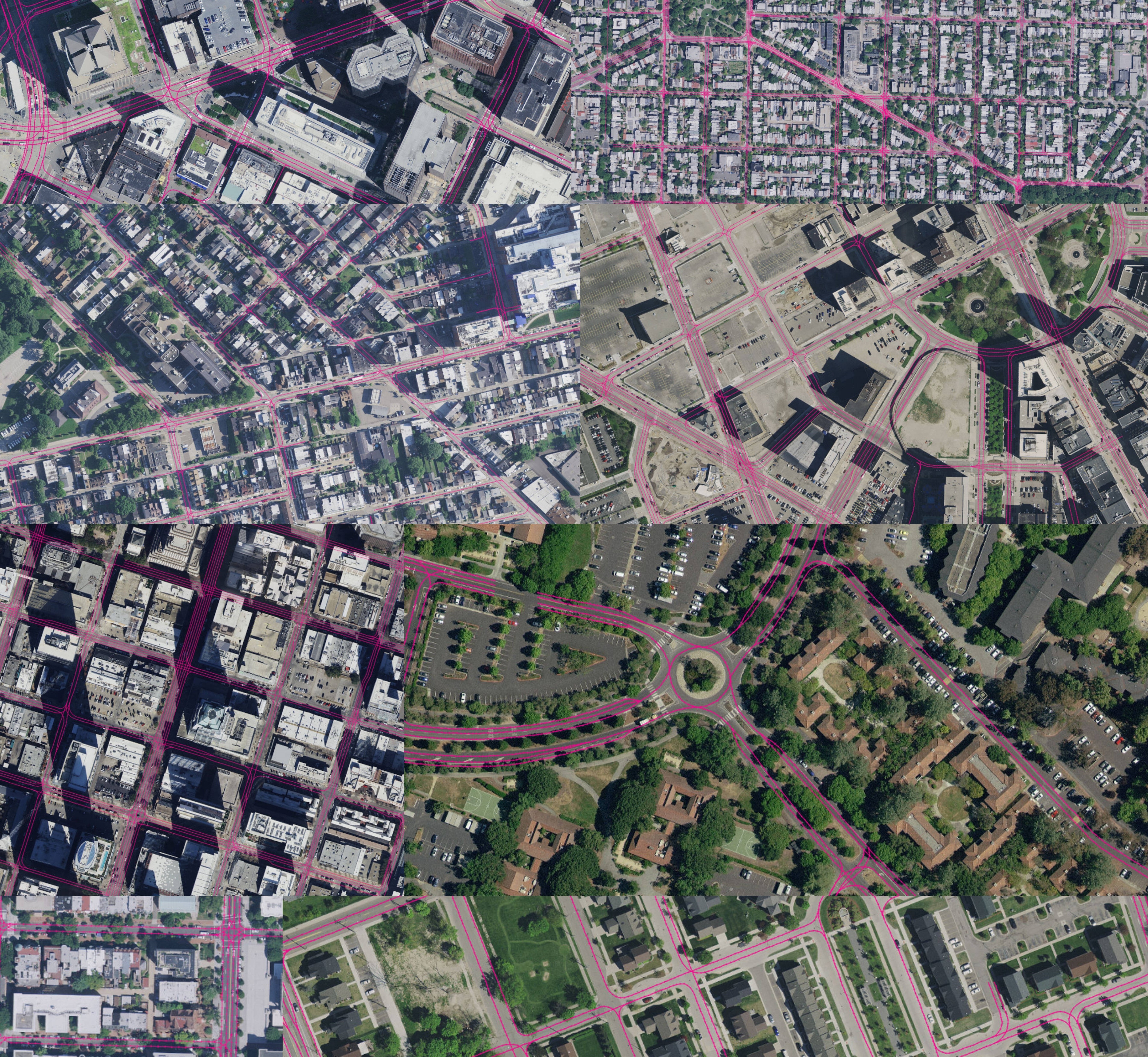}
\caption{Exemplary visualization of aerial imagery with a pixel-aligned lane graph for multiple US cities. Our \textit{UrbanLaneGraph} dataset features a wide range of environments, including rural, suburban, and urban areas. The graph annotations feature a wide range of topological complexity scales from straight road sections to large-scale intersection scenarios with multiple incoming and outgoing lanes. The aerial images provided with the dataset have challenging visual properties such as prominent shadows, and occlusions of streets due to trees and other vegetation.}
\label{fig:dataset_peek}
\end{figure*}

As detailed in the main manuscript, we introduced a large-scale lane graph dataset including aligned high-resolution aerial images. In the following, we give a thorough description of the dataset curation process and how we designed the lane graph prediction benchmark.

\subsection{Dataset Curation}

\subsubsection{Annotation Pre-Processing}

Our graph annotation source is the large-scale Argoverse2~\cite{wilson2021argoverse} autonomous driving dataset, which includes HD map annotations, including lane graphs, for all scenarios entailed in the dataset. In the context of the dataset, a scenario is a small-scale region (\SI{50}{m} diameter). Our goal is to estimate arbitrarily large lane graphs, rendering the per-scenario graph annotation scheme insufficient. We aggregated all scenario graph annotations into a per-city global graph. However, we found that many scenarios overlap while the nodes in the respective overlaps do not have a perfect positional match. Therefore, we implemented an annotation merging procedure, producing the desired globally consistent ground-truth graph.

\subsubsection{Image-Graph Alignment}

The coordinate system of the annotations is not consistent with our aerial image coordinate frame. We therefore, transform the graph annotations into the image coordinate system employing the Kabsch-Umeyama \cite{umeyama1991least} algorithm, in which a set of selected point pairs in the source and in the target frame are aligned, minimizing a least-squares objective function. The solution to the minimization problem comprises the optimal translation, rotation, and scaling that maps points from one frame into the other.

\subsubsection{Regional Train-Test Splits}

We split the dataset into disjoint train and test splits on a geospatial basis. Concretely, for the experiments carried out in this work, we select a challenging subset of the annotated regions within each city as a separate test split. The remaining regions are leveraged for model training. For all tasks and models, we consistently use the same training and testing regions.

\subsubsection{Graph Sampling into Crops}

In order to generate samples for training our LaneGNN model, we sample the dataset into crops. As illustrated in our manuscript, for our successor lane graph prediction task, the input to our model is a $\SI{256}{px} \times \SI{256}{px}$ crop of the original birds-eye view image. A sample consists of the aerial image crop and the lane successor graph that starts at the bottom center position of the crop. Crop positions are selected according to the position of nodes in the lane graph annotations. We also crop the annotated graph in order to use only graph nodes and edges as learning targets that are visible in the respective crop.

As described in the main manuscript, for aggregating the locally predicted successor graphs into a global graph, we facilitate an iterative aggregation scheme where the position and orientation of the next crop are determined according to the graph prediction in the current crop. This procedure can be interpreted as imitation learning from expert data, which are the crops present in the training data. This can produce unstable trajectories that do not follow lanes when deploying the model if the training data distribution does not cover the full distribution of crops from arbitrary positions and orientations. We, therefore, add Gaussian noise w.r.t the crop position $(x_{crop}, y_{crop})$ and orientation $\gamma_{crop}$ when sampling training data crops. Concretely, we sample $x_{crop} \sim \mathcal{N}(x_{gt},\,5)$, $y_{crop} \sim \mathcal{N}(y_{gt},\,5)$, and $\gamma_{crop} ~ \sim \mathcal{N}(\gamma_{gt},\,0.3)\,$. The units are $\SI{}{px}$ and  $\SI{}{rad}$, respectively. This crop sampling scheme allows the model to recover from crop positions and orientations that are not well-aligned with the ground-truth graph.


\begin{figure*}
     \centering
     \begin{subfigure}[b]{0.19\textwidth}
         \centering
         \includegraphics[width=\textwidth]{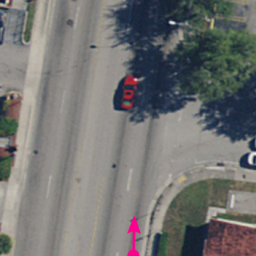}
         \caption{ }
         \label{fig:sample-1}
     \end{subfigure}
     \begin{subfigure}[b]{0.19\textwidth}
         \centering
         \includegraphics[width=\textwidth]{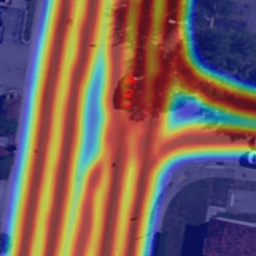}
         \caption{ }
         \label{fig:sample-2}
     \end{subfigure}
     \begin{subfigure}[b]{0.19\textwidth}
         \centering
         \includegraphics[width=\textwidth]{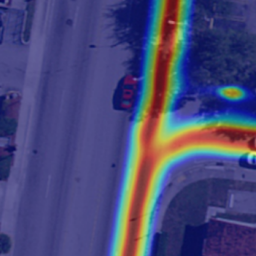}
                  \caption{ }
         \label{fig:sample-3}
     \end{subfigure}
      \begin{subfigure}[b]{0.19\textwidth}
         \centering
         \includegraphics[width=\textwidth]{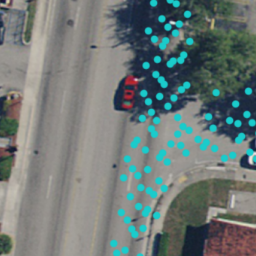}
                  \caption{ }
         \label{fig:sample-4}
     \end{subfigure}
     \begin{subfigure}[b]{0.19\textwidth}
         \centering
         \includegraphics[width=\textwidth]{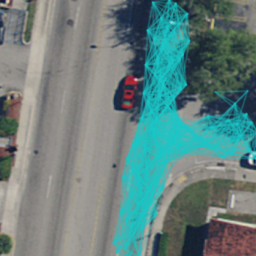}
                  \caption{ }
         \label{fig:sample-5}
     \end{subfigure}
     
        \caption{Visualization of the data modalities that serve as the input to our LaneGNN model. From left to right: a) An aerial BEV image of a local scene, including the virtual agent pose for this crop. b) The regression output $\mathbb{S}_{lane}$ of our lane regression model. c) The regression output $\mathbb{S}_{lane}^{ego}$ of the agent-centric ego lane regressor. d) The lane graph node proposal that potentially holds the successor graph. e) The edge proposal list densely connects neighboring proposal nodes.}
        \label{fig:three graphs}
\end{figure*}

\subsubsection{Centerline Regression Data}

\begin{figure*}
\centering
\includegraphics[width=\textwidth, trim={0cm 18.3cm 0cm 0cm},clip]{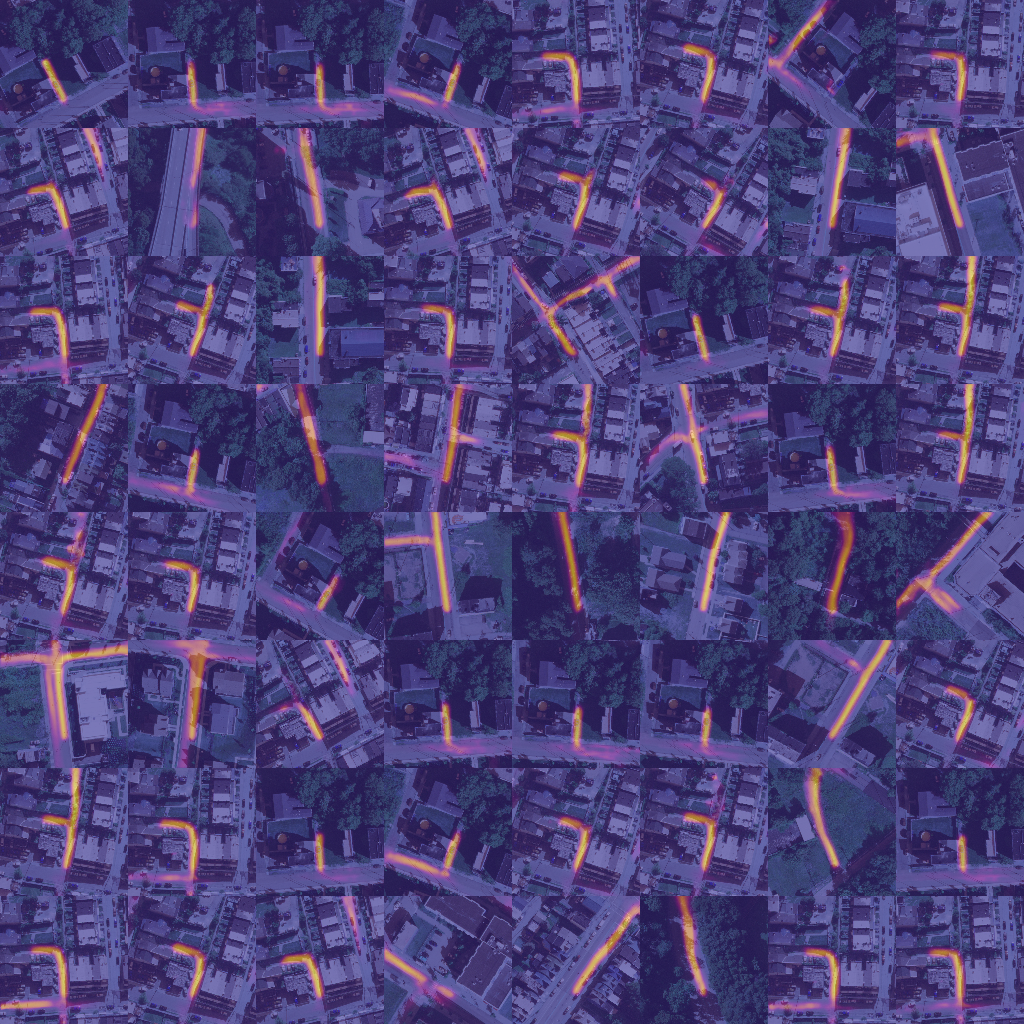}
\caption{Exemplary visualizations of our ego lane regression model on random crops from the test split of our dataset.}
\label{fig:ego_regressor}
\end{figure*}

\begin{figure*}
\centering
\includegraphics[width=\textwidth, trim={0cm 18.3cm 0cm 0cm},clip]{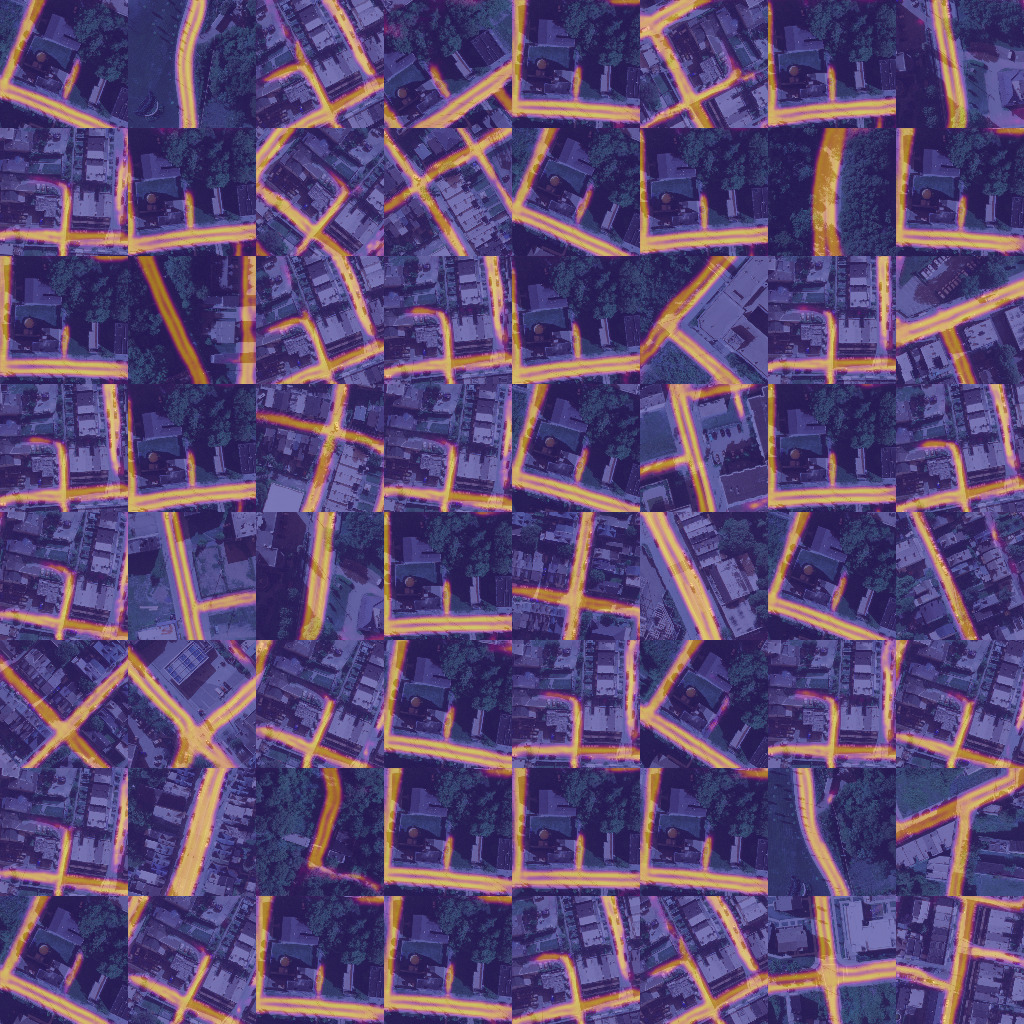}
\caption{Exemplary visualizations of our lane regression model on random crops from the test split of our dataset.}
\label{fig:context_regressor}
\end{figure*}

The centerline regression targets are obtained by rendering an inverse signed distance function from the graph as an image with the same domain as the aerial image crop. Outputs of models trained on this are visualized in Fig.~\ref{fig:context_regressor} (lane centerline regression) and in Fig.~\ref{fig:ego_regressor} (ego lane centerline regression).

\subsection{Benchmark}
\label{supp:benchmark}

We envision the previously described dataset as a reference for evaluating future approaches. To facilitate easier and quantitatively fair comparison between different approaches, we provide an easy-to-use graph prediction benchmarking API. For calculating all our metrics and generating visualizations, we leverage the networkx~\cite{hagberg2008exploring} Python library.

\begin{figure*}
\centering
\includegraphics[width=\textwidth]{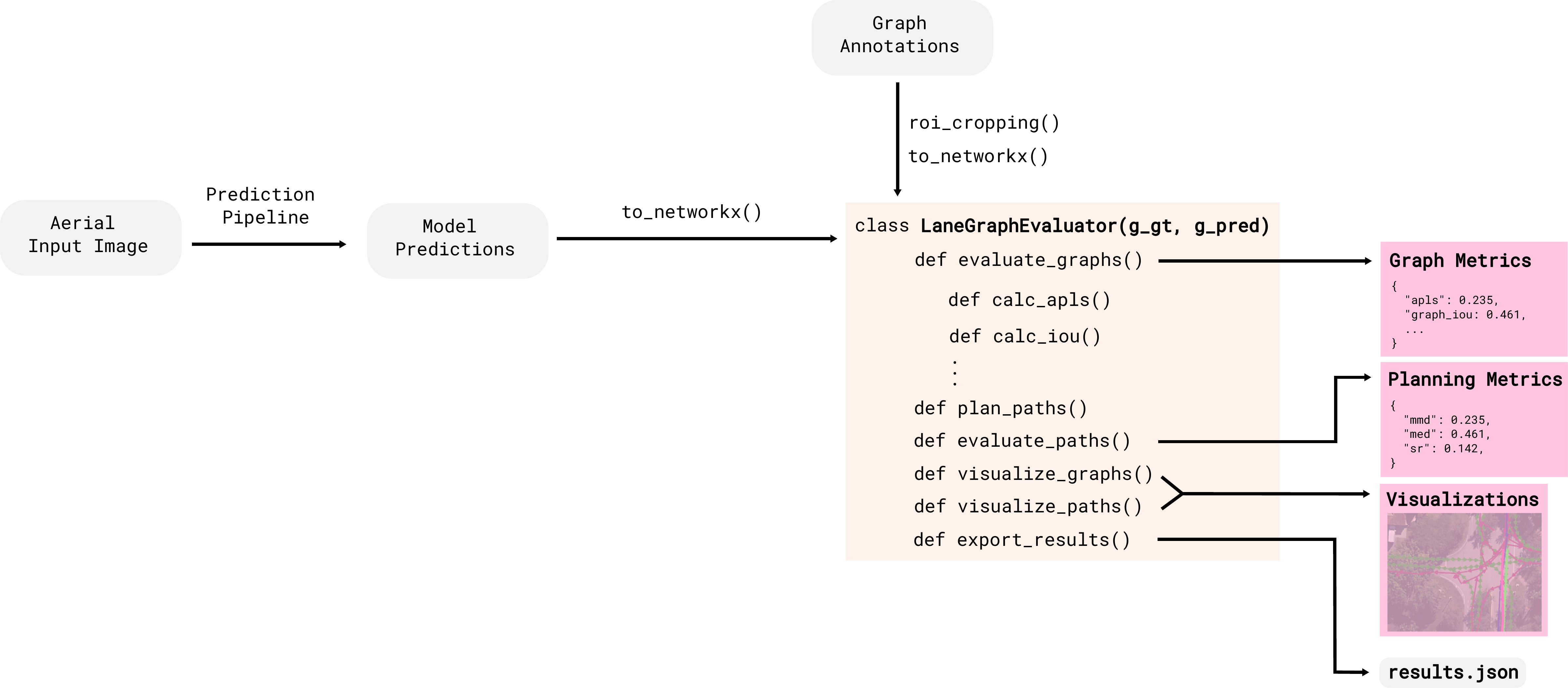}
\caption{Basic information flow for our graph evaluation benchmark. We provide basic routes for evaluating and visualizing lane graph predictions and annotations. Our evaluator expects the predicted graph(s) \texttt{g\_pred} and the ground-truth graph annotations \texttt{g\_gt} as \texttt{networkx} objects.}
\label{fig:schematics-evaluator}
\end{figure*}

The overall information flow of our benchmarking system is visualized in Fig.~\ref{fig:schematics-evaluator}. Given an aerial image, the to-be-evaluated model predicts a graph \texttt{g\_pred}. The \texttt{LaneGraphEvaluator} expects \texttt{g\_pred} to be a networkx-type object; the conversion of the model output into networkx format may be implemented by the authors in a \texttt{to\_networkx()} function. Given a predicted graph in networkx format, our evaluator queries the ground-truth graph annotations and crops the annotations to cover the same region as the predicted graph. Subsequently, all described metrics are evaluated and exported. Optionally, visualizations to rasterized or vector image formats may be produced. Finally, path-planning experiments on the predicted and ground-truth graphs may be conducted, evaluated, and visualized.

\section{Evaluation Metrics Details}
\label{supp:metrics}

In the following, we detail the GEO and TOPO metrics, proposed by He~\etal \cite{he2022lane} which were used, among others metrics, to evaluate or experiments.

\subsection{GEO Metric}

The GEO metric aims to quantify the quality of the spatial position of vertices in the predicted graph, ignoring any topological properties (the existence of edges between the vertices). To this end, following He~\etal \cite{he2022lane}, we densely interpolate the predicted graph $G_{\mathit{pred}} = \{V_{\mathit{pred}}, E_{\mathit{pred}}\}$ and the ground-truth graph $G_{\mathit{gt}} = \{V_{\mathit{gt}}, E_{\mathit{gt}}\}$ such that any two adjacent vertices have the same distance from each other. Subsequently, a 1:1 matching between $V_{\mathit{pred}}$ and $V_{\mathit{gt}}$ is computed, giving rise to the matching precision and matching recall, denoted as GEO precision and GEO recall.

\subsection{TOPO Metric}

Building on top of the GEO metric, the TOPO metric takes vertex connectivity through edges into account as well. Given a vertex pair $(V_{\mathit{gt}}, V_{\mathit{pred}})$ obtained from the GEO metric, subgraphs $S_{\mathit{gt}}^{V_{\mathit{gt}}}$ and $S_{\mathit{pred}}^{V_{\mathit{pred}}}$ are created by walking a maximum distance $D$ on the graphs. We select $D=\SI{50}{m}$. The so-created sub-graphs may be compared according to the GEO metric and averaged over all sub-graphs to obtain the overall TOPO metric. The graph-walking procedure allows penalizing missing links or false-positive graph branches as they will produce poorly aligned sub-graphs $S_{\mathit{gt}}$ and $S_{\mathit{pred}}$.

\section{Target Graph Sampling for Lane Graph Learning}
\label{supp:sampling}

As described in the main manuscript, our LaneGNN model learns to predict node and edge scores for a proposal graph. The ground-truth graph according to the dataset annotations, however, cannot without modification be used as a learning target since its node positions do not correspond to the node points obtained from the Halton-sequence-based node sampling mechanism (see \ref{subsec:halton}). In the following, we describe how a target graph used for learning the LaneGNN model can be obtained from lane graph annotations.

\begin{figure}[ht]
\centering
\includegraphics[width=0.45\textwidth, trim={0cm 12cm 0cm 0cm},clip]{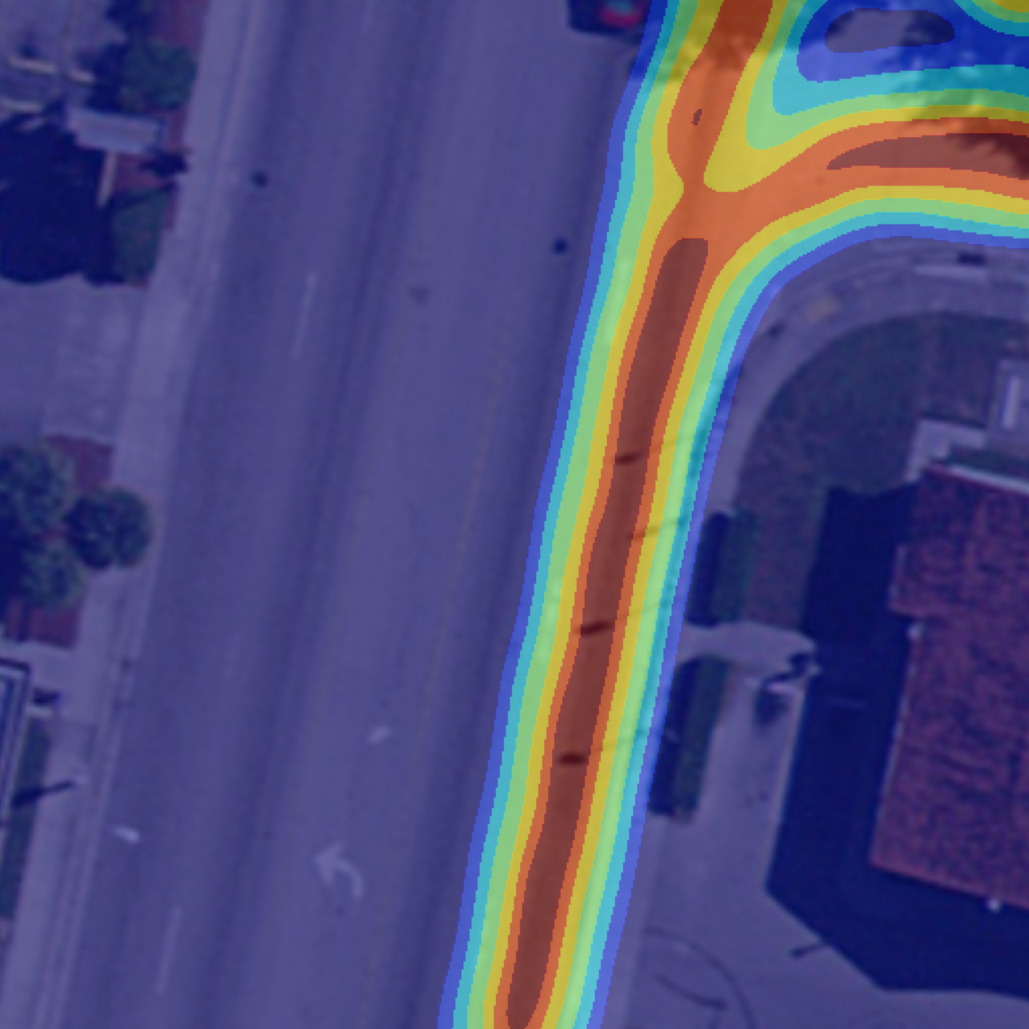}
\caption{The ego lane regressor model output, binned into 6 discrete thresholds from 0 to 1 for visualization purposes. The region of sampled node proposals for our LaneGNN model is sensitive to the choice of threshold cutoff value.}
\label{fig:binned-ego-thresholds}
\end{figure}

\subsection{Node Sampling}
\label{subsec:halton}

In order to generate uniformly distributed 2D node positions in the rectangular image domain, we use Halton sequences. Their advantage over a per-dimension uniform random sampling of points in a 2D rectangle is their comparatively low discrepancy, leading to an approximately equal spacing of neighboring points across the 2D plane, rendering them the preferred choice for sampling random uniformly but non-regular distributed node proposals. For the experiments in our manuscript, we generate $N_{node}=400$ Halton points for each sample. Finally, we filter the generated node positions based on the obtained ego centerline regression mask. Thus, the effective number of nodes is vastly decreased on average. We visualize the sampled Halton points for an exemplary scene in Fig.~\ref{fig:sample-4}.

\subsection{Edge Sampling}

Due to the large number of possible edges in the number of nodes $|E| = \mathcal{O}(N_{node}^2)$, we sample edge proposals between pairs of nodes that have a Euclidean distance $d \in [d_{min}, d_{max}]$ such that only neighboring node are assigned a proposal edge. We visualize the sampled edges for an exemplary scene in Fig.~\ref{fig:sample-5}. Furthermore, we only include edges in the edge proposal list that do not span over regions with low ego centerline regression output.

\subsection{Node and Edge Scoring}

Target node scores are a function of the proposal node distance to the ground-truth graph. More precisely, we score the node according to $s(n_i) = \big(1 - d_{L2}(n_i, G_{gt})\big)^8$ where $d_{L2}(\cdot,\cdot)$ denotes the euclidean distance. We also assign a binary \textit{is-endpoint} label to each node. As described in the main manuscript, in addition to predicting node scores, our LaneGNN model also entails a node classification head, discriminating between lane endpoint nodes and non-endpoint nodes. This distinction allows for efficient proposal graph pruning during model inference. The positive \textit{is-endpoint} label is assigned to the closest node in the proposal graph the ground-truth graph endpoint for each endpoint in the ground-truth graph.

For edge scoring, we empirically found that a binary scoring function (in contrast to a continuous scoring function for the node scores) leads to favorable graph learning performance. To produce this scoring, we leverage a two-step process: First, we evaluate a similar function to the node scoring function but add an angle penalty to the scoring function which penalizes a difference in the relative angle between the proposal edge and the closest edge $e_{gt}^{prox}$ in the ground-truth graph. Concretely: $s(e_{ij}) = \big(1 - d_{\angle}(e_{ij}, e_{gt}^{prox}) d_{L2}(e_{ij}, G_{gt})\big)^8$. Second, we find minimum-cost paths from the lane starting node to the endpoint node(s) through the proposal graph, where the edge traversal cost is the reciprocal of the edge score. All edges along the minimum-cost paths from the start node to the end nodes are assigned a score of one while all remaining edges are assigned a score of zero.

\section{Model Details}
\label{supp:training}

\subsection{Image Skeletonization Baseline}

\begin{table*}
\footnotesize
\caption{Details on the used architecture of the LaneGNN model $\mathcal{M}$. All ReLU-activated MLP layers except for the ones under \textit{Classification} which are ReLU-activated up to the last layer followed by Sigmoid.}
\renewcommand{\arraystretch}{1.5}
\centering
\setlength\tabcolsep{4.7pt}
 \begin{tabular}{rl|rll}
 \toprule
 \textbf{Stage} & \textbf{Layer} & \multicolumn{2}{l}{\textbf{Transformation}}  & \textbf{Parametrization} \\
 \midrule
\multirow{3}{*}{\textit{Encoding}} & Aerial edge features &  $ f_{enc}^{e,bev}(\mathbf{X}_{e,bev}) \,\,\, =$  &  $ \mathbf{H}_{e,bev}^{(0)} \,\,\in \quad\mathbb{R}^{B \times E \times 64}$ & $\operatorname{ResNet-18} \operatorname{(non-pretrained)}$ \\
                    & Geometric edge features &  $f_{enc}^{e,geo}(\mathbf{X}_{e,geo})  \,\,\, = $ & $\mathbf{H}_{e,geo}^{(0)} \,\,\in \quad\mathbb{R}^{B \times E \times 16}$ & $\operatorname{MLP}(4,8,16), \operatorname{ReLU}$ \\
                    & Node features &  $f_{enc}^{v}(\mathbf{X}) \,\,\, = $ & $\mathbf{H}_{v}^{(0)}  \,\, \,\,\, \,\,\in\quad\mathbb{R}^{B \times N \times 16}$ & $\operatorname{MLP}(2,8,16), \operatorname{ReLU}$ \\
\midrule
\textit{Fusion} & Edge Feature Fusion & $f_{fuse}^{e}([\mathbf{H}_{e,bev}^{(0)}, \mathbf{H}_{e,geo}^{(0)}]) \,\,\,= $ & $\mathbf{H}_{e}^{(0)} \quad \,\,\in\quad\mathbb{R}^{B \times E \times 32} $ &  $\operatorname{MLP}(16+64,64,32), \operatorname{ReLU}$  \\
\midrule
\multirow{4}{*}{\textit{\thead{Message \\ Passing \\${\scriptstyle l=1\dots L}$}}} & Edge feature update & $ f_{e}(\mathbf{H}_{i}^{(l-1)},\mathbf{H}_{ij}^{(l-1)}) \,\,\, = $  &  $\mathbf{H}^{(l)}_{e} \quad\, \,\,\in\quad\mathbb{R}^{B \times E \times 32}  $  & $\operatorname{MLP}(16+16+32, 64, 32), \operatorname{ReLU}$  \\
            & \multirow{3}{*}{Node feature update}  & $ f_{v}^{\mathit{pred}}(\mathbf{H}_{i}^{(l-1)},\mathbf{H}_{ij}^{(l-1)}, \mathbf{H}_{i}^{(0)}) \,\,\, = $  &  $\mathbf{H}^{(l)}_{v,pred} \in \quad\mathbb{R}^{B \times E \times 32}  $  & $\operatorname{MLP}(16+16+32, 64, 32), \operatorname{ReLU}$ \\
                            &  & $ f_{v}^{\mathit{succ}}(\mathbf{H}_{i}^{(l-1)},\mathbf{H}_{ij}^{(l-1)}, \mathbf{H}_{i}^{(0)}) \,\,\, = $  &  $\mathbf{H}^{(l)}_{v,succ} \in  \quad \mathbb{R}^{B \times E \times 32}  $  & $\operatorname{MLP}(16+16+32, 64, 32), \operatorname{ReLU}$ \\
                            &  & $ f_{v}(\mathbf{H}_{i,pred}^{(l)}, \mathbf{H}_{i, succ}^{(l)}) \,\,\, = $  &  $\mathbf{H}^{(l)}_{v} \,\quad\, \,\in \quad \mathbb{R}^{B \times E \times 32}  $  & $\operatorname{MLP}(16+16+32, 64, 32), \operatorname{ReLU}$  \\
\midrule
\multirow{3}{*}{\textit{Classification}} & Edge Activation Scores & $f_{cls}^{e}(\mathbf{H}_{e}^{(L)}) \,\,\,= $ & $\mathbb{E}_{e} \quad\quad \,\,\in\quad\mathbb{R}^{B \times E \times 1} $ &  $\operatorname{MLP}(32,16,8,1), \operatorname{ReLU + Sigmoid}$  \\
& Node Activation Scores & $f_{cls}^{v}(\mathbf{H}_{v}^{(L)}) \,\,\,= $ & $\mathbb{S}_{v} \quad\quad \,\,\in\quad\mathbb{R}^{B \times E \times 1} $ &  $\operatorname{MLP}(16,8,4,1), \operatorname{ReLU + Sigmoid}$   \\
& Terminal Node Scores & $f_{cls}^{v,t}(\mathbf{H}_{v}^{(L)}) \,\,\,= $ & $\mathbb{T}_{v} \quad\quad \,\,\in\quad\mathbb{R}^{B \times N \times 1} $ &  $\operatorname{MLP}(16,8,4,1), \operatorname{ReLU + Sigmoid}$   \\

 \bottomrule
 \end{tabular}
\label{tab:suppl-gnn-arch}
\end{table*}

\begin{table}
\footnotesize
\caption{Training details for the three models contained in the full LaneGNN model $\mathcal{M}$.}
\label{tab:training-details}
\centering
\setlength\tabcolsep{1.7pt}
 \begin{tabular}{r|cccc}
 \toprule
 & Batch Size & Epoch & Learning Rate & Weight Decay \\
 \midrule
 Lane Regressor & 8 & $50$   &  $10^{-3}$  &  $10^{-3}$ \\
 Ego Lane Regressor & 8 & $50$ & $10^{-3}$ & $10^{-3}$ \\ 
 Graph Neural Network & 2 & 100 & $10^{-3}$ &  $10^{-4}$ \\
 \bottomrule
 \end{tabular}
\end{table}

Our image skeletonization baseline is obtained by first thresholding the predicted ego centerline regression model output to generate a binary image of high-likelihood successor lane graph regions. We subsequently skeletonize~\cite{zhang1984fast} the binary image and obtain a 1-pixel wide representation of this region (On-pixels in the binary image). Finally, we convert this representation into a graph by considering all On-pixels that have $n=1$ neighbor pixels (lane starting point or endpoint) or $n>=3$ neighbor pixels (lane split point), forming graph nodes. To obtain the graph edges, we add a graph edge for all pairs of nodes that are connected by regions of On-pixels that do not contain any other nodes between the two nodes in consideration. 

\subsection{Centerline Regression Architectures}
For the \textit{centerline regression} and the \textit{ego centerline regression} models, we use the same model architectures. In both cases, we use a PSPNet architecture~\cite{zhao2017pyramid} with a ResNet-152 backbone. The number of input channels is 3 (RGB color channels) for the centerline regression model while it is 4 (RGB + 1 channel output of centerline regression model) for the ego centerline regression model. Additional training parameters can be found in Tab.~\ref{tab:training-details}.

\subsection{LaneGNN Architecture}
\label{supp:lanegnn_arch}
The graph neural network architecture detailed in Table~\ref{tab:suppl-gnn-arch} is trained for 100 epochs using a learning rate of $10^{-3}$ and a batch size of $2$. The used optimizer is $\operatorname{Adam}$ with weight decay $\lambda = 10^{-4}$ and $\beta=(0.9, 0.999)$. Empirically, we found that the network is relatively insensitive to variations in the learning rate, batch size, and the number of message passing steps $L=6$. This is further demonstrated in Tab.~\ref{tab:suppl-lanegnn-abl}.

\section{Graph Aggregation}
\label{supp:aggregation}

In order to aggregate our successive predictions into a globally consistent solution we take the parallelizable approach of running multiple \texttt{drive()}-instances (see Algo.~\ref{algo:drive}) each starting at a different initial pose $\mathbf{p}_{i} = (x_i,y_i,\gamma_i)$ up to a certain maximum number of steps or a maximum number of branches is reached. Each initial pose $\mathbf{p}_{i}$ could either be the actual pose obtained from localization when driving or a pose inferred from a segmentation mask of the aerial image that also includes yaw regression. Next, we follow the procedure described in Algo.~\ref{algo:drive}, which starts \textit{driving} on the birds-eye view image. Sequentially, we predict a successor lane graph $G_{pred}$ that is pruned via multiple runs of Dijkstra's algorithm using the predicted terminal node scores $\mathbb{T}_{v}$ as target nodes while neglecting already traversed corridors between runs.

\begin{figure*}
\centering
\includegraphics[width=\textwidth]{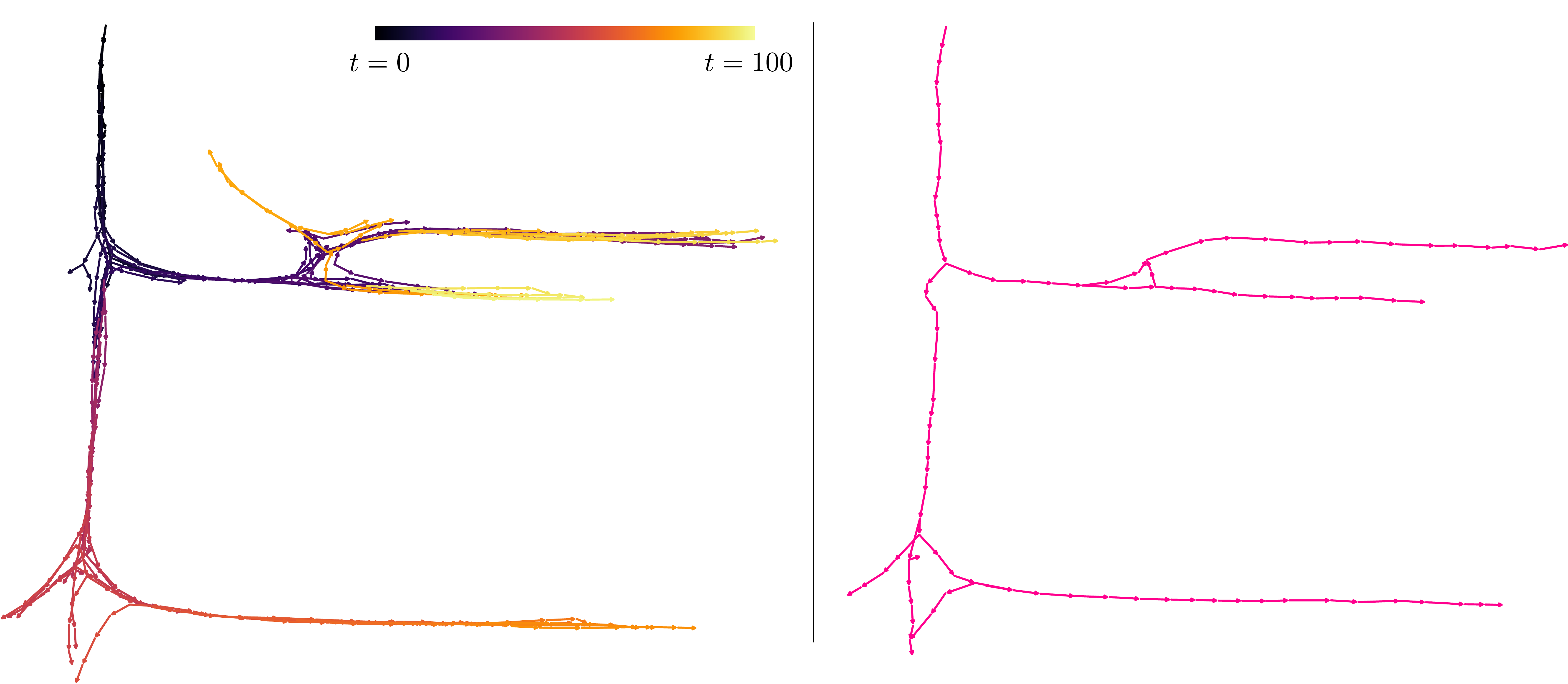}
\caption{Left: Visualization of consecutive successor graph predictions for $t \in [0, 1, ..., 100]$ time-steps. Right: The aggregated graph.}
\label{fig:agg-viz}
\end{figure*}

Before merging the predicted graph $G_{pred}$ into $G_{agg}$, we transform each predicted graph into global coordinates and employ multiple iterations of Laplacian smoothing:
\begin{equation}
    \mathbf{X} \leftarrow \left(\mathbf{I} - \gamma \mathbf{L} \right) \mathbf{X}\ ,
\end{equation}
which smoothens the original node positions $\mathbf{X}$ based on the graph Laplacian $\mathbf{L}$ of the undirected representation of $G_{pred}$. The graph Laplacian is given by 
\begin{equation}
    \mathbf{L} = \mathbf{D} - \mathbf{A},
\end{equation}
where $\mathbf{D}$ is the degree matrix and $\mathbf{A}$ is the adjacency matrix. Hereby, the position of each node is influenced by its first-degree neighbors. The scalar $\gamma$ denotes a smoothing intensity parameter while $\mathbf{I}$ is the identity matrix. The smoothing evens out position irregularities obtained from the initial Halton sampling. In the following, we aggregate the smoothed graph $G_{pred}$ with a consistent representation of all prior predictions $G_{agg}$ (see Sec.~\ref{sec:supp_lat_agg}) and traverse it by steadily exploring the edge that shows the highest successor-tree weight up to a depth of 10. The depth is limited because potentially existing loop closures of the lane graph would induce infinite tree weights. As predicted intersections provide multiple future branches of which only one is explored at a time, we queue the remaining ones for later exploration. As soon as the next edge is selected, we \textit{step forward} by one edge length, which however is subject to hyperparameter optimization. After updating the pose, our approach is able to make predictions on new grounds while further updating and improving the aggregated lane graph representation $G_{agg}$. This is illustrated over 100 time steps in Fig.~\ref{fig:agg-viz}. 

Depending on the mode of operation, we are able to terminate a current branch as soon as a pose shows significant similarity to an already visited pose stored in a list. In addition, we also terminate a branch if a certain maximum number of steps is traveled across a number of branches combined or the number of explored branches exceeds a certain value. Furthermore, every single branch can also be terminated after a designated number of steps. If a branch suddenly terminates, we filter the currently aggregated graph $G_{agg}$ by node out-degrees of two and larger to obtain split points with unvisited edges/poses that provide grounds for further exploration. The edge with the largest successor-tree weight is investigated next if exceeds a certain threshold. We have listed all used parameters in Sec.~\ref{supp:full-lgp}.

Each and every aggregated output $G_{agg}$ produced by one of the \texttt{drive()}-instances are finally aggregated one by one using the function \texttt{aggregate()} provided with Algo.~\ref{algo:aggregate}, which is able to merge arbitrary graphs. In general, it would also be possible to aggregate multiple returns of \texttt{drive()} in an agglomerative fashion to increase inference speed for HD map creation. If necessary as a postprocessing step, we delete all but one parallel branch with the same source and end node if they do not show any intermediate branching and contain less than six consecutive edges.

Overall, we observe an average runtime of $\SI{1}{s}$ per model forward-pass and aggregation step on our development machine (NVIDIA RTX 3090, AMD EPYC CPU @ 3.65GHz). With parallel execution, processing a map tile ($\SI{e6}{m^2}$) requires $\SI{\sim7}{min}$.

\begin{algorithm}
    \footnotesize
    \SetKwRepeat{Do}{do}{while}
    \SetAlgoLined
    \SetAlgoNoEnd
    $G_{agg} \leftarrow \text{ InitializeEmptyAggGraph}()$ \\
    $\textbf{p} \leftarrow \text{InitializePoseOnMap}(\textbf{p}_{init})$ \\
    $\textbf{p} \leftarrow \text{PadSatImageSymmetrically}(\textbf{I}_{bev})$ \\
    $ stepCounter = 0$ \\
    $ branchAlive = True$ \\
    $ branchAge = 0$ \\
    $ branchCounter = 0$ \\
    $ numFutBranches = 1$ \\

    \Do{$numFutBranches \lor branchAlive$}{
        \If{$stepCounter > maxSteps \lor branchCounter > maxNumBranches \lor branchAge > maxBranchAge$}{
            \textbf{break} \\
        }
        \eIf{branchAlive}{
            $branchAge = branchAge + 1$ \\
        }{
        
            $succEdges, numFutBranches \leftarrow \text{GetUntraversedEdgesAtSplits}(G_{agg})$ \\
            $\textbf{p} \leftarrow \text{GetNewPoseOfMaxScoreEdge}(succEdges)$ \\
        }
        $G_{I}, \mathbb{S}_{lane}^{ego}, \mathbb{S}_{lane} \leftarrow \text{ConstrAttribGraphManifold}(\mathbf{I}_{bev}, \mathbf{p})$ \\
        
        $\mathbb{E}_e, \mathbb{S}_v, \mathbb{T}_e \leftarrow \text{PredictSuccessorGraph}(G_{I})$ \\
        $\hat{G} \leftarrow \text{PruneTraverseLaneGraph}(G_{I}, \mathbb{E}_e, \mathbb{S}_v, \mathbb{T}_e)$ \\
        $G_{pred} \leftarrow \text{TransformLaneGraphToGlobalCoords}(\hat{G})$ \\
        $G_{pred} \leftarrow \text{ApplyLaplacianSmoothing}(G_{pred})$
        $G_{agg} \leftarrow \text{Aggregate}(G_{agg}, G_{pred})$ \\

        $branchAlive, \mathbf{p} \leftarrow \text{StepForwAlongCurrBranch}(\mathbf{p}, G_{agg})$
    }
    \textbf{return} $G_{agg}$
    \caption{$\operatorname{drive(\textbf{p}_{init}, \textbf{I}_{sat}})$}
    \label{algo:drive}
\end{algorithm}

\begin{algorithm*}
    \small
    \SetAlgoLined
    \SetAlgoNoEnd
    Add weight to each node $m \in V_{pred}$ based on weight equal to path length to starting pose \\
    Compute edge angles and mean node angles based on predecessor and successor edge angles for $G_{pred}$ and $G_{agg}$\\    
    $G_{agg} \leftarrow removeUnvalidatedSplitsMerges(G_{agg})$ \\
    $\mathbf{D}_{agg,pred}^{L2}$ Compute pair-wise Euclidean distance between $V_{agg}$ and $V_{pred}$. \\
    $\mathbf{D}_{agg,pred}^{ \angle}$ Compute pair-wise mean angle distance between $V_{agg}$ and $V_{pred}$. \\
    $\mathbf{B}_{agg,pred} \leftarrow \left(\mathbf{D}_{agg,pred}^{L2} < \lambda\right) \wedge \left(\mathbf{D}_{agg,pred}^{L2} < \psi \right) $ \\

    \For{$\textbf{A} \in V_{pred}$}{
        $LocalAggGraph(\textbf{A}) = EmptyGraph()$ \\
        \For{$k \in V_{agg}$} {
             \If{$\mathbf{B}_{agg,pred}(\textbf{A},k) == 1$}{
                $LocalAggGraph(\textbf{A}) \leftarrow \text{AddAggEdges}(k)$\\
             }
             \If{$LocalAggGraph(\textbf{A}) \text{ not empty }$}{
                $(\textbf{II},\textbf{I}), a \leftarrow \text{getNearestEdgeAndLatDist}(\textbf{A}, LocalAggGraph(\textbf{A}))$ \\
                $\textbf{I}, \textbf{II} \leftarrow \text{getNearestNodes}(m, LocalAggGraph(\textbf{A}), (\textbf{II},\textbf{I}))$ \\
                \If{$a < a_{thresh}$}{
                    Compute $\textbf{I}^{*}$ and  $\textbf{II}^{*}$ as detailed in Eq.~\ref{eq:closest_update} and Eq.~\ref{eq:sec_closest_update}  \\
                    $G_{agg} \leftarrow \text{UpdateAggGraph}(\textbf{I}^{*},\textbf{II}^{*})$ \\
                }
             }
        }
    }
    \For{$m \in V_{pred} \textbf{ if } m \text{ unmapped to } G_{agg} $}{
        $G_{agg} \leftarrow \text{AddUnmappedNode}(G_{agg},m)$ \\        
    }
    \For{$(m,l) \in E_{pred}$}{
        \If{$m \text{ not mapped to } G_{agg} \wedge l \text{ not mapped to } G_{agg} $}{
            $G_{agg} \leftarrow \text{AddUnconstrainedEdge}(G_{agg}, (m,l)))$ \\ 
        }
        \If{$m \text{ mapped to } G_{agg} \wedge l \text{ not mapped to } G_{agg} $}{
            $G_{agg} \leftarrow \text{AddLeadingEdge}(G_{agg}, (m,l)))$ \\ 
        } 
        \If{$m \text{ not mapped to } G_{agg} \wedge l \text{ mapped to } G_{agg} $}{
            $G_{agg} \leftarrow \text{AddTrailingEdge}(G_{agg}, (m,l)))$ \\ 
        } 
    }
    \textbf{return} $G_{agg}$
    \caption{$\operatorname{aggregate(G_{pred}, G_{agg})}$}
    \label{algo:aggregate}
\end{algorithm*}

\subsection{Lateral Aggregation Scheme}
\label{sec:supp_lat_agg}

\begin{figure}
\centering
\includegraphics[width=0.45\textwidth]{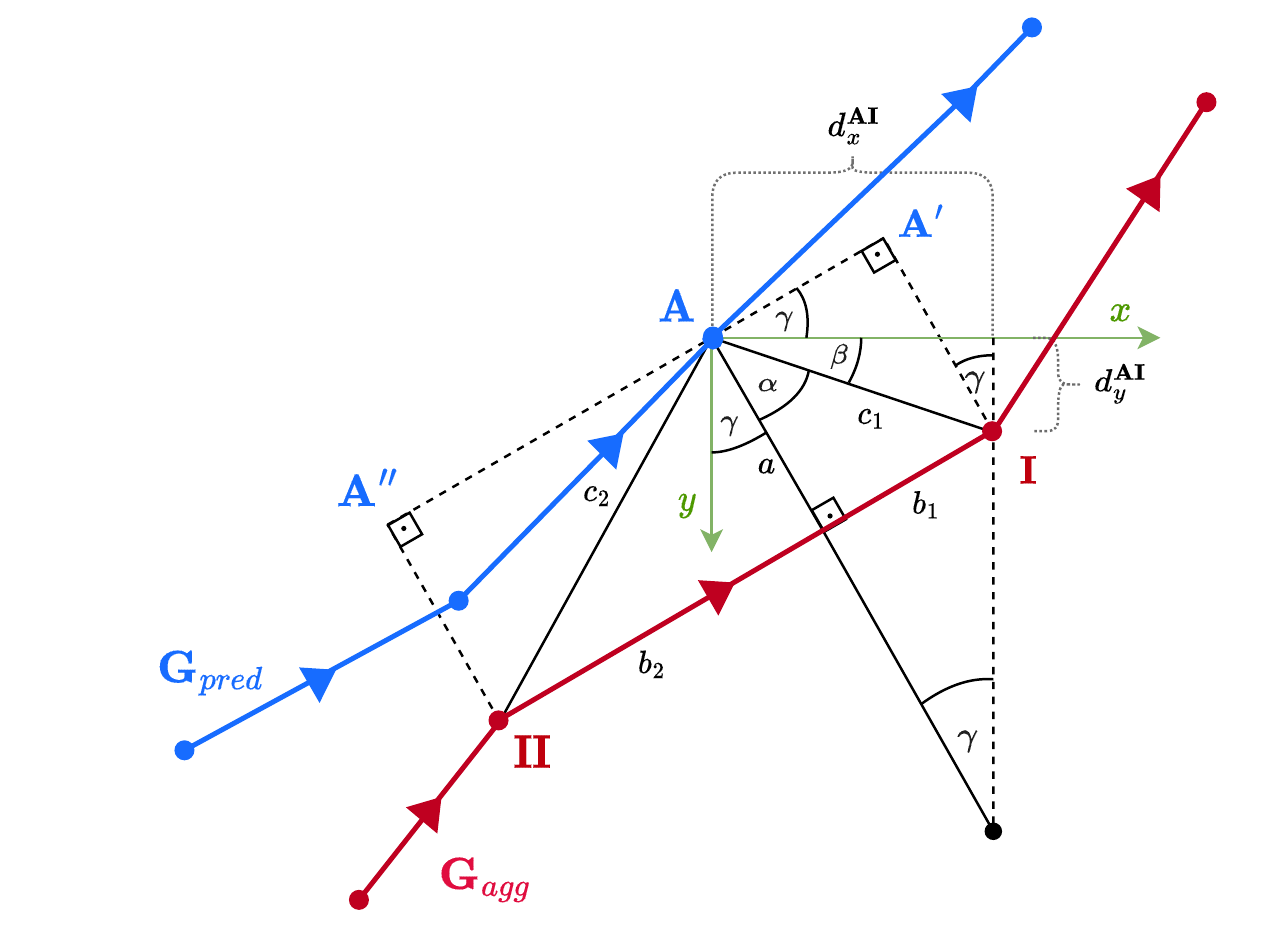}
\caption{Geometric visualization of our lateral graph aggregation scheme. We visualize the predicted graph \textcolor{blue}{$G_{pred}$} in \textcolor{blue}{\texttt{blue}} and the graph to be merged into \textcolor{red}{$G_{agg}$} in \textcolor{red}{\texttt{red}}.}
\label{fig:suppl_lateralagg}
\end{figure}

In the context of lane graphs, particular longitudinal node coordinates are merely a consequence of the chosen sampling distance. The proposed lateral graph aggregation scheme disregards deviations in longitudinal node position. Thus, we merge newly predicted graphs into an existing graph while updating it solely based on lateral deviations. Before doing so, we remove unvalidated splits and merges from the already existing aggregated graph by disregarding all splits or merges that do not exhibit a successor-edge tree of at least length three or do not show sufficient successor-tree (splits) or predecessor-tree (merges) weights. To do so, we construct local temporary aggregation graphs $LocalAggGraph(\textbf{A})$ for each newly predicted node $\textbf{A} \in G_{pred}$ as depicted in Fig.~\ref{fig:suppl_lateralagg}. This is done to decrease the number of distance calculations necessary, which is crucial for large graphs $G_{agg}$ containing thousands of nodes. Thus, each $LocalAggGraph(\textbf{A})$ represents the region of $G_{agg}$ in immediate vicinity of node $\textbf{A} \in V_{pred}$ whereas $G_{pred} = (V_{pred}, E_{pred})$. In the next step, we obtain the nearest edge contained in $G_{agg}$ as well as the lateral distance $a$ to that edge. Similarly, we obtain the closest and second closest nodes $\textbf{I}$ and $\textbf{II}$ incident to that edge. If the lateral distance $a < a_{thresh}$, we see the grounds for updating the node positions of $\textbf{I}$ and $\textbf{II}$ using a weighting-based scheme involving the weights of the involved aggregated nodes (obtained from previous aggregations) as well as the initial weight of each newly predicted node $\textbf{A}$. To do so, we calculate the angles $\alpha,\,  \beta \text{ and,}\, \gamma$ as given by Eq.~\ref{eq:agg_angles} in order to further calculate the lengths $b_1$ and $b_2$ as shown in Fig.~\ref{fig:suppl_lateralagg}:
\begin{equation}\label{eq:agg_angles}\footnotesize
    \alpha = \operatorname{arccos}\left(\frac{a}{c_1}\right), \quad \beta = \operatorname{arctan}\left(\frac{d_{y}^{\textbf{AI}}}{d_{x}^{\textbf{AI}}}\right), \quad \gamma = \frac{\pi}{2} - \alpha - \beta
\end{equation}
\begin{equation}\small
    b_1 = c_1 \operatorname{sin}(\alpha), \quad\quad b_2 = c_2 \operatorname{sin}\left( \operatorname{arccos}\left(\frac{a}{c_2}\right)\right).
\end{equation}

These lengths are used to measure the relative influence of $\textbf{A}$ onto $\textbf{I}$ as well as $\textbf{II}$, respectively. We create temporary nodes $\textbf{A}^{\prime}$ and $\textbf{A}^{\prime\prime}$ to ease merging (Eq.~\ref{eq:agg_virt_nodes}) in the next step:
\begin{equation}\label{eq:agg_virt_nodes}\small
    \textbf{A}^{\prime} = \textbf{A} + b_1 \begin{bmatrix} \operatorname{cos}(\gamma)\\ \operatorname{sin}(\gamma) \\ \end{bmatrix}, \quad
    \textbf{A}^{\prime\prime} = \textbf{A} + b_2 \begin{bmatrix} \operatorname{cos}(\gamma)\\ \operatorname{sin}(\gamma) \\ \end{bmatrix}.   
\end{equation}

Finally, we update the position of $\textbf{I}$ and $\textbf{II}$ using a weighting-based approach as described with Eq.~\ref{eq:closest_update} and Eq.~\ref{eq:sec_closest_update} in the following:
\begin{equation}\label{eq:closest_update}\small
    \textbf{I}^{*} = \frac{1}{\omega_{agg,I} + \omega_{A,I}} \left( \omega_{agg,I} \begin{bmatrix} I_{x} \\ I_{y} \\ \end{bmatrix} 
    + \omega_{A,I} \begin{bmatrix} A_{x}^{\prime} \\ A_{y}^{\prime} \\ \end{bmatrix} \right),\quad\,\,\,\,
\end{equation}
\begin{equation}\label{eq:sec_closest_update}\small
    \textbf{II}^{*} = \frac{1}{\omega_{agg,II} + \omega_{A,II}} \left( \omega_{agg,II} \begin{bmatrix} II_{x} \\ II_{y} \\ \end{bmatrix} 
    + \omega_{A,II} \begin{bmatrix} A_{x}^{\prime\prime} \\ A_{y}^{\prime\prime} \\ \end{bmatrix} \right),
\end{equation}
where $\omega_{A,I}$ and $\omega_{A,II}$ are defined using $c_1, c_2$ as follows:

\begin{equation}\small
    \omega_{A,I} = 1 - \frac{c_1}{c_1 + c_2}, \quad \omega_{A,II} = 1 - \frac{c_2}{c_1 + c_2}. 
\end{equation}

We collect all aggregation information using a merging map. All other nodes $m \in V_{pred}$ that were not mapped to any other node $k \in V_{agg}$ are added as new nodes to $G_{agg}$. Similarly, we add edges $(m,l) \in E_{pred}$ to $E_{agg}$ in a slightly different fashion depending on whether one of the involved nodes has been already mapped to a node $k \in V_{agg}$. Finally, our $\texttt{aggregate()}$ function returns $G_{agg}$.

\begin{figure}
\centering
\includegraphics[width=8.5cm]{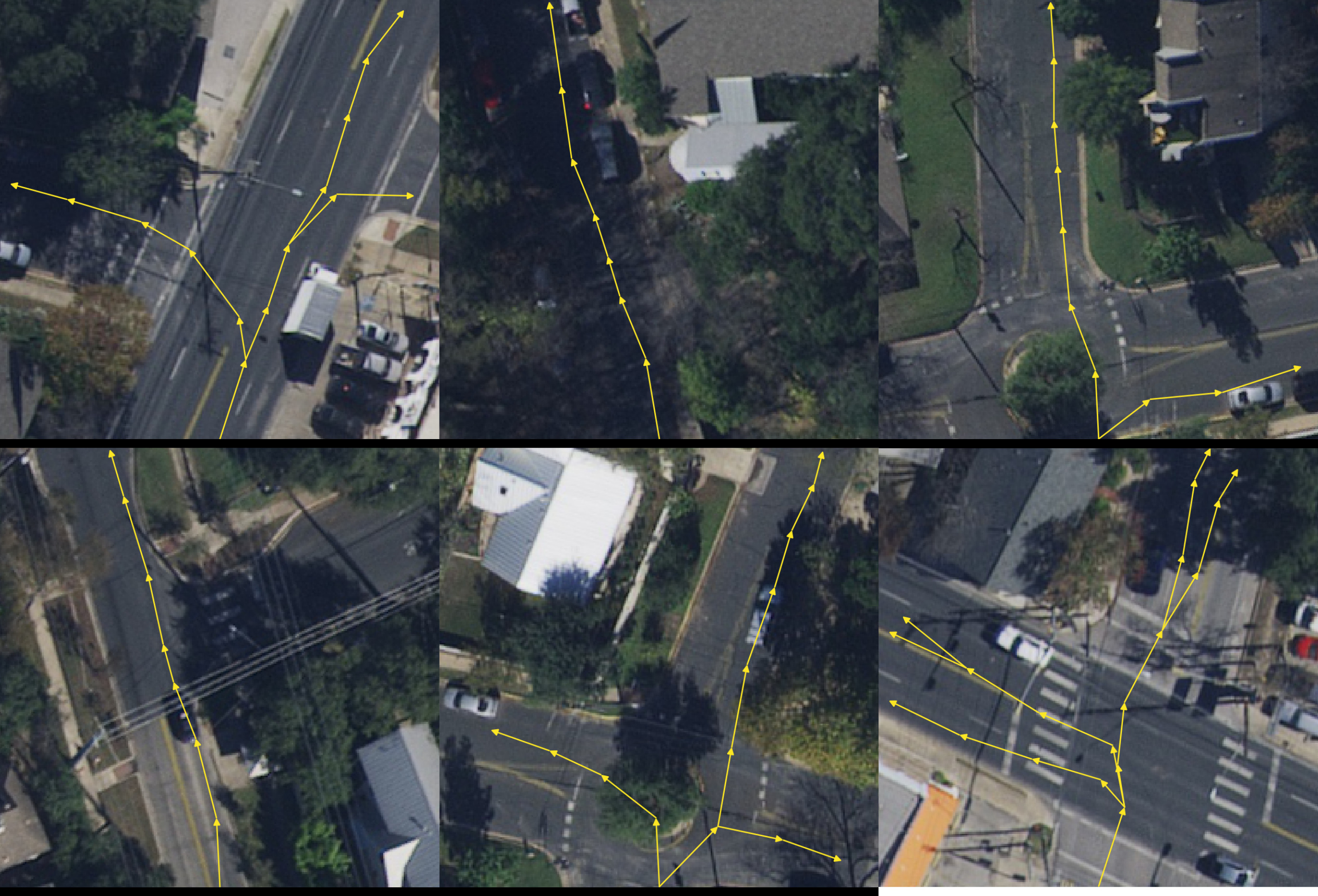}
\caption{Additional qualitative results of our LaneGNN model. In the top row, we illustrate success cases. In the bottom row, we show interesting failure cases. The depicted graphs are not Laplace-smoothed.}
\label{fig:lanegnn_qualitative}
\end{figure}

\section{Extended Results: Successor-LGP}
\label{supp:successor-lgp}

In the following section, we perform additional ablation studies for our LaneGNN model on the Successor-LGP task.

\subsection{Quantitative Results}

\subsubsection{Additional LaneGNN Ablation}
In addition to our observation regarding low sensitivity to minor parameter variations (see Sec.~\ref{supp:lanegnn_arch}), we present a parameter study in Tab.~\ref{tab:suppl-lanegnn-abl} for the LaneGNN architecture as well as the training parameters. The model is trained using both 200 as well as 400 node samples, which is a hyper-parameter that is chosen in the preprocessing stage. Note that the effective number of nodes covering the drivable corridor can be much lower in reality due to ego lane segmentation masking. We observe significantly higher recalls while the precision is roughly the same when increasing from 200 to 400 nodes. The remaining metrics, however, indicate a more drastic performance difference with plummeting values across $APLS, SDA_\text{20}, SDA_\text{50}$ and $Graph IoU$ for fewer nodes. 

In addition to the number of nodes, we show that the inclusion of the node loss term improves recall significantly while also showing higher values across all other metrics. The node regression outputs $\mathbb{S}_v$ are not used during graph traversal but still induce significant performance improvements as they guide the network towards reasonable corridors. The batch size shows the best performance for a value of 2 or 4 depending on the evaluated metric while batch size 1 produces drastically worse outputs. In addition to the batch size, we have observed that a number of 6 to 8 message passing steps produces high precision as well as high recall. A GNN depth of 0 essentially renders the network a classic edge classifier without leveraging the graph structure, which results in drastically reduced performance. Our experiments show that the produced output graph is often not even traversable (using our pruning approach) under that limitation.

Lastly, we train three different GNN architectures: a small-, medium- and, large-size LaneGNN. Our findings support our chosen network architecture presented in the main paper and underpin that a larger network does not necessarily perform better given the test sets across all 4 cities. Overall, we observe that the parameter set used in the main paper (underlined values) yields maximum performance across the metrics evaluated.

\begin{table*}
\footnotesize
\caption{Additional ablations of the LaneGNN model $\mathcal{M}$, measured across the test sets of all 4 cities combined. \underline{Underlined} parameter values denote our presented approach used in the main paper, whereas bold values represent respective \textbf{best} values for each distinct ablation. The tuples under feature dimensions represent the main LaneGNN architecture dimensions ($\texttt{map\_feat\_dim, node\_dim, edge\_dim, msg\_dim, edge\_geo\_dim}$), please see Table~\ref{tab:suppl-gnn-arch}. }
\label{tab:suppl-lanegnn-abl}
\centering
\setlength\tabcolsep{2.7pt}
 \begin{tabular}%
 {>{\raggedright\arraybackslash}p{2.4cm} | %
 >{\raggedleft\arraybackslash}p{2.4cm} | %
 >{\centering\arraybackslash}p{1.8cm} %
 >{\centering\arraybackslash}p{1.8cm} %
 >{\centering\arraybackslash}p{1.4cm} %
 >{\centering\arraybackslash}p{1.0cm} %
 >{\centering\arraybackslash}p{1.0cm} %
 >{\centering\arraybackslash}p{1.7cm}}
 \toprule
 Parameter & & TOPO P/R$\,\uparrow$ & GEO P/R$\,\uparrow$ & APLS$\,\uparrow$ & SDA$_\text{20}$$\,\uparrow$ & SDA$_\text{50}$$\,\uparrow$ & Graph IoU$\,\uparrow$ \\
 \midrule
 \multirow{2}{*}{Number of Nodes} & 200 & \textbf{0.622}/0.561 & \textbf{0.622}/0.560 & 0.077 & 0.094 & 0.162 & 0.172 \\
 & \underline{400} & 0.600/\textbf{0.699} & 0.599/\textbf{0.695} & \textbf{0.202} & \textbf{0.227} & \textbf{0.377} & \textbf{0.347} \\ 
 \midrule
 \multirow{2}{*}{Node Loss Term} & \xmark  & 0.502/\textbf{0.702} & 0.501/\textbf{0.699} & 0.144 & 0.149 & 0.288& 0.335 \\
 &  \underline{$\checkmark$}  & \textbf{0.600}/0.699 & \textbf{0.599}/0.695 & \textbf{0.202} & \textbf{0.227} & \textbf{0.377} & \textbf{0.347}\\
 \midrule
 \multirow{3}{*}{Batch Size} &  1 & 0.421/0.557 & 0.421/0.560 & 0.170 & 0.110 & 0.222 & 0.296 \\
 & \underline{2} &  \textbf{0.600}/0.699 & \textbf{0.599}/0.695 & \textbf{0.202} & \textbf{0.227} & 0.377 & 0.347\\
 & 4  & 0.581/\textbf{0.701} & 0.579/\textbf{0.696} & 0.162 & 0.220 & \textbf{0.387} & \textbf{0.349}\\
 \midrule
  \multirow{5}{*}{GNN Depth}&  0 & 0.163/0.168 & 0.163/0.166 & 0.057 & 0.017 & 0.051 & 0.098  \\
  & 2 &  0.517/0.688 & 0.517/0.684 & 0.144 & 0.152 & 0.310 & 0.331\\
  & 4 &  0.526/0.684 & 0.525/0.680 & 0.145 & 0.162 & 0.320 & 0.330 \\
  &  \underline{6} &  \textbf{0.600}/0.699 & \textbf{0.599}/0.695 & \textbf{0.202} & \textbf{0.227} & \textbf{0.377} & 0.347\\
  & 8 &  0.570/\textbf{0.710} & 0.568/\textbf{0.704} & 0.163 & 0.195 & 0.350 & \textbf{0.349}\\
  \midrule
  \multirow{3}{*}{Feature Dimensions} &  \scriptsize{$(16,8,16,16,8)$} &  0.599/0.655 & 0.598/0.650 & 0.114 & 0.072 & 0.178 & 0.234\\ 
  & \underline{\scriptsize{$(64,16,32,32,16)$}} &   0.600/\textbf{0.699} & 0.599/\textbf{0.695} & \textbf{0.202} & \textbf{0.227} & $\textbf{0.377}$ & \textbf{0.347} \\ 
  & \scriptsize{$(128,32,48,48,16)$} & \textbf{0.616}/0.695 & \textbf{0.613}/0.688 & 0.162 & 0.180 & 0.310 & 0.332\\
 \bottomrule
 \end{tabular}
\end{table*}

\begin{table*}
\footnotesize
\caption{Additional ablations of the LaneGNN model $\mathcal{M}$ for the \textit{Successor-LGP} task trained on respective cities as detailed and evaluated on various combinations of city-respective test sets. PAO, ATX, MIA, and PIT represent the cities of Palo Alto, Austin, Miami, and Pittsburgh, respectively. Bold entries denote the maximum value for the given evaluation set under different training sets. Underlined values denote the numbers presented in the main paper.}
\label{tab:suppl-lanegnn-evalset}
\centering
\setlength\tabcolsep{3.7pt}
 \begin{tabular}{cccc|cccc|cccccc}
 \toprule
 \multicolumn{4}{c|}{Train Set} & \multicolumn{4}{c|}{Eval Set} & TOPO P/R$\,\uparrow$ & GEO P/R$\,\uparrow$ & APLS$\,\uparrow$ & SDA$_\text{20}$$\,\uparrow$ & SDA$_\text{50}$$\,\uparrow$ & Graph IoU$\,\uparrow$ \\
 \midrule
PAO  & ATX & MIA & PIT  & PAO & ATX & MIA & PIT & & & & & \\
 \midrule
 \midrule
\checkmark &  &  &   & \checkmark &  &  &   & 0.584/\textbf{0.744} & 0.582/\textbf{0.739} & 0.177 & \textbf{0.220} & \textbf{0.367} & \textbf{0.378}\\
  \checkmark & \checkmark & \checkmark & \checkmark & \checkmark & & & & \textbf{0.666}/0.702 & \textbf{0.663}/0.696 & \textbf{0.206} & 0.199 & 0.337 & 0.366 \\
 \midrule
 & \checkmark &  &   &  & \checkmark &  &   & 0.468/\textbf{0.726} & 0.468/\textbf{0.727} & 0.123 & \textbf{0.227} & 0.304 & \textbf{0.327}\\
   \checkmark & \checkmark & \checkmark & \checkmark &  & \checkmark & & & \textbf{0.579}/0.660 & \textbf{0.578}/0.660 & \textbf{0.194} & 0.206 & \textbf{0.361} & 0.301 \\
 \midrule
 &  & \checkmark &   &  &  & \checkmark &   & 0.534/\textbf{0.687} & 0.532/\textbf{0.683} & 0.145 & \textbf{0.217} & 0.354 & \textbf{0.330} \\
   \checkmark & \checkmark & \checkmark & \checkmark &  & & \checkmark & & \textbf{0.643}/0.672 & \textbf{0.638}/0.666 & \textbf{0.193} & 0.198 & \textbf{0.371} & 0.315 \\
 \midrule
 &  &  &  \checkmark &  &  &  & \checkmark  & 0.530/\textbf{0.722} & 0.532/\textbf{0.714} & 0.143 & \textbf{0.151} & \textbf{0.307} & \textbf{0.336} \\
\checkmark & \checkmark & \checkmark & \checkmark &  & & & \checkmark & \textbf{0.667}/0.674 & \textbf{0.667}/0.665 & \textbf{0.191} & 0.115 & 0.243 & 0.333 \\

  \midrule\midrule
 
 \checkmark &  &  &  & \checkmark & \checkmark & \checkmark & \checkmark & \textbf{0.603}/0.675 & \textbf{0.601/}0.670 & \textbf{0.203} & 0.206 & 0.376 & 0.339 \\
   & \checkmark &  &  & \checkmark & \checkmark & \checkmark & \checkmark & 0.549/0.686 & 0.548/0.681 & 0.200 & 0.185 & 0.338 & 0.338 \\
 &  & \checkmark &  & \checkmark & \checkmark & \checkmark & \checkmark & 0.578/0.663 & 0.577/0.658 & 0.202 & 0.174 & 0.348 & 0.330 \\
 &  &  &   \checkmark & \checkmark & \checkmark & \checkmark & \checkmark & 0.577/0.643 & 0.574/0.635 & 0.171 & 0.073 & 0.179 & 0.281\\
 \checkmark & \checkmark & \checkmark & \checkmark & \checkmark & \checkmark & \checkmark & \checkmark & \underline{0.600}/\underline{\textbf{0.699}} & \underline{0.599}/\underline{\textbf{0.695}} & \underline{0.202} & \underline{\textbf{0.227}} & \underline{\textbf{0.377}} & \underline{\textbf{0.347}} \\
 \bottomrule
 \end{tabular}
\end{table*}

\subsubsection{City-wise Test Set Results}
In the following, we evaluate the used LaneGNN architecture for different training sets as well as different splits of the test set each consisting of different combinations of cities. For each respective city-split, we show the performance of a LaneGNN instance trained on either only the respective city or on all cities (see Tab.~\ref{tab:suppl-lanegnn-evalset}, upper half). In general, we observe that the specifically trained models do not necessarily perform better on the test-sets of their respective cities. In comparison, the LaneGNN model trained on all cities consistently produces higher precision while showing comparably small reductions in recall. This allows us to state that larger training sets across all cities do not necessarily harm the performance on specific cities when evaluating.

In addition to these findings, we also evaluate different LaneGNN instances trained on city-wise training sets and measure their performance across all cities combined. We observe no major performance drops when testing a city-specific model on all cities except for the model trained on Pittsburgh (see Tab.~\ref{tab:suppl-lanegnn-evalset}, lower half). This essentially means that we can use a model trained on one city and still perform reasonably well in other cities. Nonetheless, the model trained on all cities still shows the average best performance across all cities.

\begin{figure*}
    \centering
    \begin{subfigure}[b]{0.475\textwidth}
        \centering
        \includegraphics[width=\textwidth]{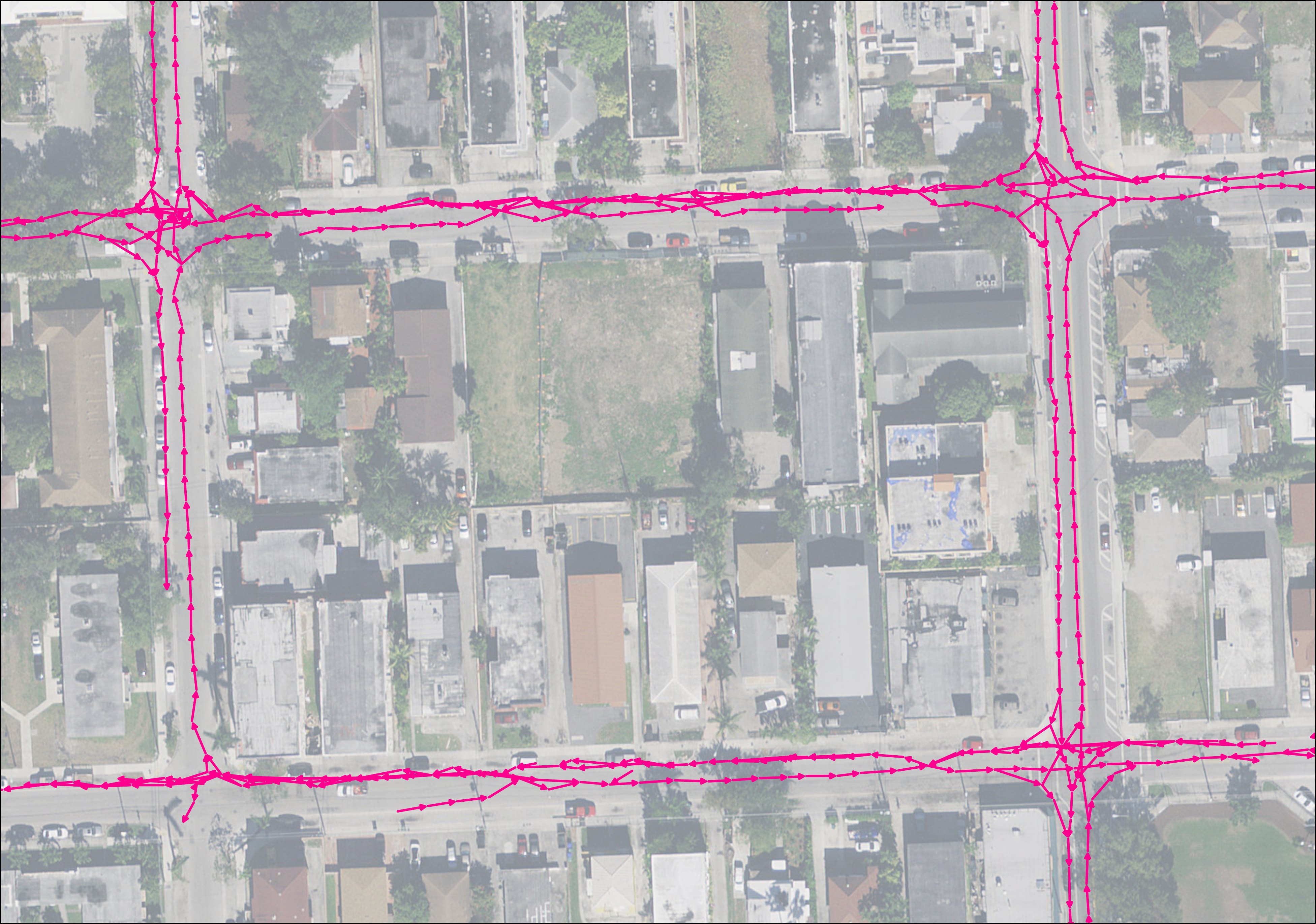}
        \caption{Region 1, naive aggregation}    
        \label{1}
    \end{subfigure}
    \hfill
    \begin{subfigure}[b]{0.475\textwidth}  
        \centering 
        \includegraphics[width=\textwidth]{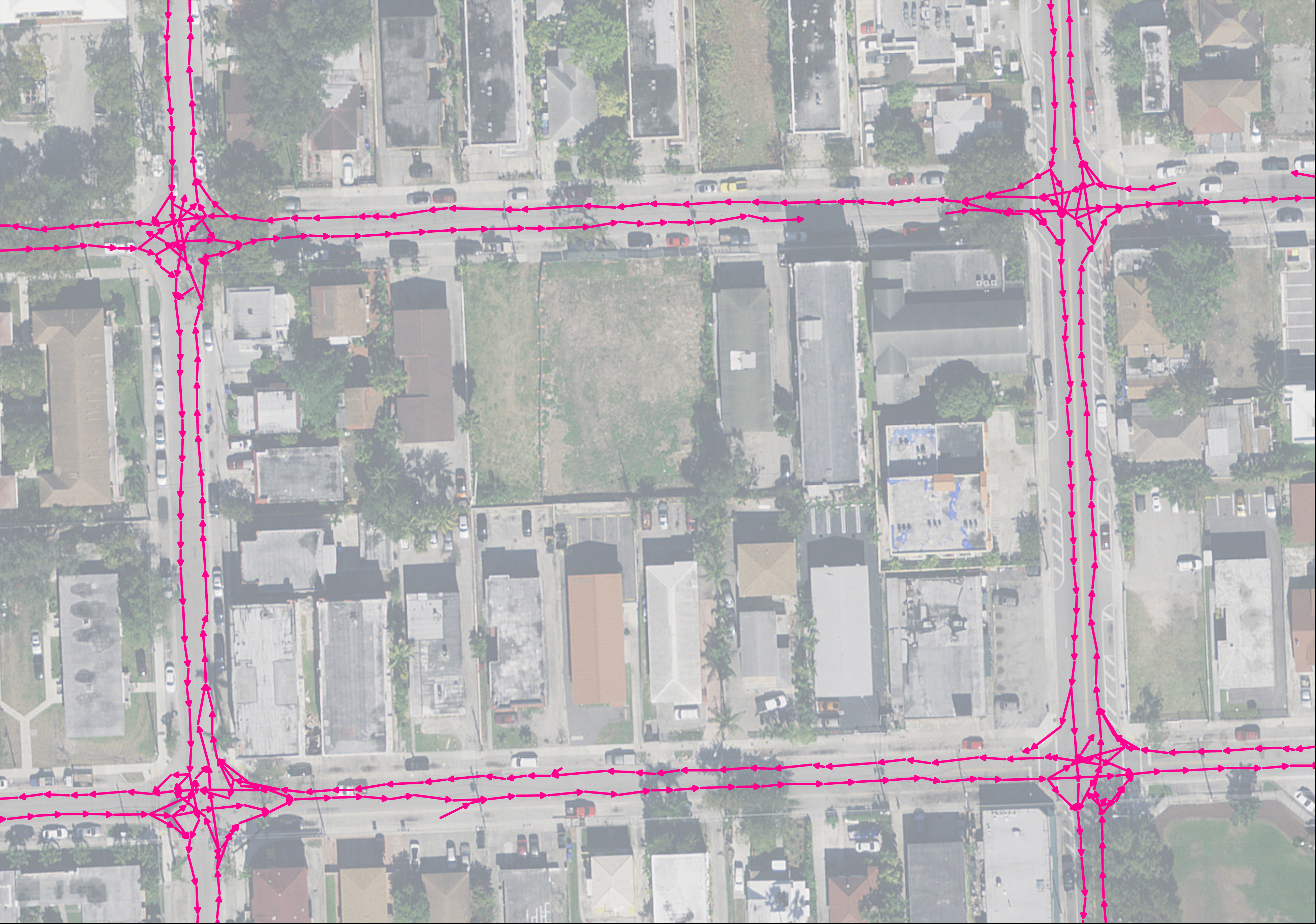}
        \caption{Region 1, our aggregation}    
        \label2{}
    \end{subfigure}
    \vskip\baselineskip
    \begin{subfigure}[b]{0.475\textwidth}   
        \centering 
        \includegraphics[trim={0 0 3cm 0},clip,width=\textwidth]{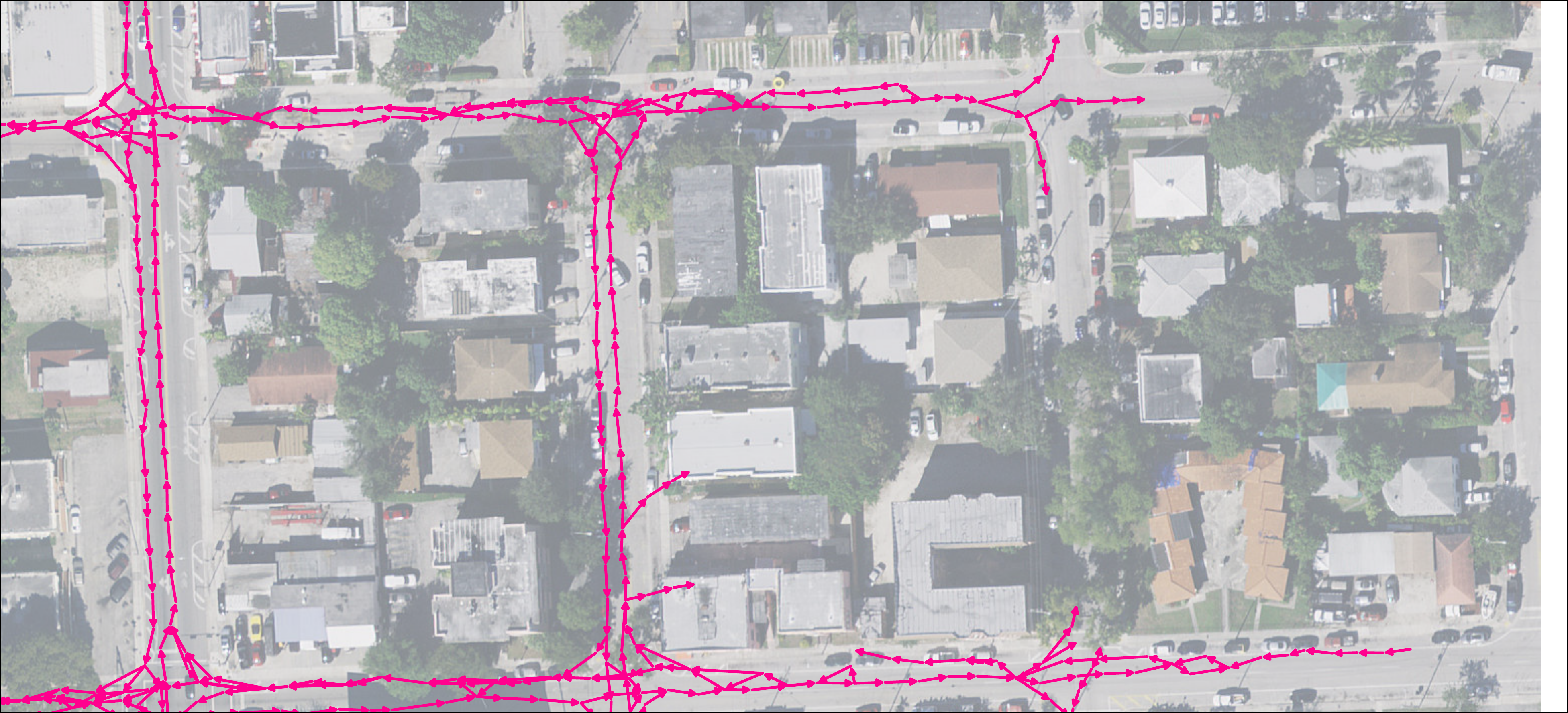}
        \caption{Region 2, naive aggregation}    
        \label{3}
    \end{subfigure}
    \hfill
    \begin{subfigure}[b]{0.475\textwidth}   
        \centering 
        \includegraphics[trim={0 0 2.5cm 0},clip,width=\textwidth]{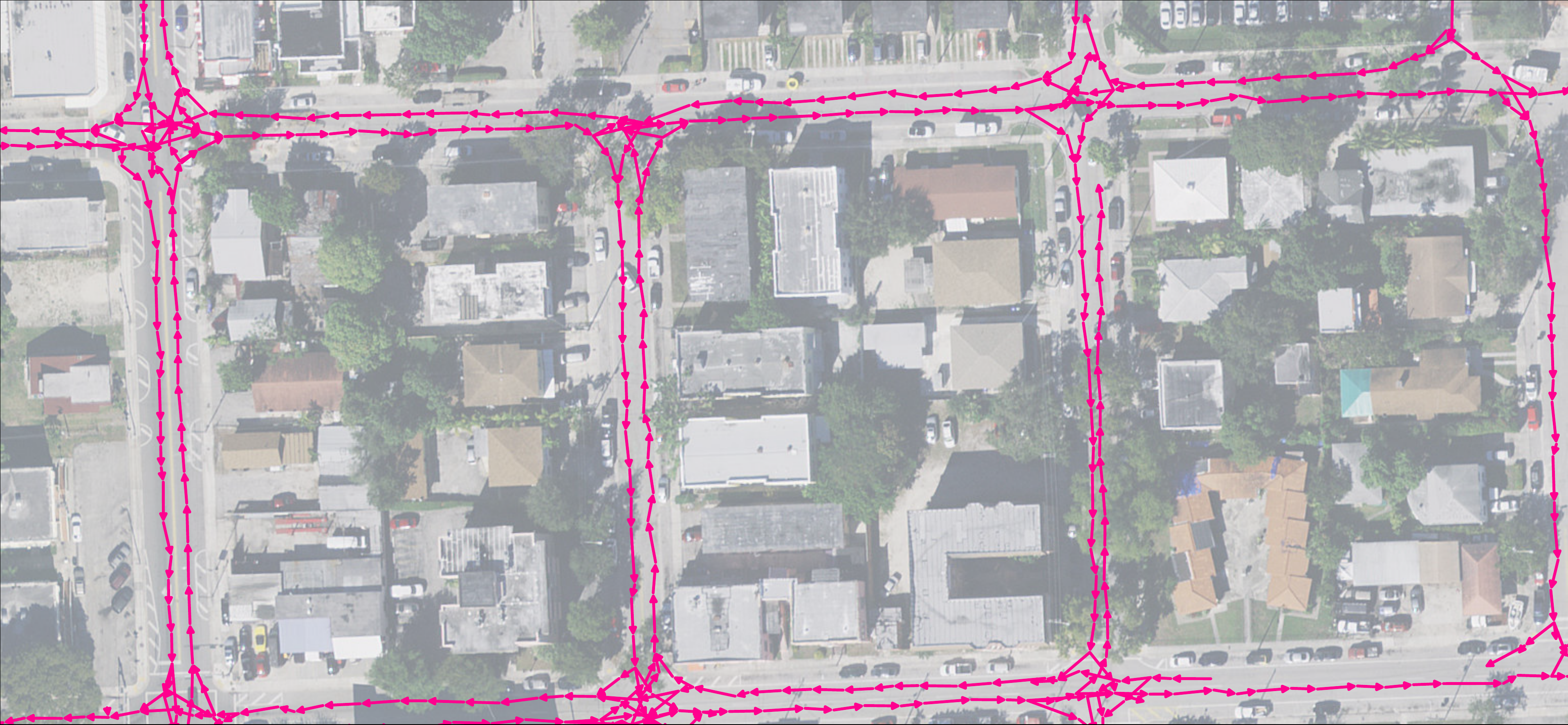}
        \caption{Region 2, our aggregation}    
        \label{4}
    \end{subfigure}
    \caption{Visualizations of the naive and our aggregation scheme for two large-scale areas within the testing region of the city of Miami.} 
    \label{fig:qualitative-fulllgp-appendix}
\end{figure*}

\begin{figure*}
\centering
\includegraphics[width=\textwidth]{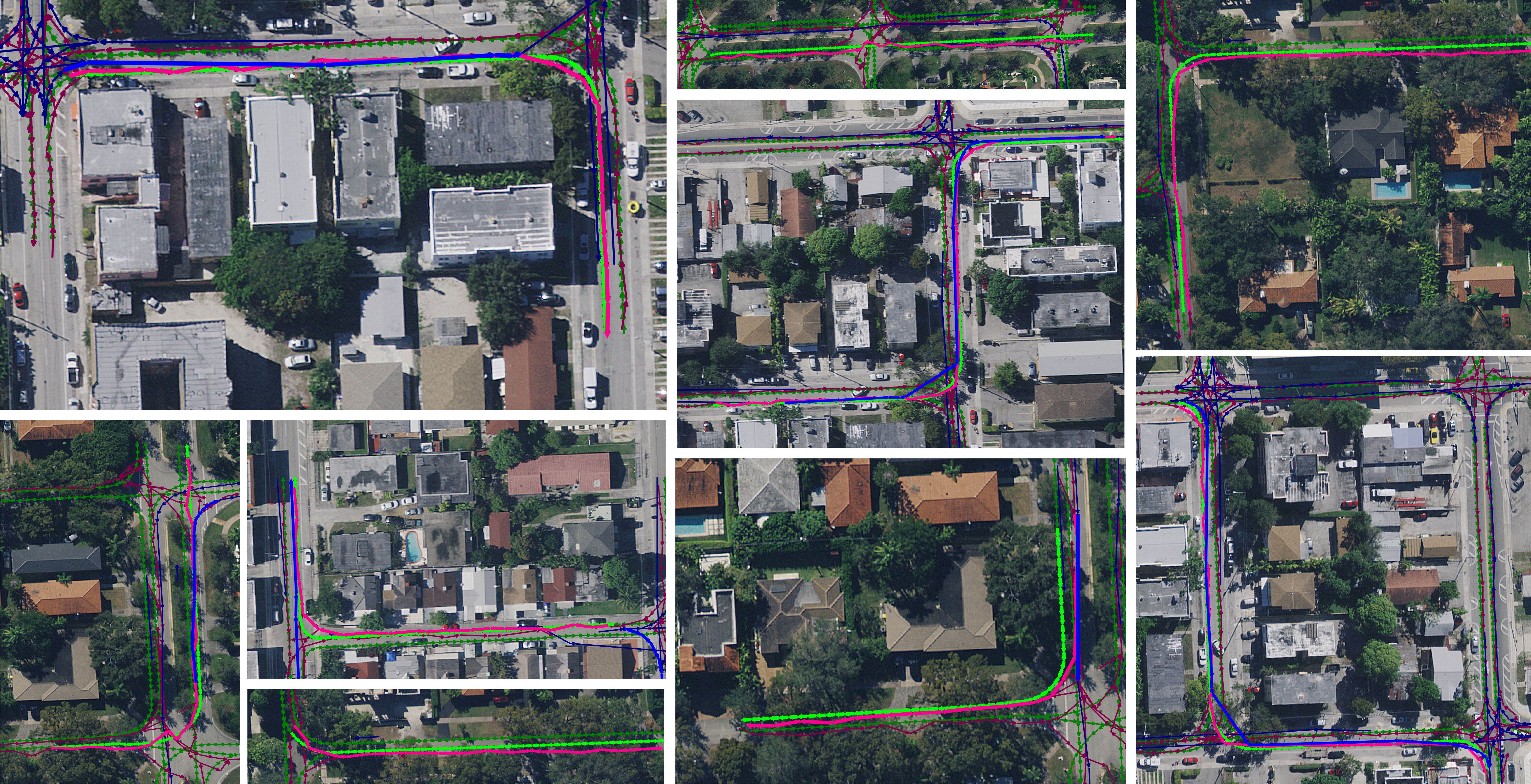}
\caption{Illustrative path planning experiment results. We visualize the ground-truth graph in thin \textcolor{green}{green} and the path planned on the ground-truth graph in bold \textcolor{green}{\textbf{green}}. We employ the same visualization scheme for the graph from \textcolor{blue}{LaneExtraction}, and \textcolor{magenta}{our aggregation scheme}.}
\label{fig:suppl_planning}
\end{figure*}

\subsection{Qualitative Results}

In Fig.~\ref{fig:lanegnn_qualitative}, we visualize the success and failure cases of our LaneGNN model for the Successor-LGP task. Challenging scenes typically entail complex topological structures, such as roundabouts, multi-lane streets, and high-contrast illumination scenarios. In the depicted high-contrast scene, the lane turning right is not detected, leading to a missing branch in the predicted lane graph. The roundabout depicted in the bottom center leads to a topologically correct predicted successor graph, but the predicted waypoints of the left-turning lane do not align with the position of the roundabout center. Finally, in the bottom right scene, an additional left-turning lane going in the opposite direction is predicted by our LaneGNN model. Our aggregation scheme can remove these false-positive graph branches if they are not consistently predicted for multiple successive LaneGNN forward passes.

\section{Extended Results: Full-LGP}
\label{supp:full-lgp}

\begin{table}
\footnotesize
\caption{Additional ablations of the aggregation scheme used in the Full-LGP task. We compare the presented naïve and full aggregation scheme with the reduced variants of the full pipeline (no smoothing of $G_{pred}$, no removal of invalidated splits and merges, and a smaller lateral aggregation threshold).}
\label{tab:add_abl_full_lgp}
\centering
\setlength\tabcolsep{1.7pt}
 \begin{tabular}{r|cccc}
 \toprule
Model &   APLS$\,\uparrow$	&  TOPO P/R$\,\uparrow$ &  GEO P/R$\,\uparrow$ & Graph IoU$\,\uparrow$ \\
 \midrule
 w/o smoothing & 0.105 & 0.452/0.671 & 0.631/0.726 & 0.377 \\ 
 w/o remove s/m  0.102 & 0.480/0.658 & 0.634/0.698 & 0.366 \\    
 $a_{thresh}=10$ & \textbf{0.108} & 0.458/\textbf{0.677} &  0.581/\textbf{0.738} & 0.354 \\  
 \midrule
Naïve               &  0.101   & 0.366/0.654 & 0.523/0.727 & 0.376 \\
Ours  & 0.103  &  \textbf{0.481}/0.670 & \textbf{0.649}/0.689 &  \textbf{0.384} \\
 \bottomrule
 \end{tabular}
\end{table}

In order to generate the test-set results provided in the main paper, we initialized our $\texttt{drive}(\textbf{p}_{init})$ function (Algo.~\ref{algo:drive}) starting at the predicted lane start points as provided by the yaw-segmentation output of LaneExtraction~\cite{he2022lane}. This results in 178 initial poses parametrized by $\textbf{p}_{i} = (x_i,y_i,\gamma_i)$ used for parallel execution of $\texttt{drive()}$ as described in Algo.~\ref{algo:drive}. Our numbers provided in the main paper are generated using the parameters detailed below. 

We make use of thresholding $\mathbf{S}_{lane}^{ego}$ at a value of 0.15, which produces relatively high recalls and thus more connections at intersections. This is complemented by only considering edges with an edge score of 0.5 or higher for graph traversal. For aggregation, we use a radius and distance threshold of $\SI{80}{px}$ and 0.5 radians to filter close nodes and edges of $G_{agg}$ for constructing $LocalAggGraph(\textbf{A})$ instances to speed up the aggregation process. For the actual merging of nodes (Eq.~\ref{eq:closest_update} and Eq.~\ref{eq:sec_closest_update}), we choose a lateral distance $a_{thresh} = 20$ that is used for $\texttt{aggregate()}$ within one $\texttt{drive()}$ instance as well as aggregation of multiple $\texttt{drive()}$ outputs (see Algo.~\ref{algo:aggregate}). Regarding the $\texttt{drive}()$ function, the maximum overall number of steps is 36 over a maximum of 4 branches with a maximum branch-age of 12 each. As described before we smooth the predicted graphs and also remove splits and merges.

In addition, we present results on three additional parameter settings for the \textit{Full-LGP} task in Tab.~\ref{tab:add_abl_full_lgp}. We observe slight performance decreases in precision while the recall is similar or higher when not smoothing predicted graphs. Without removing unvalidated splits and merges, our performance shows slight decreases for GraphIoU, GEO precision, and TOPO P/R. Finally, we lower $a_{thresh}$ to 10, which increases recall but lowers precision, specifically the GEO metric, while the Graph IoU also drops by three percent. As in the main paper, we also list our naïve aggregation scheme for comparison. The results are further illustrated in Fig.~\ref{fig:qualitative-fulllgp-appendix}.

\section{Extended Results: Planning}
\label{supp:planning}

As described in the main manuscript, we evaluate the quality of the generated lane graph on a planning task. Fig~\ref{fig:suppl_planning} illustrates exemplary planned paths obtained on the graphs from the ground-truth annotations, LaneExtraction, and our aggregation scheme, respectively. We observe that for most scenarios, the planned paths from our aggregation scheme closely match the paths planned on the ground-truth graph. Even for long routes with multiple turns, we report a close match between our path and the ground-truth path. We notice some inaccuracies in our planned path where the trajectory deviates from the correct lane and contains slightly misaligned path waypoint positions.

While some of the paths from LaneExtraction show encouraging results, we observe that for many routes, no close match between the LaneExtraction path and the ground-truth path can be observed, also mirrored in the quantitative evaluation of the planning task in the main manuscript. We hypothesize that the LaneExtraction graphs are not always complete and have missing links for occluded or topologically complex regions, resulting in the nonexistence of a path through the graph from start to goal nodes.



\end{document}